\documentclass[acmsmall]{acmart}
\AtBeginDocument{%
  }

\setcopyright{acmcopyright}
\copyrightyear{2024}
\acmYear{2024}
\acmDOI{10.1145/3700886}

\acmJournal{TELO}
\acmVolume{37}
\acmNumber{4}
\acmArticle{111}
\acmMonth{8}

\usepackage{algorithm,algpseudocode}
\usepackage{booktabs}
\usepackage{multirow}
\usepackage{makecell}
\usepackage{graphicx}
\usepackage{subfigure}
\begin{document}

\title{Island-Based Evolutionary Computation with Diverse Surrogates and Adaptive Knowledge Transfer for High-Dimensional Data-Driven Optimization}

\author{Xian-Rong Zhang}
\email{ustczxr@gmail.com}
\orcid{0000-0002-3596-3785}
\author{Yue-Jiao Gong}
\authornote{Yue-Jiao Gong is the corresponding author of this article.}
\orcid{0000-0002-5648-1160}
\email{gongyuejiao@gmail.com}
\affiliation{%
  \institution{South China University of Technology}
  \streetaddress{No. 382, East Ring Road, Panyu District.}
  \city{Guangzhou}
  \state{Guangdong}
  \country{China}
  \postcode{510006}
}

\author{Zhiguang Cao}
\email{zhiguangcao@outlook.com}
\orcid{0000-0002-4499-759X}
\affiliation{%
  \institution{Singapore Management University}
  \country{Singapore}
}

\author{Jun Zhang}
\orcid{0000-0001-7835-9871}
\email{junzhang@ieee.org}
\affiliation{%
  \institution{South Korea and Nankai University}
  \country{China}
}

\renewcommand{\shortauthors}{Xian-Rong Zhang and Yue-Jiao Gong, et al.}
\begin{abstract}
In recent years, there has been a growing interest in data-driven evolutionary algorithms (DDEAs) employing surrogate models to approximate the objective functions with limited data. However, current DDEAs are primarily designed for lower-dimensional problems and their performance drops significantly when applied to large-scale optimization problems (LSOPs). To address the challenge, this paper proposes an offline DDEA named DSKT-DDEA. DSKT-DDEA leverages multiple islands that utilize different data to establish diverse surrogate models, fostering diverse subpopulations and mitigating the risk of premature convergence. In the intra-island optimization phase, a semi-supervised learning method is devised to fine-tune the surrogates. It not only facilitates data argumentation, but also incorporates 
the distribution information gathered during the search process 
to align the surrogates with the evolving local landscapes. Then, in the inter-island knowledge transfer phase, the algorithm incorporates an adaptive strategy that periodically transfers individual information and evaluates the transfer effectiveness in the new environment, facilitating global optimization efficacy. Experimental results demonstrate that our algorithm is competitive with state-of-the-art DDEAs on problems with up to 1000 dimensions, while also exhibiting decent parallelism and scalability. Our DSKT-DDEA is open-source and accessible at: https://github.com/LabGong/DSKT-DDEA.
\end{abstract}



\received{07 September 2023}
\received[revised]{31 March 2024}
\received[revised]{20 July 2024}
\received[accepted]{25 September 2024}

\maketitle

\renewcommand{\sectionautorefname}{Section}
\renewcommand{\subsectionautorefname}{Subsection}
\renewcommand{\figureautorefname}{Fig.}
\renewcommand{\tableautorefname}{Table}
\renewcommand{\equationautorefname}{Eq.}
\newcommand{\algorithmautorefname}{Algorithm}

\section{Introduction}
In recent decades, 
evolutionary algorithms (EAs) have shown promise in solving optimization problems, under the implicit assumption that the evaluation of candidate solutions is straightforward and cheap. However, this assumption does not hold true for many real-world optimization problems. For instance, problems like high-fidelity system optimization \cite{high_fidelity} and human-involved interactive optimization \cite{peole} require computationally intensive numerical simulations or costly physical experiments to evaluate the fitness of solutions \cite{SDDObench}. Moreover, certain working scenarios such as trauma system optimization \cite{trauma_systems} and blast furnace optimization \cite{blast_furnace}, physical constraints even prevent evaluations during the iteration process of EAs.

To mitigate computational costs, surrogate models have been extensively integrated into EAs, leading to Surrogate-Assisted EAs (SAEAs)~\cite{SAEAs}. SAEAs employ a limited amount of data exclusively for training surrogate models to approximate the objective function and/or constraints. Numerous machine learning models, including polynomial regression \cite{polynomial_regression}, Kriging models \cite{Kriging-assisted}, artificial neural networks (ANNs) \cite{ANN}, and radial basis function networks (RBFNs) \cite{RBFN1,RBFN2}, have been leveraged within the SAEA framework. Since the surrogates are built upon data, this subclass of algorithms is recently known as Data-Driven EAs (DDEAs)~\cite{DDEA}. 
However, in the presence of limited training data, it is inevitable for surrogate models to incur approximation errors, potentially misleading the evolutionary search process. 

In addition, the large-scale optimization problems (LSOPs), particularly those with high-dimensional variables, present an ongoing challenge for DDEAs. 
In these scenarios, the search space expands exponentially, 
leading to ``the curse of dimensionality''  problem. 
To tackle this challenge, various decomposition-based methods have been proposed for divide and conquer~\cite{hasanzadeh2013adaptive,CCEA,CCVIL,DG,dg2,rdg, erdg,zhao2008dynamic,refer_v,decomposition, depends}. However, 
the decomposition strategies, aimed at acquiring accurate inter-dependency information, often demand substantial number of fitness evaluations (FEs), which is unsuitable within the research context of DDEAs. Therefore, alternative non-decomposition-based methods have been employed~\cite{LLSO,dynamic_neighbor,dmde}. 
However, this line of technique still carries certain limitations. Specifically, in the realm of offline DDEAs, where the evaluation of candidate solutions is infeasible throughout the process, the high dimensionality of problems makes it difficult for surrogate models to accurately predict candidate values based on limited available data. Additionally, concerns such as insufficient consideration of exploration and exploitation tradeoff, premature convergence, limited scalability, and reduced robustness persist. Consequently, when confronted with LSOPs, the performance of most existing DDEAs experiences significant deterioration.

To address these challenges, this study proposes an offline DDEA approach for solving computationally expensive large-scale problems. Specifically, we introduce an island-based EA incorporating diverse surrogate models and adaptive knowledge transfer, named DSKT-DDEA.
In summary, our contributions are as follows:
\begin{itemize}
\item Diverse surrogates for island diversification: We propose a method to construct multiple diverse surrogates based on 
local datasets, which improves the robustness of fitness prediction. By guiding the subpopulations on different islands with different surrogates, the search diversity of the entire algorithm is greatly enhanced, reducing the possibility of premature convergence.

\item Semi-supervised fine-tuning of surrogates: Prior to optimizing each island, we first evaluate a discrepancy index of individual evaluations on the current island and its neighbors, based on which we perform data augmentation on the local dataset. Using semi-supervised learning, we constantly update the surrogate models during the optimization process, allowing the surrogate models to evolve along with the population and fitting the local fitness landscape. 

\item Adaptive knowledge transfer: We introduce a mechanism for adaptive knowledge transfer by migrating individuals from source islands to target islands and reevaluating them in the new environment. Through periodic evaluation of the migration effectiveness, 
we dynamically adjust migration probability of individuals between the source and target islands and hence the grouping of the population/islands. This method not only facilitates the global optimization effectiveness but also enables adaptive allocation of search resources.

\item Multi-core parallel design: Since the island-based EAs are inherently parallelizable, we also implement a parallel DSKT-DDEA, which demonstrates faster execution speed compared to sequential algorithms. With an increasing number of cores, we achieve sublinear acceleration. Additionally, our algorithm maintains good scalability to tackle higher-dimensional problems. The number of islands can be increased or decreased depending on the number of available computing resources.
\end{itemize}

The structure of the remaining sections in this paper is as follows: \autoref{sec:related_work} offers a summary of the pertinent research and challenges encountered. Subsequently, \autoref{sec:proposed_algorithm} provides a detailed exposition of the algorithm proposed. Following this, \autoref{sec:experimental_analysis} showcases experimental comparisons and analysis. Finally, \autoref{sec:conclusion_future_work} brings forth conclusive findings and lays down the groundwork for future research endeavors.

\section{Related Works}\label{sec:related_work}

\subsection{Data-Driven Evolutionary Algorithms}

DDEAs, which use surrogate models to replace real FEs, can effectively solve computationally expensive optimization problems \cite{SAEA_DDEA}. DDEAs can be classified into two categories: online DDEAs \cite{online_DDEA} and offline DDEAs \cite{offline_DDEA}. The former refers to scenarios where partial candidate solutions can still be evaluated during the optimization process, while the latter pertains to situations where the evaluation of candidate solutions is entirely infeasible throughout the process.

In offline DDEA contexts, data collection during the optimization phase via further experiments or simulations is impractical, 
as each data point often represents a statistical summary of past events or relies on expert involvement. Examples include Trauma System Design~\cite{trauma_systems2}, Magnesium Furnace Optimization~\cite{fused_magnesium_furnaces}, Blast Furnace Optimization~\cite{blast__furnace2}, Ceramic Formulations~\cite{Ceramic_Formula}, and Hardware Accelerators Design~\cite{Hardware_Accelerators}.

 Currently, the main challenge faced by DDEAs is the sparsity of data. The direct outcome brought by data sparsity is the poor quality of the surrogate model, especially for offline DDEAs where new data cannot be generated during the optimization process. Existing research on enhancing DDEAs can be broadly categorized into two main approaches: 

(1) Enhancing the quality and quantity of data: Superior data quality and larger data volumes contribute to the development of more accurate surrogate models. 
TT-DDEA \cite{TT-DDEA} 
employs tri-training to generate pseudo-labels and update the surrogate models, facilitating the optimal utilization of offline data during the prediction process. CL-DDEA \cite{CL-DDEA} uses siamese neural networks and topological sorting mechanisms to emphasize the relativity between individuals and reduce the complexity of downstream tasks. By performing the training on paired data, the training data volume can be increased to the square of the original data volume. BDDEA-LDG \cite{BDDEA-LDG}
and DDEA-PES \cite{DDEA-PES} 
perturb the given dataset to obtain multiple variations and then update a serious of surrogate models correspondingly. 

(2) Enhancing the quality of surrogate models: A common approach is to employ various methods and then perform model ensembles \cite{ensemble_a,ensemble_b,ensemble_c,ensemble_d,ensemble1,ensemble2,ensemble3,ensemble6,ensemble8}. For instance, SRK-DDEA \cite{SRK-DDEA} employs four distinct RBFN as surrogate models. To manage these surrogate models effectively, SRK-DDEA introduces a stochastic ranking strategy, similar to bubble sort. DDEA-SE \cite{DDEA-SE} utilizes the bagging method to construct hundreds of RBFN base learners. Multiple base learners are selectively combined for joint decision-making. Additionally, in the case of MS-DDEO \cite{MS-DDEO}, a model pool is constructed using four RBF models with distinct hyperparameters. Offline DDEA design incorporates two model selection criteria: the model error criterion and the distance bias criterion. This combination yields promising outcomes in selecting the most suitable models for offline optimization.

\subsection{Large-Scale Optimization Methods}
Currently, methods for solving LSOPs can be broadly categorized into two main groups: decomposition-based methods and non-decomposition-based methods:

(1) Decomposition-based methods encompass the process of breaking down a complex, high-dimensional problem into several lower-dimensional subproblems. These subproblems are then independently optimized and collaboratively resolved, aiming to ascertain the optimal solution for the original problem. Cooperative coevolutionary algorithms (CCEAs) \cite{CCEA} serve as widely employed frameworks rooted in decomposition strategies. 
It mainly consists of three steps \cite{step1,step2}: problem decomposition, subcomponent optimization, and collaborative combination. The problem decomposition step can be further divided into static decomposition and dynamic decomposition. 
In the static decomposition-based CCEAs, the size of the subspecies and the number of decision variables are determined prior to the coevolution process and remain fixed during the optimization process~\cite{fix_1,fix_2}. 
However, choosing a fixed number of sub-problems and the interdependencies among them significantly affect the performance of the algorithm \cite{affect1,affect2}. 

Therefore, dynamic decomposition-based CCEAs have been introduced, which explore interdependencies of variables and dynamically allocate variables to different groups 
during the optimization process.  
Chen et al. proposed the Cooperative Coevolution with Variable Interaction Learning (CCVIL) \cite{CCVIL}, which consists of learning and optimization phases. In the learning phase, the interactions between decision variables are detected, grouped, and adaptively optimized. Omidvar et al. proposed an automatic decomposition strategy (DECC-DG) based on testing the permutation of variables in affecting each other \cite{DG}, as well as a variant of the differential grouping algorithm (DG2) \cite{dg2} to detect reliable thresholds for reusing sample points. Ma et al. proposed a merging differential grouping (MDG) \cite{MDG} algorithm based on subset interaction and binary search to improve problem decomposition efficiency. 
Mei et al. proposed routing distance grouping (RDG) \cite{rdg} as an effective decomposition solution. Further, ERDG~\cite{erdg} utilizes historical information to reduce the computational cost of decomposition, and the study in \cite{refer_v} achieved an adaptive balance between convergence and diversity by utilizing reference vector guidance, conducting controlled variable analysis, and optimizing decision variables.

(2) In contrast, non-decomposition-based methods adopt a holistic perspective, considering the high-dimensional original problem as a unified entity. These methods employ diverse algorithms derived from metaheuristics to enhance the search capability within the high-dimensional space. Illustrative examples of such methods include the competitive swarm optimizer (CSO) \cite{CSO,CSO2}, social learning particle swarm optimization (SL-PSO) \cite{SL-PSO}, and level-based learning swarm optimizer (LLSO) \cite{LLSO}. 
There are also some non-decomposition methods that reduce dimensions through feature selection or feature extraction \cite{feature3,feature4,feature5}. 
For instance, Qian et al. \cite{feature10} adopted random embedding techniques to 
enable the algorithm optimize in the low-dimensional solution space but evaluate in the original high-dimensional space. 

\subsection{Large-Scale Data-Driven Evolutionary Algorithms}

For large-scale DDEAs, there are a few decomposition-based algorithms: 
SAEA-RPG \cite{SAEA-RPG} follows the divide-and-conquer approach of CCEAs~\cite{CC-DDEA}. The original problem is decomposed into multiple low-dimensional subproblems, and a surrogate model is built for each subproblem to guide optimization. The results of each subpopulation are then merged into a complete solution. In contrast, SAEA-RFS \cite{SAEA-RFS} performs optimization on multiple subproblems in a sequential manner, where a random subset of dimensions is selected for a specific optimization period. The optimal solutions found for each subproblem replace the corresponding dimensions of the global optimal solution.

Regarding non-decomposition methods, 
SA-COSO \cite{SA-COSO} adopts a multi-population strategy where PSO and SL-PSO alternate 
and share the global best individuals. 
SAMSO \cite{SAMSO} utilizes global and local surrogate models to assist the evolution of two populations, and the algorithm improves search performance through mutual learning between the populations. 
Furthermore, CA-LLSO \cite{CA-LLSO} replaces the conventional real fitness function of LLSO with a gradient-enhanced classifier, selecting the optimal solution through a hierarchical approach instead of considering specific fitness values. ESCO \cite{ESCO} trains a substantial number of local models and employs KLD for selective ensemble. In each generation, it selects an auxiliary subproblem and simultaneously optimizes it alongside the main problem.

Although these online DDEAs have achieved good results in hundreds or even 1000 dimensions, they still possess certain limitations. (1) The majority of research efforts are concentrated on online DDEAs where a few online FEs are still available. Utilizing offline DDEAs based soley on historical data poses a significant challenge in addressing LSOPs. (2) Despite the prevailing performance of decomposition-based methods in LSOPs, their advanced decomposition strategies are challenging to apply in the context of DDEAs due to the requirement for a substantial number of designated FEs. Consequently, the current decomposition-based DDEAs utilize simple random grouping strategies, which challenges their effectiveness. In light of these limitations, this paper aims to develop a offline, non-decomposition-based DDEA. 

\subsection{Island model}
The island model is a popular and effective approach in the field of evolutionary computation, where the population is divided into subpopulations called islands. These islands evolve in parallel through their algorithms, and they periodically exchange solutions through a given topology and migration strategy. The independent evolution on islands allows for focused exploitation of promising regions in the search space, while the periodic migration enables sharing of knowledge and genetic material across islands, promoting exploration of new areas. This combination facilitates a more comprehensive search, increasing the algorithm's chances of finding high-quality solutions in complex optimization problems. Besides, given that high-dimensional problems often demand significant computational resources, the structure of the island model is particularly well-suited for distributed computing environments. This alignment ensures both computational efficiency and scalability for solvers of high-dimensional problems.

When implementing the island model, the number of islands must be customized by the user. If all islands run the same algorithm instance, the island model is considered to be homogeneous. If different algorithms are applied within the islands, the island model is considered to be heterogeneous. In addition, the user must choose the topology structure of the island model \cite{island_cate1, island_cate2}, where the ring topology \cite{control_parameters,island_large,best,constrained,crow_search}, master-slave topology \cite{bee,flower} and fully connected topology \cite{Stigmergy,berth_scheduling} are most commonly used. Furthermore, the connections between islands can be either unidirectional or bidirectional. If the connections between the islands remain unchanged throughout the entire running process, the topology is static. If the connections change during the evolution process, the topology is dynamic. 

Meanwhile, various island migration strategies have been proposed in previous work. In \cite{best}, a strategy of migrating excellent individuals within neighboring islands is proposed. In \cite{dynamic}, a dynamic migration strategy was introduced to avoid premature convergence by dynamically calculating migration probabilities. In \cite{master_slave}, an island model implemented based on a server-worker architecture was described. This strategy controls migration by calculating allelic diversity, thereby avoiding premature convergence. 

The island model, when integrated with the context of DDEA, additionally possess the following advantages. (1)~By introducing diverse surrogate models, we not only enhance the diversity of the model's search strategies, but also enable a more robust exploration of the search space. These diverse surrogates are capable of capturing different aspects or features of the high-dimensional landscape, ensuring a comprehensive investigation of a broader range of potential solutions. (2)~The periodic migration of individuals 
aids in updating and refining the surrogate models. This continuous learning process allows the surrogates to adapt over time, thereby improving their predictions and, consequently, the efficiency of the search process. 

\subsection{Basics of RBFN}
\begin{figure}[h]
  \centering
  \includegraphics[width=0.4\linewidth]{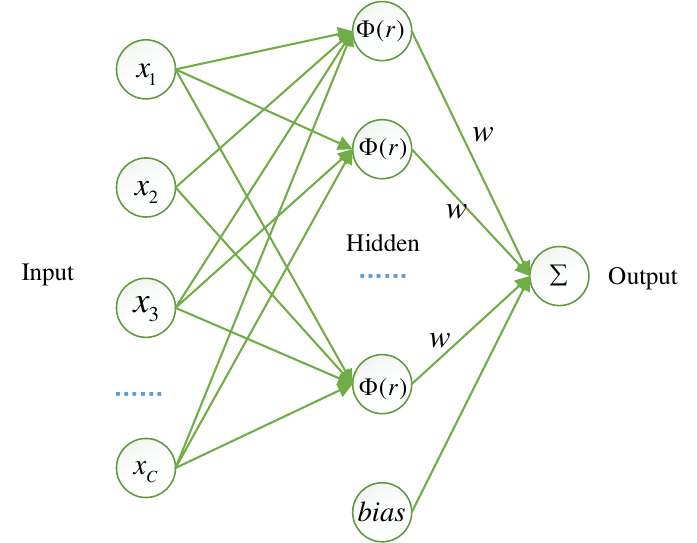}
  \caption{RBFN structure.}
  \label{fig:rbfn}
\end{figure}

RBFN typically consists of three layers, including the input layer, hidden layer, and output layer, as shown in \autoref{fig:rbfn}.

In RBFN, each hidden layer neuron serves as the center of a radial basis function, and the distance between the input and the center point is used as the variable for the function. The output of RBFN can be described by the following equation:
\begin{equation}
Output=\sum\limits_{c=1}^{C}{{{\varpi }_{c}}{{\Phi }_{c}}}+bias
\end{equation}
where $C$ represents the number of hidden layer neurons, ${{\varpi _c}}$ represents the weight between the ${c}$-th hidden layer neuron and the output layer, ${{\Phi _c}}$ represents the output of the ${c}$th hidden layer neuron, and bias is the bias term.

The main difference between RBFN and a typical three-layer neural network lies in the hidden layer. The hidden layer in RBFN uses radial basis functions to transform the inputs into nonlinear outputs. The role of radial basis functions is to map the inputs from a low-dimensional space to a high-dimensional space, thus transforming the low-dimensional linearly inseparable problem into a high-dimensional linearly separable problem. Radial basis functions are a class of functions where the output values are only related to the distance to the center point, for which many kernel functions can be used. 
This study uses the Gaussian kernel function: 
\begin{equation}
\Phi (\| x-ct \|)=\exp (-\frac{{{\left\| x-ct \right\|}^{2}}}{2{{\sigma }^{2}}})
\end{equation}
where $ct$ denotes a center point; $\sigma $ represents the expansion constant, which controls the influence range of each neuron represented by the center point. 
The k-means clustering algorithm is used to select the center points of the radial basis network \cite{rbfn_cluster}. 

Once the radial basis function is determined, it can be trained.
Due to the fact that any input-output in the training data satisfies the following equation:

\begin{equation}
\sum\limits_{c=1}^{C}{{{\varpi }_{c}}}\Phi (\left\| x-ct{_{c}} \right\|)+bias=y
\end{equation}
A commonly used method is to compute the pseudoinverse of the matrix (e.g., singular value decomposition or least squares method) to obtain the solution to the linear equation system \cite{rbfn_gc}. Hence, compared to other feedforward neural networks, RBFN requires lower computational costs and has a faster training speed.

\section{Proposed Method}
\label{sec:proposed_algorithm}
\subsection{Overall Framework}
\begin{figure}[h]
  \centering
  \includegraphics[width=1\linewidth]{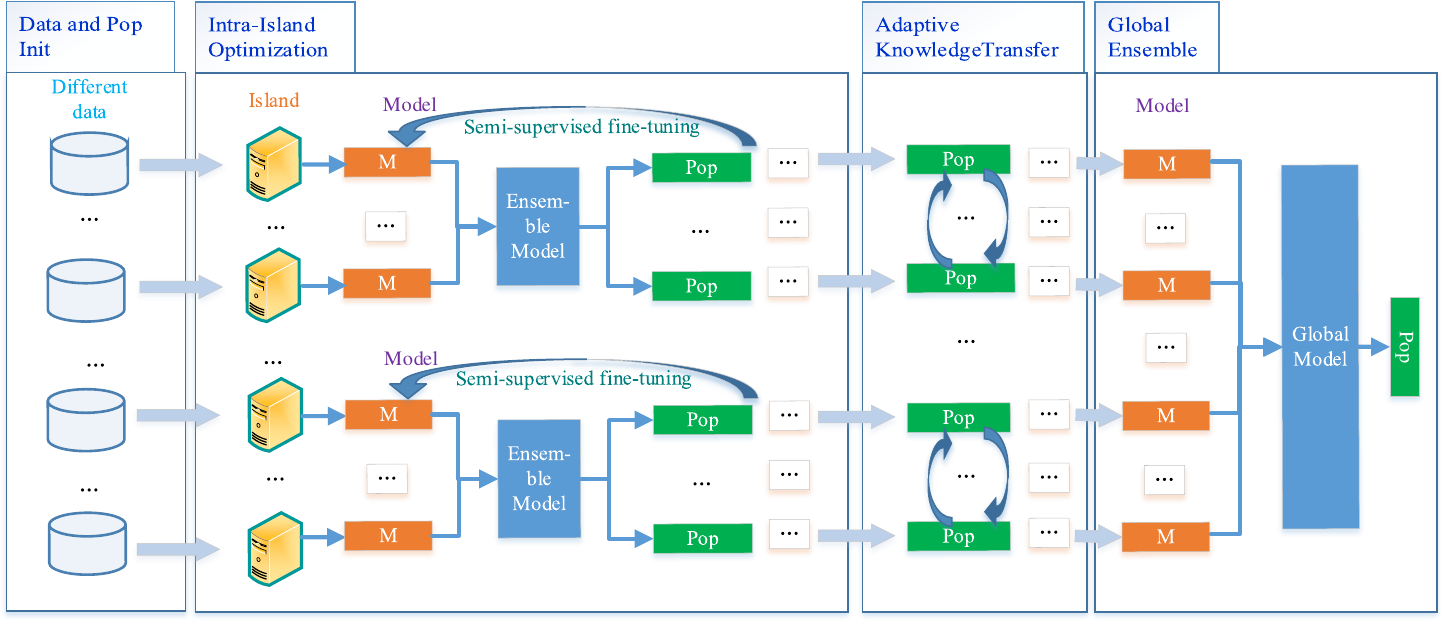}
  \caption{The overall framework of DSKT-DDEA.}
  \label{fig:1}
\end{figure}

The overall architecture of DSKT-DDEA is shown in \autoref{fig:1}. Generally speaking, our optimization consists of two key phases: the intra-island optimization phase and the adaptive knowledge transfer phase. 

In the intra-island optimization phase, to achieve local optimization and enhance group diversity, we first construct diverse surrogate models (\autoref{sec:3.2}) and multiple isolated local sub-populations based on offline data. Each sub-population is guided and environmentally selected by its specific model. Then, in the process of local optimization, we design semi-supervised ensemble learning to enhance the surrogate models (\autoref{sec:3.3}) so that they can receive timely feedback on the distribution information during the continuous optimization process of the population. In this way, in each generation of the optimization process, the surrogate models can continuously learn knowledge from the new population and fine-tune to suitable fitness landscapes. Then, environmental selection is performed through ensemble integration of different surrogate models (\autoref{sec:3.4}).

The adaptive knowledge transfer phase (\autoref{sec:3.5}) can enhance the global exploration capability of the model. In this phase, individuals in the source island are regularly transferred to the target island and re-evaluated in the new environment. During the migration, an attractiveness mechanism based on rank effectiveness is used to evaluate the effectiveness of knowledge transfer and update the migration probability between islands. 
This realizes adaptive allocation and dynamic update of search resources.

\subsection{Diverse Surrogates} \label{sec:3.2}

Before conducting the optimization in DSKT-DDEA, it is necessary to train the surrogate model for each island based on offline data. However, the high dimensionality of problems presents a significant challenge for training the surrogate models to accurately predict fitness values with limited available data. As a result, the performance of most existing DDEAs significantly deteriorates in such cases. To address this, we propose a diverse surrogate model construction method that makes full use of the available data to enhance the robustness of predictions, which thereafter guides diverse populations to evolve and explore in different islands.

First, it is necessary for each island to sample specific data subsets from the overall offline dataset in order to ensure differently distributed data across the islands. Let $D$ denote the overall dataset. For each island, $(2/3)|D|$ data points are sampled from $D$ to construct the local training set ${{D}_{i}}$, while the remaining $(1/3)|D|$ data points are designated as the validation dataset ${{V}_{i}}$ for that island. Once the data sampling is completed, each island trains its own surrogate model ${{M}_{i}}$ utilizing its respective training set ${{D}_{i}}$. 

This work adopts the radial basis function network (RBFN) as the surrogate model for each island. Data with different distributions can be used to train diverse RBFNs. In this case, the number of RBFNs is equal to the number of islands $T$, with each island corresponding to an surrogate model. Consequently, we obtain the models $({{M}_{1}},{{M}_{2}},\ldots ,{{M}_{i}},\ldots ,{{M}_{T}})$. Subsequently, following the prediction of the validation dataset ${{V}_{i}}$ by the surrogate model ${{M}_{i}}$ of the $i$-th island, we can calculate the root mean square error ($RMSE$) between the predicted values and the true values. This allows us to assess the performance of the surrogate model for the $i$-th island.

\subsection{Semi-supervised Model Fine-tuning} \label{sec:3.3}

In order to automatically control and provide feedback to the surrogate model using different distribution information of the population during the population optimization process, and enable the surrogate model to evolve to a suitable fitness landscape, the surrogate model needs to be updated. To achieve this, after data and model initialization, we employ semi-supervised learning to fine-tune the model ${M_i}$ of the $i$-th island. 

Generally, in DSKT-DDEA, the update of each island's surrogate model involves cooperation with its neighboring islands. 
A predefined topology is used to determine $\Omega$ neighbors for each island.
For instance, for the Von Neumann topology we adopted, each island will have $\Omega = 4$ neighbors. For island $i$, we select $l$ individuals for pseudo label annotation, based on the discrepancy of the fitness approximations among its $\Omega$ neighbors. 
Subsequently, these  $l$ individuals, along with their pseudo labels, are combined with the original training data ${D_i}$ of island $i$ to collectively train the surrogate model ${M_i}$.

The detailed process is described as follows. First, individuals on the $i$-th island are represented as $({{x}_{i1}},{{x}_{i2}},\cdots ,{{x}_{ij}},\cdots ,{{x}_{in}})$. We can also assess the approximated fitness values of these individuals that are predicted by the $\Omega$ neighbor models of island $i$, denoted as $\left(f^1(x_{ij}) ,f^2( x_{ij}) ,\cdots ,f^{\Omega}(x_{ij}) \right)$. 
The mean fitness value predicted by the $\Omega$ neighbor models is denoted as $\overline{f_{ij}}$:

\begin{equation}
\label{eq:fij}
  \overline{{{f}_{ij}}}=\frac{1}{\Omega }\sum\limits_{k=1}^{\Omega }{{{f}^{k}{(x_{ij})}}}
\end{equation}
where $k$ represents the $k$-th neighbor of island $i$. Then, we can obtain the discrepancy of evaluation for the $j$-th individual on island $i$, denoted as $Div_{ij}$, as

\begin{equation}
\label{eq:div}
Di{{v}_{ij}}=\sqrt{\frac{1}{\Omega }\sum\limits_{k=1}^{\Omega }{{{({{f}^{k}{(x_{ij})}}-\overline{{{f}_{ij}}})}^{2}}}}
\end{equation}

Next, we sort the $n$ individuals on island $i$ based on their discrepancy indices  from smallest to largest. The top $l$ individuals with the smallest discrepancy are used to form new temporary data on island $i$. The $\ell$-th new temporary data on island $i$ is ${{x}_{i\ell }}$, $\ell \in [1, \cdots, l]$, while its  pseudo-label is calculated as 

\begin{equation}
\label{eq:p_label}
{{y}_{i\ell }}=\sum\limits_{k=1}^{\Omega }{{{w}^{k}}{{f}^k{(x_{i\ell}) }}}
\end{equation}
Here, ${{w}^{k}}$ represents the weight of the surrogate model for the $k$-th neighboring island, which is defined as 

\begin{equation}
\label{eq:w}
{{w}^{k}}=\frac{\sum\limits_{q=1}^{\Omega }{{{RMSE}^{q}}}-{{RMSE}^{k}}}{(\Omega -1)\sum\limits_{q=1}^{\Omega }{{{RMSE}^{q}}}}
\end{equation}

The $RMSE$ mentioned here is calculated using the validation data specific to each island. So far, on island $i$, we have obtained $l$ new data points with their pseudo labels, denoted as a tuple: $\left\langle {{X}_{i}},{{Y}_{i}} \right\rangle$.
Finally, the data on island $i$ is updated as
\begin{equation}
\label{eq:update_d}
{{D}_{i}}={{D}_{i}}\cup \left\langle {{X}_{i}},{{Y}_{i}} \right\rangle
\end{equation}

Then, island $i$ fine-tunes the surrogate model with the updated ${{D}_{i}}$. Since the distribution of individuals within each island undergoes changes during the local optimization process, the temporary training data is discarded after each training cycle. It should be noted that before each round of iteration, each island updates its own surrogate model. This iterative process allows the surrogate models of each island to continuously adapt and improve their performance based on the evolving data and local fitness landscape.

~\autoref{algo:0} details the semi-supervised fine-tuning process for island $i$ before generating offspring each time.

\begin{footnotesize}

\begin{algorithm} 
    \floatname{algorithm}{Algorithm}
    \setcounter{algorithm}{0}
    \caption{: Semi-supervised Model Fine-tuning}
    \begin{flushleft}
        \hspace*{0.02in} {\bf Input:}
        Training data ${{D}_{i}}$,
        Population ${{P}_{i}}$, 
        Surrogate set ${{M}_{set}}$, 
        RMSE set ${{E}_{set}}$, 
        Population size $n$.
        \\
        \hspace*{0.02in} {\bf Output:} ${{M}_{i}}$.
    \end{flushleft}
    \begin{algorithmic}[1]
        \State
        Geting local model ${{M}_{i}}$ from ${{M}_{set}}$
        \State
        Geting neighbor Models $N{{M}_{i}}=\{{{M}_{i}}^{1},{{M}_{i}}^{2},\cdots ,{{M}_{i}}^{\Omega}\}$ from ${{M}_{set}}$
        \State
        $\delta_{n\times\Omega}$ = Initializing an empty matrix with $n$ rows and $\Omega$ columns.
        \For{$j$ =0 to $n$}
            \State
            ${x}_{ij}$ $\leftarrow$ $j$-th population of the $i$-th island 
            \State
            $\left(f^1(x_{ij}) ,f^2( x_{ij}) ,\cdots ,f^{\Omega}(x_{ij}) \right)$ $\leftarrow$ Using $\Omega$ neighbor models to predict ${x}_{ij}$
            \State
            $\delta_{j, :}$ $\leftarrow$ $\left(f^1(x_{ij}) ,f^2( x_{ij}) ,\cdots ,f^{\Omega}(x_{ij}) \right)$
        \EndFor
        \State
        mean = []
        \State
        vars = []
        \For{$j$ =0 to $n$}
            \State
            $\overline{f_{ij}}$ = Calculate the mean of the \(j\)-th row of \(\delta_{n \times \Omega}\). (\autoref{eq:fij})
            \State
            $Div_{ij}$ = Calculate the standard deviation of the \(j\)-th row of \(\delta_{n \times \Omega}\). (\autoref{eq:div})
            \State
            mean.append($\overline{f_{ij}}$)
            \State
            vars.append($Div_{ij}$)
        \EndFor
        \State
        inds = Find the indices of the smallest \(l\) elements in the vars.
        \State
        ${X}_{i}$ = Extract \(l\) individuals according to the inds. 
        \State
        ${Y}_{i}$ = Calculate the label for these \(l\) individuals according to \autoref{eq:p_label}.
        \State
        ${{D}_{i}}={{D}_{i}}\cup \left\langle {{X}_{i}},{{Y}_{i}} \right\rangle$
        \State
        Using ${{D}_{i}}$ to fine-tune surrogate ${{M}_{i}}$
    \end{algorithmic}
    \begin{flushleft}
    \hspace*{0.02in} {\bf Output:} ${{M}_{i}}$.
    \end{flushleft}
        \label{algo:0}

\end{algorithm}

\end{footnotesize} 

\subsection{Intra-island Optimization} \label{sec:3.4}
After updating the local surrogate model as \autoref{sec:3.3}, the population undergoes the process of crossover and mutation. During environmental selection, we select individuals using ensembled local surrogates of the current island and its neighboring islands. This approach is adopted to mitigate the low accuracy of the surrogate models, particularly in high-dimensional spaces. The detailed process is described as follows.

For the $j$-th individual on island $i$, the fitness values predicted by the $\Omega$ neighboring models and the local surrogate model are denoted as a total of $\Omega +1$ surrogate models: $(f^1(x_{ij})$, $f^2(x_{ij})$, $\cdots $, $f^\Omega(x_{ij})$, $f^{(\Omega+1)}(x_{ij}))$. Then, the weighted fitness value of the $j$-th individual on island $i$ is given by
\begin{equation}
\label{eq:f_select}
\widehat{f_{ij}}=\sum\limits_{k=1}^{\Omega +1}{{{w}^{k}}f^{k}(x_{ij})}
\end{equation}
where ${w}^k$ represents the weight of the $k$-th surrogate model. 
The weight calculation follows the same formulation as \autoref{eq:w}. The individuals are then sorted in ascending order according to their fitness values, and the top $n$ individuals are selected as the next generation population. 
\begin{footnotesize}
\begin{algorithm}
    \floatname{algorithm}{Algorithm}
    \setcounter{algorithm}{1}
    \caption{: Intra-island Optimization}
    \begin{flushleft}
        \hspace*{0.02in} {\bf Input:}
        Training data ${{D}_{i}}$,
        Validation data ${{V}_{i}}$,
        Number of iterations before migration ${{t}_{iter}}$,
        Population ${{P}_{i}}$, 
        Surrogate set ${{M}_{set}}$, 
        RMSE set ${{E}_{set}}$, 
        Population size $n$.
        \\
        \hspace*{0.02in} {\bf Output:} ${{P}_{i}}$, $eli{{t}_{i}}$.
    \end{flushleft}
    \begin{algorithmic}[1]
        \While {${{Opt}_{iter}}\leq{{t}_{iter}}$ }
            \State
            {Use} \autoref{algo:0} to fine-tune surrogate ${{M}_{i}}$
            \State
            {Calculate} the latest $RMSE^{i}$ using the updated ${{M}_{i}}$ and ${{V}_{i}}$.
            \State
            {Update} ${{M}_{set}},{{E}_{set}}$ 
            \State
            {Share} ${{M}_{set}},{{E}_{set}}$ with other islands
            \State
            ${{D}_{i}}\leftarrow {{D}_{i}}\backslash \left\langle {{X}_{i}},{{Y}_{i}} \right\rangle $
            \State
            C $\leftarrow$ apply $SBX$ and $PM$ on ${{P}_{i}}$
            \State
            ${{P}_{i}}\leftarrow {{P}_{i}}\cup C$
            \State
            EM$\leftarrow$$N{{M}_{i}}\cup {{M}_{i}}$
            \State
            Calculate the weighted fitness of ${{P}_{i}}$ using EM and \autoref{eq:f_select}
            \State
            Sort ${{P}_{i}}$ base on the weighted fitness
            \State
            ${{P}_{i}}$ $\leftarrow$ Select top $n$ population
            \State
            $eli{{t}_{i}}$ $\leftarrow$ Select top 1 population
            \State
            ${{Opt}_{iter}}$ ++
        \EndWhile
    \end{algorithmic}
    \begin{flushleft}
    \hspace*{0.02in} {\bf Output:} ${{P}_{i}}$, $eli{{t}_{i}}$.
    \end{flushleft}
        \label{algo:1}
\end{algorithm}

\end{footnotesize}

The pseudocode for the intra-island optimization within an island is shown in \autoref{algo:1}. Each local subpopulation optimizes independently on each island, and the islands can be executed in parallel. Before starting the local optimization, data preprocessing is required, including sampling the population, training data, and validation data, as well as training the initial surrogate model for each island. 
Then, during the iterations, Algorithm line 2 {updates the local} surrogate model in a semi-supervised manner. Lines 3-5 are for updating the global surrogate model set and the RMSE set for other islands to use.
Optimization continues from line 6 of the algorithm {until Line} 14, where most evolutionary algorithms can serve as optimizers within each island. In this study, we use 
simulated binary crossover (SBX) \cite{SBX} and polynomial mutation (PM) \cite{PM1, PM2} as the methods for generating offspring. For evaluation, $EM$ refers to a collection of neighboring island surrogate models and local island surrogate models, and their weighted-average predictions are used to obtain the fitness to select individuals to the next generation.
Before conducting knowledge transfer, multiple island-based optimizations are performed to obtain the latest subpopulation ${{P}_{i}}$ and elite individual $eli{{t}_{i}}$ of the $i$-th island.

\subsection{Inter-island Knowledge Transfer} \label{sec:3.5}

To avoid the premature convergence of local islands and strengthen the overall global optimization effectiveness of the entire algorithm, inter-island communication is required. We design an adaptive knowledge transfer mechanism that facilitates the migration of individuals among neighboring islands and adjusts the migration probability based on their effectiveness.

\begin{figure}[h]
  \centering
  \includegraphics[width=0.4\linewidth]{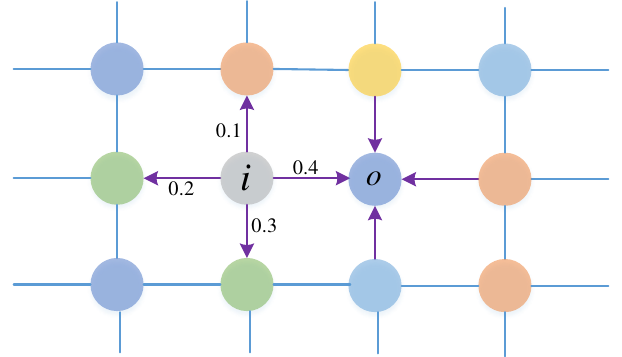}
  \caption{Example of source island and target island. Source island "$i$" can only migrate to the neighboring island in Von Neumann topology, and similarly, target island "$o$" can only receive immigrants from the neighboring islands in Von Neumann. The sum of migration probability of source island emitting is 1.}
  \label{fig:3}
\end{figure}

Specifically,
as shown in \autoref{fig:3}, taking the migration from source island $i$ to target island $o$ as an example, the migrated individuals from the source island will migrate only to $\Omega$ neighboring islands, and the sum of migration probability for all $\Omega$ edges should be 1. Inversely, the target island $o$ can only receive immigrants from the neighboring islands. 

Generally, the calculation of the migration probability from the
source island to the target island is as follows:
\begin{equation}
M{{P}_{io}}=\frac{{{\tau }_{io}}\times {{v}_{io}}}{\sum\limits_{k=1}^{\Omega }{{{\tau }_{ik}}\times {{v}_{ik}}}}
\label{eq:mp}
\end{equation}
In the equation, ${\tau }_{io}$ represents the \textit{attractiveness} of source island $i$ to target island $o$, and ${v}_{io}$ represents the \textit{differentiation factor} of source island $i$ to target island $o$.  
Next, we formulate the attractiveness and the differentiation factor in detail.

\subsubsection{Attractiveness Calculation}

The attractiveness of source island $i$ to target island $o$ is defined as 
\begin{equation}
    \tau_{io} = (1-\rho) \tau_{io}^{\rm{pre}} + \Delta \tau_{io}
\label{eq:a}
\end{equation}
where $\rho \in \left( 0,1 \right)$ represents the decay factor, $\tau_{io}^{\rm{pre}}$ represents the accumulated attractiveness up until the current migration (this design strives to retain historical attractiveness), and $\Delta \tau_{io}$ represents the attractive part that needs to be added during the current migration. 

The importance of $\Delta \tau_{io}$ is evident. We believe that the attractiveness of target island $o$ towards source island $i$ is related to two attraction factors concerning the current migration effectiveness. The first factor is the average fitness improvement of all individuals on target island $o$ caused by this migration, denoted as
\begin{equation}
{{\theta }_{o}}={{\overline{f}}_{o}}-\overline{{{f}_{o}}^{*}}
\label{eq:theta}
\end{equation}
where ${\overline{f}}_{o}$ represents the average fitness of the population on the target island $o$ before migration. $\overline{{{f}_{o}}^{*}}$ represents the average fitness of the population on the island $o$ after migration and another $t_{iter}$ rounds of optimization. The fitness here is estimated by the surrogate model on island $o$. Please note that both the population and the surrogate model are constantly changing. When calculating $\overline{{{f}_{o}}^{*}}$, we suggest to go through the migration and multiple rounds of optimization, with the aim of making not only the immigrants useful for  island $o$, but also their descendants.
Finally,  $\theta _o$ is normalized into range $[0, 1]$ by min-max normalization. 

Given that the above attraction factor is {calculated on} the population-level, the second attraction factor is {calculated on} a more fine-grained manner, considering the individual-level. 
Specifically, we first define the migration effectiveness $r_{io}^{m}$ of the $m$-th individual when it migrated from the island $i$ to the target island $o$ as
\begin{equation}
r_{io}^{m}=n-rank_{io}^{m}
\end{equation}
where $n$ represents the population size on the target island $o$, $rank_{io}^{m}$ represents the rank of the immigrant in the fitness ranking of the target island. 
For example, suppose there are 100 individuals on the target island $o$, and one individual migrates from island $i$ to join the local population. Then, the fitness of these 101 individuals is predicted and ranked using the surrogate model of the target island. If the rank of this new immigrant is 3, it indicates that the immigrant is highly important for the target island. Therefore, $r_{io}^{m} = 100-3 = 97$.

Note that in the migration, there is not only one individual migrating from the source island $i$ to the target island $o$, and there are also multiple source islands involved. Our second attraction factor is calculated by considering the effectiveness ranking of all immigrants from the source island $i$ to the target island $o$, which is defined as
\begin{equation}
\varphi _{io}=\frac{\sum_{m=1}^{\beta _i}{r_{io}^{m}}}{\sum_{i=1}^{\Omega}{\sum_{m=1}^{\beta _i}{r_{io}^{m}}}}
\end{equation}
where 
$\beta_i$ refers to the total number of immigrants which migrated from the source island $i$ to the target island $o$. The denominator in the above equation denotes the rank sum for all immigrates received by the target island $o$ in the current migration process. 

Finally, the calculation of $\Delta {{\tau }_{io}}$ is given as
\begin{equation}
\Delta {{\tau }_{io}}={{\theta }_{o}} \times {{\varphi }_{io}} 
\end{equation}

\subsubsection{Differential Factor}

In computing the attractiveness, the effectiveness of migration is calculated based on a surrogate model on the target island solely. As a supplement, when calculating the migration probability, we also employ a differential factor that considers the approximation difference of individuals by surrogates  of the source and target islands. The larger the difference, the better, which helps to improve the diversity of immigrants.

The difference factor $v_{io}$ between source island $i$ and target island $o$ is defined as 

\begin{equation}
v_{io} = \frac{1}{n}\sum_{j=1}^{n}\left|f^o(x_{ij}) - f^i(x_{ij})\right|
\label{eq:v}
\end{equation}
where $x_{ij}$ represents the $j$-th individual on island $i$, $f^o(\cdot)$ represents the fitness prediction for an individual using the surrogate model on the target island $o$, and $f^i(\cdot)$ represents the fitness prediction using the surrogate model on the source island $i$. We further normalize the $v_{io}$ into the range $[0,1]$ by min-max normalization.

Above, we have discussed in detail the calculation process for the migration probability from a source island $i$ to a target island $o$. 
During the migration phase, the source island $i$ randomly selects 10\% of individuals to be migrated. For each individual to be migrated, its target island is determined using a roulette wheel selection method according to the migration probability.
After each migration event, our algorithm proceeds through subsequent evolutionary stages, including an environmental selection step. Regardless of the number of individuals present post-migration, the population on each island is restricted by selecting only the top $n$ individuals based on their ``fitness'' levels (evaluated using the integrated surrogate model), ensuring that only the most promising individuals are retained for subsequent generations.
\begin{footnotesize}

\begin{algorithm}
    \floatname{algorithm}{Algorithm}
    \setcounter{algorithm}{2}
    \caption{: DSKT-DDEA}\label{algo:2}
    \begin{flushleft}
        \hspace*{0.02in} {\bf Input:}
        Training data $D$,
        Maximum generation $Ma{{x}_{iter}}$,
        Migration gap ${{t}_{iter}}$,
        Population size $n$,
        Island count $T$.
        \\
        \hspace*{0.02in} {\bf Output:} $elit[-1]$.
    \end{flushleft}
    \begin{algorithmic}[1]
        \State
        ${{D}_{set}},{{V}_{set}},{{P}_{set}},{{M}_{set}},{{E}_{set}} \leftarrow \varnothing$ 
        \For{$i$ =0 to $T$}
            \State
            ${{D}_{i}}$ $\leftarrow$ Sample a training data set for the $i$-th island from $D$
            \State
            ${{V}_{i}}$ $\leftarrow$ $D$ $\backslash$ ${{D}_{i}}$
            \State
            ${{P}_{i}}$ $\leftarrow$ Using LHS to sample population
            \State
            Training surrogate ${{M}_{i}}$
            \State
            ${{RMSE}^{i}}$ $\leftarrow$ Calculate the RMSE
            \State
            Storing ${{D}_{i}},{{S}_{i}},{{P}_{i}},{{M}_{i}},{{RMSE}^{i}}$ into ${{D}_{set}},{{V}_{set}},{{P}_{set}},{{M}_{set}},{{E}_{set}}$
        \EndFor
        \State
        Alloc a worker to each island
        \State
        All islands share ${{D}_{set}},{{V}_{set}},{{P}_{set}},{{M}_{set}},{{E}_{set}}$
        \State
        Compute migration probability based on \autoref{eq:mp} for each island.
        \State
        $elit=\varnothing $
        \While {$iter\leq Ma{{x}_{iter}}$}
            \State
            $eli{{t}_{pool}}=\varnothing$
            \For{$i$ =0 to $T$}
                /*This loop is executed in parallel across multiple workers.*/
                \State
                ${{P}_{i}}$, $eli{{t}_{i}}$ $\leftarrow$ $i$-th island performs optimization based on  \autoref{algo:1}
                \State
                Using ${{P}_{i}}$ to update ${{P}_{set}}$
                \State
                $eli{{t}_{pool}}=eli{{t}_{pool}}\cup eli{{t}_{i}}$
                \State
                $iter$ = $iter$ + ${{t}_{iter}}$
            \EndFor
            \State
            $eli{{t}_{g}}$ $\leftarrow$ Select a global elite from $eli{{t}_{pool}}$ using \autoref{eq:gf}
            \State
            $elit=elit\cup \{eli{{t}_{g}}\}$
            \State
            $idx$ = Get the index of the minimum value of ${F}^{g}(elit)$ using \autoref{eq:gf}
            \If {($idx < |elit| - es$)} /*This is early stopping mechanism.*/
                \State
                Break
            \Else
                \State
    	        Each island migrates individuals to neighboring islands based on migration probabilities and Roulette Wheel
                \State
                Update the migration probabilities based on \autoref{eq:mp}
            \EndIf
        \EndWhile
    \end{algorithmic}
    \begin{flushleft}
    \hspace*{0.02in} {\bf Output:} $elit[-1]$.
    \end{flushleft}
\end{algorithm}

\end{footnotesize}
\subsection{Early Stop Mechanism and the Entire Workflow}
The last issue to manage is when to terminate the DSKT-DDEA process and output the final solution. Usually, the algorithm is assigned a maximum number of generations as a limited budget. Additionally, we propose a mechanism that may halt the algorithm early according to the changes in global fitness (estimated ones, since offline DDEAs do not allow real functions evaluations). This early stop mechanism enables the optimization process to save a significant amount of time and effectively avoid overfitting, thereby preventing search stagnation and degradation.

Before each migration, we integrate the models from all islands and use the global integrated model to predict fitness from elites of each island. This provides us with the {globally best seen} solution. Here, the weight of the $k$-th surrogate model, denoted as $w_k$, is calculated as

\begin{equation}
w^k=\frac{\sum_{q=1}^{T} RMSE^q-RMSE^k}{(T-1) \sum_{q=1}^{T} RMSE^q}
\end{equation}
which is similar to \autoref{eq:w}, but we consider all islands, namely $q\in [1,2,\cdots ,T]$.

The global surrogate model $F^g(\cdot)$ is then defined as the weighted sum of all islands' surrogate models:
\begin{equation}
F^g(\cdot)=\sum\limits_{k=1}^{T}{w^k f^k(\cdot)}
\label{eq:gf}
\end{equation}

According to \autoref{algo:1}, we obtain the elites of all islands as $elit=\{elit_1,elit_2,\cdots ,elit_T\}$. We calculate the fitness values of all island elites using the global surrogate model ${F}^{g}(\cdot)$, and the smallest fitness value is denoted as $fit_{\min}$. If $fit_{\min}$ does not decrease for $es$ consecutive migration stages, the entire optimization process is stopped.

\autoref{algo:2} presents the overall pseudocode of DSKT-DDEA. In lines 1-9, the training set, validation set, population, surrogate model, and RMSE are initialized and stored in sets $D_{set}$, $V_{set}$, $P_{set}$, $M_{set}$, and $E_{set}$, respectively, identified by the island number. Since DSKT-DDEA is a parallel optimization method, line 10 assigns a worker to each island to enable parallelism using methods such as process-based or multi-server parallelism. 
During the main loop, lines 16 to 21 perform intra-island optimization in parallel among multiple workers. The updated latest population of each optimized island is promptly updated into the population collection.
Lines 22-26 check if the global best solution is no longer improving its global fitness. If it is not improving, it is recommended to stop the entire process in order to save optimization time. Otherwise, the inter-island migration is adaptively conducted in lines 28-29. 
Notably, the restart mechanism is often employed to enhance the performance of EAs. However, for the sake of simplicity, our algorithm currently does not incorporate this feature.

\section{Experimental results}\label{sec:experimental_analysis}
In this section, we first introduce the experimental setup and algorithm parameter settings. Then, we compare DSKT-DDEA with the state-of-the-art (SOTA) DDEAs, which will be discussed in detail in \autoref{subsec:DDEA}. 
Additionally, we conducted in-depth analysis on DSKT-DDEA, which is presented in \autoref{subsec:ablation}. Lastly, we investigate the parallel speedup of our algorithm in \autoref{subsec:para}.

\subsection{Experimental Setup}
We evaluated our algorithm using the following test suites: {(1) BBOB\footnote{Please note that the large-scale BBOB test suite provided by COCO (https://github.com/numbbo/coco) supports up to 640 dimensions, while the BBOB test suite provided by MetaBox (https://github.com/GMC-DRL/MetaBox) and pycma (https://github.com/CMA-ES/pycma) can support 1000 dimensions.}\cite{COCO,bbob-largescale, MetaBox, lq-CMA-ES}: The BBOB noiseless sub-testsuite contains 24 noiseless, single-objective test functions, each with 15 test instances. We selected the 0-th test instance from each of the 24 test functions in this sub-testsuite suite for evaluation.
The search range of the decision variables was set to [-5, 5], with a dimensionality of 1000. The test functions are denoted as `F1' to `F24'. (2) CEC2010~\cite{CEC2010}: The CEC2010 test suite includes 20 test functions, where we applied the official default search range with a dimensionality of 1000. The test functions are denoted as `F1' to `F20'.}


For the offline DDEAs, including our DSKT-DDEA, all the data was obtained through LHS prior to the start of the EA and used as training samples. No new data sampling was allowed during the optimization process. 

  The fitness of the population is estimated using the surrogate model. At the end of the algorithm, the best individual (estimated by the surrogate) is {output. For a fair comparison, the maximum available FEs count is set to 500, meaning that the global dataset has a size $|D|=500$. For the online DDEAs, The algorithm configurations adhere to the settings described in their original publications. They are also allowed $500$ real FEs, with a portion of them conducted during the algorithm iterations according to the algorithms' flowchart}. The particular division of offline and online sample size is indicated in \autoref{tab:init_data}.

\begin{table}
    \centering
    \caption{Division of Offline and Online Sample Sizes for Different Algorithms}
    \resizebox{0.5\linewidth}{!}{
    \begin{tabular}{|c|c|c|c|}
    \hline
    \textbf{Algorithms} & \textbf{Offline FEs} & \textbf{Online FEs}&\textbf{Total FEs}\\
    \hline
    lq-CMA-ES & 0 & 500 & 500\\
    \hline
    SADE-AMSS & 200 & 300& 500\\
    \hline
    CA-LLSO & 200 & 300& 500\\
    \hline
    SA-COSO & 230 & 270& 500\\
    \hline
    SAMSO & 200 & 300& 500\\
    \hline
    SAEA-RFS & 200 & 300& 500\\
    \hline
    ESCO & 200 & 300& 500\\
    \hline
    All offline DDEAs including ours & 500 &0& 500\\
    \hline
    \end{tabular}
    }
    \label{tab:init_data}
\end{table}
  To assess the performance of each DDEA, we employ an essential step of real fitness evaluation for the best individual produced by each algorithm. This approach ensures that the final optimization results are grounded in the original problem's context, providing a meaningful assessment of the algorithm's practical utility. 
Each problem underwent 20 independent runs, and the average and standard deviation of these 20 runs were provided. Additionally, we employed the Wilcoxon rank-sum test \cite{wilcoxon} to verify if there was a significant difference in performance between the two algorithms on specific problems. The significance parameter $\alpha$ was set to 0.05, while "+", "$\approx$", and "-" indicate that the proposed method outperforms, is comparable to, or is worse than the competing methods. For each problem, the results of the best-performing algorithm are displayed in bold. The testing environment for all experiments was an Intel(R) Xeon(R) CPU 2.30GHz, 18 cores, 128GB memory.

\begin{table}
  \caption{Algorithm Parameters}
      \label{tab:parameters}
  \resizebox{0.55\linewidth}{!}{
  \begin{tabular}{|c|c|l|}
    \hline
    Module&Parameter&Value\\
    \hline
\multirow{6}{*}{Global}
&Island count $T$ &36\\
&Population size $n$ &100\\
&Early stop $es$&3\\
&Migration gap ${{t}_{iter}}$&90\\
&Number of migrated individuals &10\\
&Max iter & 1800\\
&Attraction decay factor $\rho$& 0.1 \\
\hline
\multirow{2}{*}{Surrogate}
&Number of augmented data(Pseudo-label counts)&3\\
&Number of RBF centers&$\left\lceil \sqrt{(2/3){|D|}} 
\right\rceil $\\
\hline
\multirow{4}{*}{EA}
&Probability of crossover&1\\
&Distribution index ${{\eta}_{c}}$ in SBX&15\\
&Distribution index ${{\eta}_{m}}$ in PM&15\\
&Probability of mutation&1/$d$\\
  \hline
\end{tabular}
}
\end{table}

We list the DSKT-DDEA parameters in \autoref{tab:parameters}. In terms of global settings, We define the number of islands as 36\footnote{Since the target island of a certain island during migration is its von Neumann neighbor, in order to maintain an even distribution of neighbor numbering, we suggest that the number of rows and columns in the topology should be the same, and the number of islands should be a square. {So} the number of islands is 36, with a von Neumann topology of 6 rows and 6 columns.}. The population size of each island is 100. All islands are iterated in parallel for maximum 1800 times. The migration gap ${{t}_{iter}}$ is set to 90, which means migration occurs synchronously every 90 iterations, so the migration frequency is 20. As previously stated, the number of individuals migrated each time is 10\% of the population number randomly selected from each island, therefore the number of migrated individuals is 10. The early stopping parameter $es$ is set to 3, which means that if the global best fitness does not improve for 3 consecutive migrations, the entire process will stop, even if the maximum iteration limit has not been reached. Attraction decay factor $\rho$ is 0.1. 
The number of RBF centers is set to $\left\lceil \sqrt{(2/3)|D|} \right\rceil$\footnote{Since the data size of each island is $\left\lceil (2/3){|D|} \right\rceil$, the actual number of centers is the square root of the data size, according to past conventions.}. 
In the semi-supervised fine-tuning module, the number of augmented data is $l=3$.
In the EA module, we use SBX crossover and polynomial mutation as operators to generate new populations. The distribution index ${{\eta}_{c}}$ of SBX is set to 15, and the crossover probability is 100\%. For polynomial mutation PM, the distribution index ${{\eta}_{m}}$ is set to 15, and the mutation probability is 1/$d$, where $d$ is the dimension of the decision variables.

\subsection{Performance Comparisons}\label{subsec:DDEA}
\subsubsection{Comparing with Offline DDEAs}
\label{subsec:offDDEA}
We compare our algorithm with the following six offline DDEAs.

(1) CL-DDEA \cite{CL-DDEA}: This algorithm establishes a comparative learning model to determine the superiority and inferiority of individuals through binary classification. It also enhances the training data by predicting the pairing relationships between individuals, constructing a directed graph, and proposing a topological sorting algorithm on the graph to obtain the ranking of the population. This algorithm exhibits advantages in solving relatively high-dimensional problems ($d=200$).

(2) SRK-DDEA \cite{SRK-DDEA}: This algorithm introduces four RBFNs based on four different kernel functions as surrogate models. In order to more effectively manage approximation error and utilize initial data more efficiently, a combination strategy of Stochastic Ranking is introduced to rank individuals. It uses selection sort to determine the rank of the population. In the inner loop, only one RBFN is randomly selected to get the best individual among the remaining ones.

(3) DDEA-SE \cite{DDEA-SE}: This algorithm applies bootstrap to generate a large number of subsets from the offline data set and then independently constructs many RBFNs on these subsets. The fitness of individuals is estimated by combining these models.

(4) BDDEA-LDG \cite{BDDEA-LDG}: This algorithm proposes a local data generation  method to generate synthetic data and uses a boosting strategy similar to AdaBoost learning to construct multiple base learners on the enhanced training data.

(5) TT-DDEA \cite{TT-DDEA}: The semi-supervised learning method of tri-training is used to update the surrogate models in the proposed offline data-driven evolutionary algorithm. In the optimization process, each surrogate model is dynamically updated.

(6) MS-DDEO \cite{MS-DDEO}: The offline samples are divided into training and testing data, and a model pool is constructed via four RBFNs with different smoothness degrees. Each guides corresponding DDEAs to find the global optimal solution. 
The performance of RBFN on the test set, as well as the average distance between a solution and offline data, is used as the criterion for choosing the optimal solution.

{
The comparison results for the 1000-dimensional BBOB and CEC2010 benchmarks are shown in \autoref{tab:Offline} and \autoref{tab:compare off cec2010}, respectively. 
According to the comparison results presented in \autoref{tab:Offline}, DSKT-DDEA demonstrates exceptional performance with an average ranking of 2.54. The subsequent best performing models include CL-DDEA, MS-DDEO, SRK-DDEA, TT-DDEA, BDDEA-LDG, DDEA-SE. }

\begin{table}
\centering
\caption{Comparison with Offline DDEAs on 1000-Dimensional BBOB}
\resizebox{0.9\linewidth}{!}{
\arrayrulewidth=0.6pt 

\begin{tabular}{|c|c|c|c|c|c|c|c|}
\hline
Benchmark & DSKT-DDEA & CL-DDEA & SRK-DDEA & DDEA-SE & BDDEA-LDG & TT-DDEA & MS-DDEO \\\hline
F1 & 4.76E+03$\pm$1.75E+01 & 4.98E+03$\pm$6.31E+01(+) & 5.23E+03$\pm$4.87E+01(+) & 6.48E+03$\pm$1.37E+02(+) & 6.20E+03$\pm$1.32E+02(+) & 5.30E+03$\pm$1.23E+02(+) & \textbf{4.62E+03$\pm$1.60E+02($\approx$)} \\\hline
F2 & 3.04E+08$\pm$1.24E+07 & 3.86E+08$\pm$8.82E+06(+) & 3.87E+08$\pm$2.94E+07(+) & 4.96E+08$\pm$3.31E+07(+) & 4.42E+08$\pm$2.86E+07(+) & 4.03E+08$\pm$1.77E+07(+) & \textbf{2.51E+08$\pm$3.84E+07($\approx$)} \\\hline
F3 & \textbf{3.73E+04$\pm$4.50E+02} & 3.99E+04$\pm$5.82E+02(+) & 4.27E+04$\pm$1.77E+03(+) & 5.73E+04$\pm$2.38E+03(+) & 5.44E+04$\pm$2.88E+03(+) & 4.25E+04$\pm$1.74E+03(+) & 3.93E+04$\pm$1.62E+03($\approx$) \\\hline
F4 & \textbf{3.09E+04$\pm$1.46E+02} & 3.21E+04$\pm$4.42E+02(+) & 3.62E+04$\pm$1.64E+03(+) & 6.95E+04$\pm$1.12E+04(+) & 5.52E+04$\pm$7.03E+03(+) & 3.56E+04$\pm$1.20E+03(+) & 4.83E+04$\pm$2.73E+03(+) \\\hline
F5 & 1.63E+04$\pm$2.68E+02 & 1.92E+04$\pm$2.73E+02($\approx$) & 1.92E+04$\pm$1.11E+02($\approx$) & 1.73E+04$\pm$8.68E+02($\approx$) & 1.82E+04$\pm$4.61E+02($\approx$) & 1.94E+04$\pm$3.43E+02(+) & \textbf{1.51E+04$\pm$8.04E+02($\approx$)} \\\hline
F6 & 1.81E+07$\pm$6.38E+04 & 1.86E+07$\pm$4.84E+04(+) & 1.89E+07$\pm$5.72E+04(+) & 2.15E+07$\pm$4.46E+05(+) & 2.00E+07$\pm$6.46E+05(+) & 1.90E+07$\pm$1.78E+05(+) & \textbf{1.45E+07$\pm$1.39E+06(-)} \\\hline
F7 & 4.67E+04$\pm$1.03E+03 & 5.04E+04$\pm$6.07E+02(+) & 5.09E+04$\pm$1.47E+03(+) & 6.46E+04$\pm$2.74E+03(+) & 6.07E+04$\pm$1.99E+03(+) & 5.35E+04$\pm$1.22E+03(+) & \textbf{4.13E+04$\pm$3.58E+03($\approx$)} \\\hline
F8 & \textbf{4.11E+08$\pm$4.51E+06} & 5.09E+08$\pm$1.16E+07(+) & 6.40E+08$\pm$4.17E+07(+) & 1.62E+09$\pm$1.05E+08(+) & 1.23E+09$\pm$9.00E+07(+) & 6.26E+08$\pm$3.85E+07(+) & 8.32E+08$\pm$9.34E+07(+) \\\hline
F9 & \textbf{1.24E+04$\pm$1.30E+03} & 1.47E+06$\pm$2.53E+05(+) & 1.50E+07$\pm$1.84E+06(+) & 2.47E+08$\pm$2.82E+07(+) & 1.61E+08$\pm$3.65E+07(+) & 1.33E+07$\pm$3.13E+06(+) & 1.55E+08$\pm$3.30E+07(+) \\\hline
F10 & \textbf{2.75E+08$\pm$7.89E+06} & 3.26E+08$\pm$1.29E+07(+) & 3.60E+08$\pm$1.75E+07(+) & 4.25E+08$\pm$3.08E+07(+) & 4.26E+08$\pm$2.86E+07(+) & 3.50E+08$\pm$1.71E+07(+) & 2.97E+08$\pm$2.09E+07($\approx$) \\\hline
F11 & 6.22E+06$\pm$2.99E+06 & 2.27E+06$\pm$3.27E+06($\approx$) & 1.84E+06$\pm$2.83E+06($\approx$) & 8.57E+04$\pm$1.33E+05(-) & \textbf{3.51E+04$\pm$5.28E+04(-)} & 4.49E+06$\pm$4.67E+06($\approx$) & 2.46E+05$\pm$2.07E+05(-) \\\hline
F12 & 1.29E+11$\pm$1.59E+11 & \textbf{1.78E+10$\pm$7.50E+08($\approx$)} & 2.20E+10$\pm$3.25E+09($\approx$) & 3.35E+10$\pm$9.50E+09($\approx$) & 2.96E+10$\pm$4.78E+09($\approx$) & 2.09E+10$\pm$1.58E+09($\approx$) & 2.92E+10$\pm$2.67E+09($\approx$) \\\hline
F13 & 1.47E+04$\pm$4.40E+01 & 1.50E+04$\pm$4.52E+01(+) & 1.53E+04$\pm$1.14E+02(+) & 1.70E+04$\pm$2.96E+02(+) & 1.65E+04$\pm$3.18E+02(+) & 1.54E+04$\pm$9.30E+01(+) & \textbf{1.41E+04$\pm$4.73E+02($\approx$)} \\\hline
F14 & \textbf{3.57E+02$\pm$1.97E+01} & 4.25E+02$\pm$1.46E+01(+) & 4.80E+02$\pm$2.89E+01(+) & 7.09E+02$\pm$8.07E+01(+) & 5.92E+02$\pm$5.86E+01(+) & 4.87E+02$\pm$1.91E+01(+) & 5.70E+02$\pm$1.66E+02(+) \\\hline
F15 & \textbf{3.67E+04$\pm$3.15E+02} & 3.81E+04$\pm$4.89E+02(+) & 4.04E+04$\pm$7.18E+02(+) & 4.95E+04$\pm$2.12E+03(+) & 4.78E+04$\pm$3.00E+03(+) & 4.02E+04$\pm$1.28E+03(+) & 4.51E+04$\pm$1.88E+03(+) \\\hline
F16 & -1.79E+02$\pm$4.11E+00 & -1.80E+02$\pm$1.95E+00($\approx$) & -1.80E+02$\pm$2.89E+00($\approx$) & \textbf{-1.88E+02$\pm$1.80E+00(-)} & -1.87E+02$\pm$1.05E+00(-) & -1.79E+02$\pm$2.24E+00($\approx$) & -1.83E+02$\pm$4.24E+00($\approx$) \\\hline
F17 & -1.17E+01$\pm$7.31E+00 & \textbf{-1.43E+01$\pm$1.50E+00($\approx$)} & -1.16E+01$\pm$3.95E+00($\approx$) & -5.30E+00$\pm$2.01E+00($\approx$) & -1.22E+01$\pm$2.70E+00($\approx$) & -1.21E+01$\pm$1.89E+00($\approx$) & -1.16E+01$\pm$2.23E+00($\approx$) \\\hline
F18 & 6.64E+01$\pm$3.10E+01 & 1.01E+02$\pm$9.39E+01($\approx$) & 6.53E+01$\pm$8.88E+00($\approx$) & 8.89E+01$\pm$1.37E+01($\approx$) & 6.52E+01$\pm$8.98E+00($\approx$) & 7.65E+01$\pm$4.24E+01($\approx$) & \textbf{6.07E+01$\pm$1.81E+00($\approx$)} \\\hline
F19 & \textbf{4.96E+01$\pm$1.04E-01} & 5.45E+01$\pm$3.81E-01(+) & 8.39E+01$\pm$3.86E+00(+) & 7.33E+02$\pm$1.75E+02(+) & 4.89E+02$\pm$1.28E+02(+) & 8.67E+01$\pm$8.81E+00(+) & 4.24E+02$\pm$9.24E+01(+) \\\hline
F20 & \textbf{6.50E+05$\pm$3.88E+04} & 9.56E+05$\pm$2.08E+04(+) & 1.25E+06$\pm$6.62E+04(+) & 3.10E+06$\pm$3.12E+05(+) & 2.24E+06$\pm$1.09E+05(+) & 1.26E+06$\pm$1.03E+05(+) & 1.16E+06$\pm$1.06E+05(+) \\\hline
F21 & \textbf{3.95E+02$\pm$1.15E-02} & 3.95E+02$\pm$7.63E-02(+) & 3.95E+02$\pm$1.07E-01(+) & 3.96E+02$\pm$1.74E-01(+) & 3.96E+02$\pm$8.00E-02(+) & 3.95E+02$\pm$8.16E-02(+) & 3.95E+02$\pm$1.87E-01(+) \\\hline
F22 & 1.26E+02$\pm$2.99E-02 & 1.27E+02$\pm$9.63E-02(+) & 1.27E+02$\pm$1.03E-01(+) & 1.28E+02$\pm$1.97E-01(+) & 1.28E+02$\pm$6.51E-02(+) & 1.27E+02$\pm$1.02E-01(+) & \textbf{1.25E+02$\pm$9.90E-01($\approx$)} \\\hline
F23 & 1.50E+03$\pm$1.87E-02 & 1.50E+03$\pm$1.92E-02(-) & 1.50E+03$\pm$3.06E-02($\approx$) & 1.50E+03$\pm$9.41E-03(-) & 1.50E+03$\pm$1.94E-02(-) & 1.50E+03$\pm$3.88E-02($\approx$) & \textbf{1.50E+03$\pm$1.94E-02(-)} \\\hline
F24 & \textbf{1.62E+04$\pm$7.73E+01} & 1.68E+04$\pm$1.49E+02(+) & 1.79E+04$\pm$1.96E+02(+) & 2.35E+04$\pm$8.21E+02(+) & 2.19E+04$\pm$3.70E+02(+) & 1.78E+04$\pm$1.88E+02(+) & 2.07E+04$\pm$2.00E+02(+) \\\hline
+/$\approx$/- & NA & 17/6/1 & 17/7/0 & 17/4/3 & 17/4/3 & 18/6/0 & 9/12/3 \\\hline
Average Rank & 2.54 & 2.96 & 4.12 & 6.04 & 5.08 & 4.29 & 2.96 \\\hline
\end{tabular}
}
\label{tab:Offline}
\end{table}

\begin{table}
    \centering
    \caption{
    Comparison with Offline DDEAs on the 1000-Dimensional CEC2010 Benchmark Functions
    }
    \resizebox{0.9\linewidth}{!}{
\arrayrulewidth=0.6pt 
\begin{tabular}{|c|c|c|c|c|c|c|c|}
\hline
Benchmark & DSKT-DDEA & CL-DDEA & SRK-DDEA & DDEA-SE & BDDEA-LDG & TT-DDEA & MS-DDEO \\\hline
F1 & 1.61E+11$\pm$8.47E+09 & 1.94E+11$\pm$6.81E+09(+) & 2.01E+11$\pm$1.13E+10(+) & 2.45E+11$\pm$1.55E+10(+) & 2.34E+11$\pm$1.69E+10(+) & 2.06E+11$\pm$6.00E+09(+) & \textbf{1.14E+11$\pm$6.18E+09(-)} \\\hline
F2 & 1.70E+04$\pm$1.16E+02 & 1.74E+04$\pm$1.79E+02(+) & 1.74E+04$\pm$2.28E+02(+) & 1.91E+04$\pm$3.58E+02(+) & 1.85E+04$\pm$1.15E+02(+) & 1.75E+04$\pm$1.94E+02(+) & \textbf{1.68E+04$\pm$1.44E+02($\approx$)} \\\hline
F3 & \textbf{2.10E+01$\pm$2.69E-02} & 2.11E+01$\pm$1.77E-02(+) & 2.11E+01$\pm$3.64E-02(+) & 2.13E+01$\pm$3.24E-02(+) & 2.12E+01$\pm$4.44E-02(+) & 2.11E+01$\pm$1.96E-02(+) & 2.12E+01$\pm$5.78E-02(+) \\\hline
F4 & 2.08E+16$\pm$1.64E+16 & 7.41E+15$\pm$4.41E+15($\approx$) & 1.61E+16$\pm$2.03E+16($\approx$) & 1.42E+16$\pm$6.95E+15($\approx$) & \textbf{6.89E+15$\pm$1.07E+15($\approx$)} & 1.06E+16$\pm$1.54E+16($\approx$) & 2.64E+16$\pm$2.04E+15($\approx$) \\\hline
F5 & 7.59E+08$\pm$8.02E+07 & 9.29E+08$\pm$8.66E+07(+) & 9.14E+08$\pm$6.70E+07(+) & 9.47E+08$\pm$7.34E+07(+) & 8.92E+08$\pm$9.56E+07($\approx$) & 9.44E+08$\pm$1.06E+08(+) & \textbf{6.08E+08$\pm$4.27E+07($\approx$)} \\\hline
F6 & 2.06E+07$\pm$4.13E+05 & 2.09E+07$\pm$1.09E+05($\approx$) & 2.11E+07$\pm$3.42E+05($\approx$) & 2.14E+07$\pm$9.01E+04($\approx$) & 2.11E+07$\pm$2.85E+05($\approx$) & 2.11E+07$\pm$3.16E+05(+) & \textbf{2.02E+07$\pm$5.29E+05($\approx$)} \\\hline
F7 & 2.16E+14$\pm$6.87E+13 & \textbf{9.89E+12$\pm$6.06E+12(-)} & 6.96E+13$\pm$1.02E+14(-) & 2.08E+13$\pm$2.21E+13(-) & 1.43E+13$\pm$6.14E+12(-) & 5.23E+13$\pm$7.96E+13(-) & 1.06E+14$\pm$8.56E+12(-) \\\hline
F8 & 2.44E+16$\pm$4.20E+15 & 8.17E+16$\pm$7.58E+16(+) & 5.60E+16$\pm$4.35E+15(+) & 8.77E+16$\pm$4.47E+16(+) & 9.07E+16$\pm$7.62E+15(+) & 6.69E+16$\pm$2.11E+16(+) & \textbf{1.71E+16$\pm$1.09E+16($\approx$)} \\\hline
F9 & 2.02E+11$\pm$8.68E+09 & 2.33E+11$\pm$6.09E+09(+) & 2.42E+11$\pm$1.18E+10(+) & 2.76E+11$\pm$1.85E+10(+) & 2.83E+11$\pm$1.33E+10(+) & 2.44E+11$\pm$9.34E+09(+) & \textbf{1.67E+11$\pm$2.79E+10($\approx$)} \\\hline
F10 & 1.74E+04$\pm$1.51E+02 & 1.75E+04$\pm$1.70E+02($\approx$) & 1.77E+04$\pm$1.55E+02($\approx$) & 1.93E+04$\pm$5.97E+02(+) & 1.86E+04$\pm$3.96E+02(+) & 1.78E+04$\pm$1.93E+02(+) & \textbf{1.71E+04$\pm$3.37E+02($\approx$)} \\\hline
F11 & \textbf{2.31E+02$\pm$2.76E-01} & 2.31E+02$\pm$4.35E-01(+) & 2.32E+02$\pm$4.63E-01(+) & 2.34E+02$\pm$4.08E-01(+) & 2.34E+02$\pm$5.17E-01(+) & 2.32E+02$\pm$4.79E-01(+) & 2.35E+02$\pm$4.24E-01(+) \\\hline
F12 & 2.70E+07$\pm$4.74E+07 & \textbf{4.47E+06$\pm$1.21E+06($\approx$)} & 1.89E+07$\pm$1.24E+07($\approx$) & 1.62E+07$\pm$3.90E+06($\approx$) & 1.07E+07$\pm$1.25E+06($\approx$) & 6.23E+06$\pm$1.38E+06($\approx$) & 4.10E+07$\pm$2.16E+07($\approx$) \\\hline
F13 & \textbf{6.32E+11$\pm$1.15E+10} & 7.21E+11$\pm$1.82E+10(+) & 8.36E+11$\pm$1.44E+10(+) & 1.36E+12$\pm$1.49E+11(+) & 1.33E+12$\pm$1.41E+11(+) & 8.26E+11$\pm$5.54E+10(+) & 7.68E+11$\pm$1.20E+11($\approx$) \\\hline
F14 & 2.22E+11$\pm$1.82E+10 & 2.71E+11$\pm$4.86E+09(+) & 2.63E+11$\pm$1.73E+10(+) & 3.14E+11$\pm$3.58E+10(+) & 3.16E+11$\pm$1.52E+10(+) & 2.74E+11$\pm$1.14E+10(+) & \textbf{1.68E+11$\pm$1.19E+10(-)} \\\hline
F15 & 1.72E+04$\pm$1.81E+02 & 1.76E+04$\pm$1.79E+02(+) & 1.79E+04$\pm$2.64E+02(+) & 1.92E+04$\pm$2.34E+02(+) & 1.91E+04$\pm$4.99E+02(+) & 1.78E+04$\pm$2.57E+02(+) & \textbf{1.68E+04$\pm$4.67E+02($\approx$)} \\\hline
F16 & \textbf{4.20E+02$\pm$5.17E-01} & 4.21E+02$\pm$5.29E-01(+) & 4.22E+02$\pm$5.59E-01(+) & 4.25E+02$\pm$6.86E-01(+) & 4.24E+02$\pm$5.39E-01(+) & 4.22E+02$\pm$6.68E-01(+) & 4.23E+02$\pm$9.92E-01(+) \\\hline
F17 & 4.34E+02$\pm$2.94E-01 & 4.34E+02$\pm$4.76E-01($\approx$) & 4.34E+02$\pm$3.74E-01($\approx$) & 4.34E+02$\pm$4.53E-01($\approx$) & \textbf{4.34E+02$\pm$1.47E-01($\approx$)} & 4.34E+02$\pm$2.61E-01($\approx$) & 4.35E+02$\pm$4.26E-01($\approx$) \\\hline
F18 & \textbf{1.39E+12$\pm$7.90E+09} & 1.55E+12$\pm$2.47E+10(+) & 1.70E+12$\pm$3.96E+10(+) & 2.88E+12$\pm$4.65E+10(+) & 2.48E+12$\pm$3.72E+10(+) & 1.73E+12$\pm$4.17E+10(+) & 1.60E+12$\pm$1.12E+11(+) \\\hline
F19 & 7.93E+11$\pm$1.19E+12 & 9.36E+09$\pm$1.30E+10(-) & 2.31E+12$\pm$1.21E+12($\approx$) & 9.87E+09$\pm$9.23E+09($\approx$) & \textbf{1.74E+09$\pm$1.22E+09(-)} & 5.59E+11$\pm$5.40E+11($\approx$) & 8.44E+10$\pm$2.96E+10($\approx$) \\\hline
F20 & \textbf{1.58E+12$\pm$1.05E+10} & 1.70E+12$\pm$4.33E+10(+) & 1.91E+12$\pm$4.57E+10(+) & 3.05E+12$\pm$2.28E+11(+) & 2.60E+12$\pm$1.55E+11(+) & 1.92E+12$\pm$7.98E+10(+) & 1.80E+12$\pm$1.08E+11(+) \\\hline
+/$\approx$/- & NA & 13/5/2 & 13/6/1 & 14/5/1 & 13/5/2 & 15/4/1 & 5/12/3 \\\hline
Average Rank & 2.75 & 2.80 & 4.20 & 5.80 & 4.75 & 4.35 & 3.35 \\\hline
\end{tabular}
    }
    \label{tab:compare off cec2010}
\end{table}

As mentioned in Section 2, both data and surrogate models are crucial for offline data-driven approaches. 
In high-dimensional spaces, training data {is sparse}, leading to diminished surrogate model accuracy and subsequently affecting optimization performance. 
Among the compared algorithms, MS-DDEO 
constructs a model pool consisting of four RBFNs with different smoothness for model selection. However, this diversity is still insufficient in high-dimensional spaces.
Although CL-DDEA increases the amount of data by using the comparative model, the computational expense of training comparative models limits the algorithm to only one such model, causing its performance to deteriorate on large-scale problems. 
Both SRK-DDEA and DDEA-SE integrate multiple surrogate models to improve the performance, yet they do not consider the dynamic update of the surrogate models. As population distributions evolve, the reliability of constructed models decreases. 
BDDEA-LDG increases the data quantity by approximate replacement and has the potential to improve performance. However, the method of generating neighboring sampling points and performing approximate replacement may flatten the fitting landscape around the samples, 
deviating from the true landscape. Therefore, the performance of BDDEA-LDG remains mediocre. TT-DDEA introduces pseudo-labels for individuals in the current population through a semi-supervised approach, enriching the information of surrogate models in the local neighborhood. This grants it good performance. 
However, it 
encounters search stagnation and premature convergence in high-dimentional space. In contrast, the proposed DSKT-DDEA guides diverse subpopulations through distinct surrogate models, and these subpopulations collaborate through adaptive knowledge transfer to optimize diversity in population grouping, avoiding search stagnation and premature convergence. 
Simultaneously, our algorithm enhances the surrogate model capabilities for addressing large-scale optimization problems by leveraging semi-supervised integration and incorporating feedback from diverse distributions within the environment.

\subsubsection{Comparing with the Online DDEAs}
\label{subsec:onlineDDEA}

In this section, we compare our proposed algorithm with seven online DDEAs.
It is important to note that our DSKT-DDEA is still implemented in an offline manner, with the offline dataset fixed at a size of 500. {Since most of these algorithms employ a multi-population technique, the purpose of comparing online DDEAs with our offline DDEA is not to claim the superiority of one over the other but to provide additional reference information regarding the effectiveness of our algorithm in optimizing large-scale expensive problems. Moreover, the offline DDEA's optimization approach makes it a robust starting point or preprocessing step for online algorithms, potentially enhancing their performance.}

Below is a brief overview of the algorithms being compared.

(1) lq-CMA-ES \cite{lq-CMA-ES}: 
This work introduces a global surrogate-assisted CMA-ES method, where the parameters of the surrogate model are calculated using the Moore-Penrose pseudoinverse. The surrogate model is employed only when the rank correlation between the true fitness values and the surrogate fitness values of the most recently sampled data points exceeds a certain threshold. If the accuracy of the surrogate model is found to be insufficient, additional samples are taken from the current population and integrated into the model.

(2) SADE-AMSS \cite{SADE-AMSS}: 
This algorithm integrates Surrogate-Assisted Differential Evolution and Adaptive Multi-Subspace Search techniques, constructing search subspaces through Principal Component Analysis (PCA) and random decision variable selection, thereby effectively accelerating the optimization process. 

(3) CA-LLSO \cite{CA-LLSO}: A classifier-assisted hierarchical learning swarm optimizer founded on the hierarchical learning swarm optimizer and gradient boosting classifier. 

(4) SA-COSO \cite{SA-COSO}: A surrogate-assisted cooperative group optimization algorithm 
employing a multi-group strategy  that alternates between PSO and SL-PSO to enhance the search capability of evolutionary algorithms.

(5) SAMSO \cite{SAMSO}: The algorithm consists of two swarms to collaboratively explore and exploit the search space, and a dynamic group size adjustment scheme is introduced to balance the sizes of the two groups. It avoids premature convergence in certain problems by introducing dynamic group size adjustment methods to enhance communication among different groups.

(6) SAEA-RFS \cite{SAEA-RFS}: Through random feature selection techniques, this algorithm forms multiple subproblems. 
 It updates the solution of the original problem via surrogate-assisted sequential optimization for each subproblem, replacing dimensions linked to the global optimal solution.

(7) ESCO \cite{ESCO}: The algorithm utilizes selective integration surrogates based on KLD to approximate the objective values. 
In each generation, the Kendall
rank correlation coefficient (KRCC) selects an auxiliary subproblem and optimizes it simultaneously with the main problem. The cooperation between populations and subpopulations helps balance the trade-off between exploration and exploitation.

Among these online DDEAs,
lq-CMA-ES is tested on 3, 5, 10, and 20 dimensions. CA-LLSO, SA-COSO, and SAMSO are designed to handle problems with dimensions up to 300, while SADE-AMSS, SAEA-RFS and ESCO are applied to dimensions up to 1000. 

{
The comparison results of our algorithm and the online DDEA on the 1000-dimensional BBOB and CEC2010 test suites are shown in \autoref{tab:base_bench} and \autoref{tab:compare on cec2010}, respectively. According to \autoref{tab:base_bench}, DSKT-DDEA demonstrates potential performance with the same FE budget. Specifically, DSKT-DDEA has an average rank of 2.83, followed by lq-CMA-ES, CA-LLSO, SAMSO, SADE-AMSS, ESCO, SA-COSO, and SAEA-RFS. SAEA-RFS exhibits the poorest performance due to its failure to adequately account for approximation errors in the surrogate model and refine the surrogate within the grouping scheme. Although both SA-COSO and SAMSO adopt a dual-population strategy, the use of a simple surrogate model significantly reduces the search efficiency of these advanced evolutionary algorithms in high-dimensional problems, especially when the number of fitness evaluations is limited. While ESCO divides the population into subpopulations by constructing multiple surrogate models, the efficiency of computing variable correlations declines significantly as dimensionality increases, leading to poor performance on high-dimensional problems. Despite SADE-AMSS constructing subspaces in the mapping space based on past population search information and devising strategies to balance exploration and exploitation, insufficient attention to the surrogate model hampers the algorithm's effectiveness in tackling large-scale problems. CA-LLSO employs a classifier as the surrogate model; however, it struggles to capture subtle changes in population fitness, resulting in poor prediction accuracy. lq-CMA-ES enhances the performance of the surrogate model through active sampling, but the single-population strategy struggles to effectively balance exploration and exploitation in large-scale problems with limited fitness evaluation budgets.

}

\begin{table}
    \centering
    \caption{
    Comparison with Online DDEAs on 1000-Dimensional BBOB
    }
    \resizebox{1\linewidth}{!}{
\arrayrulewidth=0.6pt 

\begin{tabular}{|c|c|c|c|c|c|c|c|c|}
\hline
Benchmark & DSKT-DDEA & lq-CMA-ES & SADE-AMSS & CA-LLSO & SA-COSO & SAMSO & SAEA-RFS & ESCO \\\hline
F1 & \textbf{4.76E+03$\pm$1.75E+01} & 4.91E+03$\pm$5.22E+00(+) & 7.27E+03$\pm$1.59E+01(+) & 6.12E+03$\pm$4.83E+01(+) & 7.91E+03$\pm$1.63E+02(+) & 6.25E+03$\pm$3.32E+02(+) & 8.05E+03$\pm$2.03E+02(+) & 6.38E+03$\pm$2.17E+02(+) \\\hline
F2 & 3.04E+08$\pm$1.24E+07 & 3.77E+08$\pm$9.32E+05(+) & \textbf{2.68E+08$\pm$6.07E+06(-)} & 3.49E+08$\pm$1.92E+07(+) & 3.99E+08$\pm$1.47E+07(+) & 4.31E+08$\pm$3.50E+07(+) & 3.05E+08$\pm$2.65E+07($\approx$) & 3.76E+08$\pm$2.51E+07(+) \\\hline
F3 & \textbf{3.73E+04$\pm$4.50E+02} & 3.96E+04$\pm$9.65E+01(+) & 5.48E+04$\pm$1.68E+03(+) & 4.71E+04$\pm$5.63E+02(+) & 4.87E+04$\pm$1.35E+03(+) & 4.65E+04$\pm$2.23E+03(+) & 5.94E+04$\pm$1.67E+03(+) & 5.46E+04$\pm$9.99E+03(+) \\\hline
F4 & 3.09E+04$\pm$1.46E+02 & \textbf{3.00E+04$\pm$2.04E+02(-)} & 7.46E+04$\pm$2.54E+03(+) & 5.20E+04$\pm$3.88E+03(+) & 6.21E+04$\pm$1.19E+03(+) & 5.21E+04$\pm$5.05E+03(+) & 9.23E+04$\pm$9.41E+03(+) & 7.01E+04$\pm$9.86E+03(+) \\\hline
F5 & 1.63E+04$\pm$2.68E+02 & 1.96E+04$\pm$1.15E+01($\approx$) & 1.60E+04$\pm$5.67E+01(-) & 1.79E+04$\pm$7.13E+01($\approx$) & \textbf{1.54E+04$\pm$3.72E+02($\approx$)} & 1.66E+04$\pm$1.82E+02($\approx$) & 1.56E+04$\pm$5.07E+01($\approx$) & 1.75E+04$\pm$3.92E+02(+) \\\hline
F6 & \textbf{1.81E+07$\pm$6.38E+04} & 1.86E+07$\pm$2.47E+04(+) & 1.88E+07$\pm$3.91E+05(+) & 1.90E+07$\pm$1.76E+05(+) & 2.24E+07$\pm$1.26E+06(+) & 2.00E+07$\pm$2.34E+05($\approx$) & 1.96E+07$\pm$2.42E+05(+) & 1.95E+07$\pm$2.97E+05(+) \\\hline
F7 & \textbf{4.67E+04$\pm$1.03E+03} & 4.96E+04$\pm$2.85E+02(+) & 5.57E+04$\pm$3.12E+03(+) & 5.30E+04$\pm$1.90E+03(+) & 6.48E+04$\pm$2.15E+03(+) & 5.75E+04$\pm$1.88E+03(+) & 6.10E+04$\pm$2.11E+03(+) & 5.90E+04$\pm$2.06E+03(+) \\\hline
F8 & \textbf{4.11E+08$\pm$4.51E+06} & 4.40E+08$\pm$1.30E+06(+) & 2.04E+09$\pm$1.58E+08(+) & 1.24E+09$\pm$3.41E+07(+) & 2.40E+09$\pm$1.50E+08(+) & 1.55E+09$\pm$2.44E+08(+) & 3.05E+09$\pm$1.87E+08(+) & 1.45E+09$\pm$1.10E+08(+) \\\hline
F9 & 1.24E+04$\pm$1.30E+03 & \textbf{9.24E+03$\pm$4.44E+02(-)} & 7.36E+08$\pm$7.79E+07(+) & 2.40E+08$\pm$1.80E+07(+) & 8.89E+08$\pm$3.35E+07(+) & 3.07E+08$\pm$8.35E+07(+) & 2.03E+09$\pm$1.00E+08(+) & 3.45E+08$\pm$5.42E+07(+) \\\hline
F10 & \textbf{2.75E+08$\pm$7.89E+06} & 3.28E+08$\pm$1.19E+06(+) & 2.82E+08$\pm$2.00E+07($\approx$) & 3.27E+08$\pm$1.32E+07(+) & 3.52E+08$\pm$3.76E+07(+) & 3.71E+08$\pm$1.07E+07(+) & 3.12E+08$\pm$3.45E+07($\approx$) & 3.72E+08$\pm$7.47E+07(+) \\\hline
F11 & 6.22E+06$\pm$2.99E+06 & 5.60E+04$\pm$2.39E+04(-) & 1.25E+04$\pm$4.90E+02(-) & \textbf{7.50E+03$\pm$9.19E+01(-)} & 1.08E+04$\pm$6.04E+02(-) & 9.38E+03$\pm$8.03E+02(-) & 1.24E+04$\pm$5.05E+02(-) & 1.31E+04$\pm$4.56E+02(-) \\\hline
F12 & 1.29E+11$\pm$1.59E+11 & \textbf{1.66E+10$\pm$1.60E+08($\approx$)} & 1.79E+10$\pm$1.82E+09($\approx$) & 2.17E+10$\pm$1.97E+09($\approx$) & 3.95E+10$\pm$4.48E+09($\approx$) & 2.45E+10$\pm$4.49E+09($\approx$) & 2.81E+10$\pm$2.71E+09($\approx$) & 3.59E+10$\pm$8.27E+09($\approx$) \\\hline
F13 & \textbf{1.47E+04$\pm$4.40E+01} & 1.49E+04$\pm$4.40E+00(+) & 1.69E+04$\pm$2.28E+02(+) & 1.61E+04$\pm$2.06E+02(+) & 1.81E+04$\pm$3.56E+02(+) & 1.64E+04$\pm$1.47E+02(+) & 1.78E+04$\pm$3.81E+02(+) & 1.69E+04$\pm$4.12E+02(+) \\\hline
F14 & \textbf{3.57E+02$\pm$1.97E+01} & 4.32E+02$\pm$4.71E+00(+) & 4.96E+02$\pm$2.91E+01(+) & 4.97E+02$\pm$8.44E+00(+) & 6.70E+02$\pm$3.24E+01(+) & 5.91E+02$\pm$5.47E+01(+) & 6.31E+02$\pm$4.28E+01(+) & 6.35E+02$\pm$1.39E+02(+) \\\hline
F15 & \textbf{3.67E+04$\pm$3.15E+02} & 3.80E+04$\pm$2.34E+02(+) & 5.10E+04$\pm$2.18E+03(+) & 4.36E+04$\pm$9.39E+02(+) & 5.30E+04$\pm$2.85E+03(+) & 4.36E+04$\pm$2.05E+03(+) & 5.77E+04$\pm$1.54E+03(+) & 4.71E+04$\pm$1.62E+03(+) \\\hline
F16 & -1.79E+02$\pm$4.11E+00 & -1.82E+02$\pm$2.66E-01($\approx$) & -1.91E+02$\pm$9.65E-01(-) & -1.89E+02$\pm$1.46E+00(-) & -1.89E+02$\pm$8.29E-01(-) & -1.90E+02$\pm$1.10E+00(-) & -1.90E+02$\pm$4.63E-01(-) & \textbf{-1.91E+02$\pm$2.36E+00(-)} \\\hline
F17 & -1.17E+01$\pm$7.31E+00 & -1.47E+01$\pm$9.29E-02($\approx$) & -1.44E+01$\pm$2.86E+00($\approx$) & \textbf{-1.64E+01$\pm$6.84E-01($\approx$)} & -1.02E+01$\pm$2.71E+00($\approx$) & -1.16E+01$\pm$1.18E+00($\approx$) & -8.16E+00$\pm$2.83E+00($\approx$) & -1.45E+01$\pm$1.74E+00($\approx$) \\\hline
F18 & 6.64E+01$\pm$3.10E+01 & 5.54E+01$\pm$1.27E+00($\approx$) & 6.42E+01$\pm$6.56E+00($\approx$) & \textbf{4.65E+01$\pm$2.35E+00($\approx$)} & 7.14E+01$\pm$5.89E+00($\approx$) & 6.62E+01$\pm$2.45E+00($\approx$) & 8.04E+01$\pm$8.58E+00($\approx$) & 5.91E+01$\pm$5.06E+00($\approx$) \\\hline
F19 & 4.96E+01$\pm$1.04E-01 & \textbf{4.96E+01$\pm$1.70E-01($\approx$)} & 1.73E+03$\pm$3.92E+02(+) & 6.59E+02$\pm$8.69E+01(+) & 2.60E+03$\pm$2.06E+02(+) & 3.40E+02$\pm$6.33E+01(+) & 4.95E+03$\pm$4.10E+02(+) & 9.61E+02$\pm$1.92E+02(+) \\\hline
F20 & \textbf{6.50E+05$\pm$3.88E+04} & 7.98E+05$\pm$9.88E+03(+) & 3.24E+06$\pm$2.60E+05(+) & 2.04E+06$\pm$2.45E+05(+) & 3.92E+06$\pm$2.39E+05(+) & 2.20E+06$\pm$2.05E+05(+) & 3.99E+06$\pm$1.47E+05(+) & 2.39E+06$\pm$2.02E+05(+) \\\hline
F21 & 3.95E+02$\pm$1.15E-02 & \textbf{3.95E+02$\pm$2.48E-02($\approx$)} & 3.96E+02$\pm$1.18E-01(+) & 3.96E+02$\pm$7.89E-02(+) & 3.96E+02$\pm$8.55E-02(+) & 3.96E+02$\pm$1.40E-01(+) & 3.96E+02$\pm$8.42E-02(+) & 3.97E+02$\pm$2.08E-01(+) \\\hline
F22 & \textbf{1.26E+02$\pm$2.99E-02} & 1.27E+02$\pm$1.52E-02(+) & 1.28E+02$\pm$1.19E-01(+) & 1.28E+02$\pm$3.76E-02(+) & 1.28E+02$\pm$1.50E-01(+) & 1.28E+02$\pm$1.21E-01(+) & 1.28E+02$\pm$1.86E-01(+) & 1.29E+02$\pm$2.85E-01(+) \\\hline
F23 & 1.50E+03$\pm$1.87E-02 & 1.82E+03$\pm$7.57E-01(+) & 1.50E+03$\pm$0.00E+00(-) & \textbf{1.50E+03$\pm$7.73E-03(-)} & 1.50E+03$\pm$2.02E-02(-) & 1.50E+03$\pm$9.70E-03(-) & 1.50E+03$\pm$4.84E-03(-) & 1.50E+03$\pm$9.91E-03(-) \\\hline
F24 & \textbf{1.62E+04$\pm$7.73E+01} & 1.64E+04$\pm$3.23E+02($\approx$) & 2.76E+04$\pm$8.77E+02(+) & 2.25E+04$\pm$5.18E+02(+) & 3.03E+04$\pm$1.27E+03(+) & 2.11E+04$\pm$5.01E+02(+) & 3.76E+04$\pm$6.00E+02(+) & 2.30E+04$\pm$9.61E+02(+) \\\hline
+/$\approx$/- & NA & 13/8/3 & 15/4/5 & 17/4/3 & 17/4/3 & 16/5/3 & 15/6/3 & 18/3/3 \\\hline
Average Rank & 2.83 & 3.00 & 4.54 & 3.25 & 6.21 & 4.50 & 6.42 & 5.25 \\\hline
\end{tabular}
    }
    \label{tab:base_bench}
\end{table}

\begin{table}
    \centering
    \caption{
   Comparison with Online DDEAs on the 1000-Dimensional CEC2010 Benchmark Functions
    }
    \resizebox{1\linewidth}{!}{

\begin{tabular}{|c|c|c|c|c|c|c|c|c|}
\hline
Benchmark & DSKT-DDEA & lq-CMA-ES & SADE-AMSS & CA-LLSO & SA-COSO & SAMSO & SAEA-RFS & ESCO \\\hline
F1 & 1.61E+11$\pm$8.47E+09 & 2.00E+11$\pm$5.85E+07(+) & \textbf{1.40E+11$\pm$3.59E+09(-)} & 1.85E+11$\pm$1.80E+09(+) & 1.97E+11$\pm$1.34E+10(+) & 1.99E+11$\pm$5.65E+09(+) & 1.40E+11$\pm$3.93E+09(-) & 1.98E+11$\pm$3.03E+10(+) \\\hline
F2 & 1.70E+04$\pm$1.16E+02 & \textbf{1.69E+04$\pm$1.08E+02($\approx$)} & 2.00E+04$\pm$1.51E+02(+) & 1.83E+04$\pm$3.01E+02(+) & 1.97E+04$\pm$5.05E+02(+) & 1.84E+04$\pm$1.78E+02(+) & 1.93E+04$\pm$1.25E+02(+) & 1.85E+04$\pm$2.30E+02(+) \\\hline
F3 & \textbf{2.10E+01$\pm$2.69E-02} & 2.10E+01$\pm$1.15E-02($\approx$) & 2.13E+01$\pm$2.73E-02(+) & 2.12E+01$\pm$1.19E-02(+) & 2.13E+01$\pm$2.87E-02(+) & 2.13E+01$\pm$2.62E-02(+) & 2.13E+01$\pm$1.17E-02(+) & 2.14E+01$\pm$7.73E-02(+) \\\hline
F4 & 2.08E+16$\pm$1.64E+16 & 7.66E+15$\pm$1.14E+13($\approx$) & \textbf{6.22E+14$\pm$1.15E+14(-)} & 1.51E+15$\pm$3.91E+14(-) & 1.30E+15$\pm$3.78E+14(-) & 1.45E+15$\pm$2.64E+14(-) & 7.14E+14$\pm$2.53E+14(-) & 2.48E+15$\pm$9.23E+14(-) \\\hline
F5 & 7.59E+08$\pm$8.02E+07 & 9.12E+08$\pm$5.25E+07(+) & 9.11E+08$\pm$1.96E+07(+) & 7.14E+08$\pm$4.14E+06($\approx$) & \textbf{6.29E+08$\pm$3.46E+07(-)} & 7.85E+08$\pm$2.80E+07($\approx$) & 7.95E+08$\pm$6.17E+07($\approx$) & 8.17E+08$\pm$1.47E+08($\approx$) \\\hline
F6 & 2.06E+07$\pm$4.13E+05 & 2.09E+07$\pm$5.42E+04($\approx$) & 2.11E+07$\pm$1.38E+05($\approx$) & 2.04E+07$\pm$2.55E+05($\approx$) & \textbf{1.95E+07$\pm$9.01E+05($\approx$)} & 2.03E+07$\pm$7.60E+04($\approx$) & 2.09E+07$\pm$2.95E+05($\approx$) & 2.08E+07$\pm$1.72E+05($\approx$) \\\hline
F7 & 2.16E+14$\pm$6.87E+13 & 2.03E+13$\pm$6.99E+09(-) & 1.68E+11$\pm$1.16E+10(-) & 7.97E+11$\pm$6.08E+10(-) & 4.73E+11$\pm$1.73E+11(-) & 3.81E+11$\pm$9.08E+10(-) & \textbf{1.56E+11$\pm$3.58E+10(-)} & 9.82E+11$\pm$4.60E+11(-) \\\hline
F8 & 2.44E+16$\pm$4.20E+15 & 6.71E+16$\pm$1.22E+14(+) & \textbf{2.63E+15$\pm$5.19E+14(-)} & 2.97E+16$\pm$3.86E+15($\approx$) & 2.53E+16$\pm$9.86E+15($\approx$) & 3.52E+16$\pm$1.24E+16($\approx$) & 1.09E+16$\pm$6.88E+15(-) & 3.83E+16$\pm$2.74E+16($\approx$) \\\hline
F9 & 2.02E+11$\pm$8.68E+09 & 2.41E+11$\pm$5.20E+07(+) & 1.69E+11$\pm$1.13E+10(-) & 2.11E+11$\pm$9.74E+09($\approx$) & 2.11E+11$\pm$6.96E+09($\approx$) & 2.40E+11$\pm$1.89E+10(+) & \textbf{1.65E+11$\pm$6.77E+09(-)} & 2.42E+11$\pm$5.33E+10($\approx$) \\\hline
F10 & 1.74E+04$\pm$1.51E+02 & \textbf{1.73E+04$\pm$1.56E+02($\approx$)} & 1.99E+04$\pm$1.46E+02(+) & 1.82E+04$\pm$1.43E+02(+) & 2.00E+04$\pm$2.67E+02(+) & 1.86E+04$\pm$3.09E+02(+) & 2.13E+04$\pm$4.55E+02(+) & 1.92E+04$\pm$1.97E+02(+) \\\hline
F11 & \textbf{2.31E+02$\pm$2.76E-01} & 2.31E+02$\pm$1.31E-01(+) & 2.36E+02$\pm$1.90E-01(+) & 2.32E+02$\pm$3.70E-01(+) & 2.34E+02$\pm$5.15E-01(+) & 2.33E+02$\pm$5.31E-01(+) & 2.36E+02$\pm$3.19E-01(+) & 2.34E+02$\pm$4.95E-01(+) \\\hline
F12 & 2.70E+07$\pm$4.74E+07 & 5.53E+06$\pm$6.54E+03($\approx$) & 9.16E+06$\pm$4.01E+05($\approx$) & \textbf{4.49E+06$\pm$3.67E+05($\approx$)} & 1.03E+07$\pm$1.16E+06($\approx$) & 1.59E+07$\pm$1.49E+06($\approx$) & 6.07E+06$\pm$6.71E+05($\approx$) & 8.54E+06$\pm$1.11E+06($\approx$) \\\hline
F13 & \textbf{6.32E+11$\pm$1.15E+10} & 7.01E+11$\pm$2.78E+08(+) & 1.17E+12$\pm$4.07E+10(+) & 9.59E+11$\pm$5.91E+10(+) & 1.64E+12$\pm$1.50E+11(+) & 1.18E+12$\pm$1.01E+11(+) & 1.24E+12$\pm$1.35E+11(+) & 1.12E+12$\pm$6.43E+10(+) \\\hline
F14 & 2.22E+11$\pm$1.82E+10 & 2.73E+11$\pm$3.36E+07(+) & \textbf{1.74E+11$\pm$9.40E+09(-)} & 2.27E+11$\pm$4.81E+09($\approx$) & 2.20E+11$\pm$1.28E+10($\approx$) & 2.68E+11$\pm$1.08E+10(+) & 1.76E+11$\pm$1.13E+10(-) & 2.75E+11$\pm$3.27E+10(+) \\\hline
F15 & \textbf{1.72E+04$\pm$1.81E+02} & 1.73E+04$\pm$7.30E+01($\approx$) & 2.14E+04$\pm$7.74E+02(+) & 1.82E+04$\pm$1.20E+02(+) & 2.00E+04$\pm$2.91E+02(+) & 1.88E+04$\pm$4.38E+02(+) & 2.25E+04$\pm$2.74E+02(+) & 1.94E+04$\pm$1.40E+02(+) \\\hline
F16 & 4.20E+02$\pm$5.17E-01 & \textbf{4.19E+02$\pm$1.76E-01(-)} & 4.29E+02$\pm$1.73E-01(+) & 4.22E+02$\pm$4.00E-01(+) & 4.26E+02$\pm$5.01E-01(+) & 4.24E+02$\pm$9.95E-01(+) & 4.29E+02$\pm$3.21E-01(+) & 4.26E+02$\pm$2.48E-01(+) \\\hline
F17 & 4.34E+02$\pm$2.94E-01 & 4.34E+02$\pm$1.62E-01($\approx$) & 1.97E+07$\pm$1.30E+06(+) & 5.33E+02$\pm$1.93E-01($\approx$) & 4.33E+02$\pm$1.01E-01(-) & 3.85E+07$\pm$5.25E+06(+) & \textbf{4.29E+02$\pm$3.68E-01(-)} & 4.33E+02$\pm$3.67E-01(-) \\\hline
F18 & \textbf{1.39E+12$\pm$7.90E+09} & 1.48E+12$\pm$4.47E+08(+) & 3.19E+12$\pm$1.68E+11(+) & 2.36E+12$\pm$4.20E+10(+) & 3.72E+12$\pm$1.53E+11(+) & 2.72E+12$\pm$2.74E+11(+) & 3.79E+12$\pm$1.85E+11(+) & 2.78E+12$\pm$1.16E+11(+) \\\hline
F19 & 7.93E+11$\pm$1.19E+12 & 3.10E+09$\pm$4.33E+06(-) & \textbf{5.89E+07$\pm$5.39E+06(-)} & 9.72E+07$\pm$1.06E+07(-) & 1.59E+08$\pm$2.00E+07(-) & 9.57E+07$\pm$1.05E+07(-) & 6.23E+07$\pm$9.31E+06(-) & 1.61E+08$\pm$5.14E+07(-) \\\hline
F20 & \textbf{1.58E+12$\pm$1.05E+10} & 1.66E+12$\pm$5.67E+08(+) & 3.31E+12$\pm$1.24E+11(+) & 2.59E+12$\pm$7.25E+10(+) & 4.22E+12$\pm$2.30E+11(+) & 3.01E+12$\pm$1.10E+11(+) & 4.32E+12$\pm$1.70E+11(+) & 2.89E+12$\pm$1.75E+11(+) \\\hline
+/$\approx$/- & NA & 9/8/3 & 11/2/7 & 10/7/3 & 10/5/5 & 13/4/3 & 9/3/8 & 11/5/4 \\\hline
Average Rank & 3.45 & 4.35 & 4.80 & 3.60 & 4.85 & 4.75 & 4.60 & 5.60 \\\hline
\end{tabular}
    }
    \label{tab:compare on cec2010}
\end{table}
\subsection{In-depth Analysis}
\label{subsec:ablation}

\subsubsection{Ablation Study}
In this section, we discuss the contributions of the main components in DSKT-DDEA. We first provide a brief overview of the relevant variants of DSKT-DDEA and then compare them with each other.

(1) DSKT-DDEA-w/o-H: In this variant, the surrogate models across each island exhibit isomorphism, meaning that the training datasets are consistent across all islands. It is used to investigate the effectiveness of our diverse surrogates for island diversification.

(2) DSKT-DDEA-w/o-F: This model removes the semi-supervised model fine-tuning process, where each subpopulation trains the surrogate model once at the beginning and does not update it afterwards. 
Meanwhile, each subpopulation only uses its own surrogate model for selecting offspring.

(3) DSKT-DDEA-w/o-M: This variant removes the adaptive knowledge transfer process to study the effectiveness of our designed migration strategy. 

(4) DSKT-DDEA-w/o-a: This variant removes the calculation of attractiveness (\autoref{eq:a}) when calculating migration probabilities, that is, the attractiveness of each island is 1.

(5) DSKT-DDEA-w/o-d: This variant removes the differential factor (\autoref{eq:v}) when calculating migration probabilities, that is, the Differential Factor of each island is 1.

(6) DSKT-DDEA-blank: This variant serves as a blank controller, eliminating several new components including diverse surrogate models, semi-supervised model fine-tuning process, and adaptive knowledge transfer. 

\begin{table}
\centering
\caption{Results of Ablation Experiment 1000-Dimensional CEC2010 Benchmark Functions}
\resizebox{0.9\linewidth}{!}{

\begin{tabular}{|c|c|c|c|c|c|c|c|}
\hline
Benchmark & DSKT-DDEA & DSKT-DDEA-w/o-H & DSKT-DDEA-w/o-F & DSKT-DDEA-w/o-M & DSKT-DDEA-w/o-a & DSKT-DDEA-w/o-d & DSKT-DDEA-w/o-blank \\\hline
F1 & \textbf{1.61E+11$\pm$8.47E+09} & 1.78E+11$\pm$9.14E+09($\approx$) & 1.76E+11$\pm$4.58E+09($\approx$) & 1.79E+11$\pm$4.49E+09(+) & 1.80E+11$\pm$3.87E+09(+) & 1.78E+11$\pm$4.89E+09(+) & 1.82E+11$\pm$9.70E+08(+) \\\hline
F2 & 1.70E+04$\pm$1.16E+02 & 1.71E+04$\pm$1.80E+02($\approx$) & 1.71E+04$\pm$2.09E+02($\approx$) & 1.70E+04$\pm$6.84E+01($\approx$) & \textbf{1.70E+04$\pm$1.40E+02($\approx$)} & 1.70E+04$\pm$5.36E+01($\approx$) & 1.73E+04$\pm$4.02E+01(+) \\\hline
F3 & 2.10E+01$\pm$2.69E-02 & \textbf{2.10E+01$\pm$1.25E-02($\approx$)} & 2.11E+01$\pm$1.06E-02($\approx$) & 2.11E+01$\pm$2.29E-02($\approx$) & 2.11E+01$\pm$1.57E-02($\approx$) & 2.11E+01$\pm$2.78E-02($\approx$) & 2.12E+01$\pm$1.27E-02(+) \\\hline
F4 & 2.08E+16$\pm$1.64E+16 & 7.93E+15$\pm$5.26E+15($\approx$) & 7.69E+15$\pm$7.16E+15($\approx$) & 1.09E+16$\pm$1.13E+16($\approx$) & \textbf{5.05E+15$\pm$1.37E+15($\approx$)} & 9.21E+15$\pm$9.27E+15($\approx$) & 6.32E+15$\pm$3.77E+14($\approx$) \\\hline
F5 & \textbf{7.59E+08$\pm$8.02E+07} & 8.14E+08$\pm$3.59E+07($\approx$) & 8.26E+08$\pm$7.28E+07($\approx$) & 8.91E+08$\pm$3.75E+07(+) & 8.96E+08$\pm$1.40E+08(+) & 8.60E+08$\pm$5.54E+07($\approx$) & 9.05E+08$\pm$3.13E+06($\approx$) \\\hline
F6 & 2.06E+07$\pm$4.13E+05 & 2.10E+07$\pm$5.28E+05($\approx$) & 2.07E+07$\pm$4.46E+04($\approx$) & 2.08E+07$\pm$4.73E+05($\approx$) & 2.08E+07$\pm$3.70E+05($\approx$) & \textbf{2.06E+07$\pm$1.72E+05($\approx$)} & 2.17E+07$\pm$1.01E+05(+) \\\hline
F7 & 2.16E+14$\pm$6.87E+13 & \textbf{2.93E+13$\pm$1.72E+13(-)} & 6.62E+13$\pm$3.91E+13(-) & 7.57E+13$\pm$8.41E+13(-) & 7.77E+13$\pm$6.18E+13(-) & 9.61E+13$\pm$8.05E+13(-) & 3.52E+13$\pm$2.79E+13(-) \\\hline
F8 & \textbf{2.44E+16$\pm$4.20E+15} & 3.91E+16$\pm$3.76E+15(+) & 3.71E+16$\pm$7.16E+15($\approx$) & 3.59E+16$\pm$7.13E+15($\approx$) & 3.26E+16$\pm$9.81E+15($\approx$) & 3.21E+16$\pm$8.19E+15($\approx$) & 9.16E+16$\pm$1.12E+17(+) \\\hline
F9 & \textbf{2.02E+11$\pm$8.68E+09} & 2.10E+11$\pm$1.51E+10($\approx$) & 2.16E+11$\pm$5.04E+09($\approx$) & 2.19E+11$\pm$7.63E+09(+) & 2.19E+11$\pm$5.83E+09(+) & 2.18E+11$\pm$5.59E+09(+) & 2.07E+11$\pm$7.52E+09($\approx$) \\\hline
F10 & 1.74E+04$\pm$1.51E+02 & 1.75E+04$\pm$1.41E+02($\approx$) & 1.74E+04$\pm$2.28E+02($\approx$) & 1.74E+04$\pm$1.04E+02($\approx$) & 1.74E+04$\pm$7.54E+01($\approx$) & \textbf{1.73E+04$\pm$9.09E+01($\approx$)} & 1.74E+04$\pm$1.44E+02($\approx$) \\\hline
F11 & \textbf{2.31E+02$\pm$2.76E-01} & 2.31E+02$\pm$2.76E-01($\approx$) & 2.31E+02$\pm$5.18E-02($\approx$) & 2.31E+02$\pm$1.70E-01(+) & 2.31E+02$\pm$3.41E-01($\approx$) & 2.31E+02$\pm$3.33E-01($\approx$) & 2.31E+02$\pm$1.19E-01(+) \\\hline
F12 & 2.70E+07$\pm$4.74E+07 & \textbf{2.63E+06$\pm$4.13E+05($\approx$)} & 5.10E+06$\pm$2.84E+06($\approx$) & 1.68E+07$\pm$2.80E+07($\approx$) & 1.40E+07$\pm$2.01E+07($\approx$) & 1.41E+07$\pm$2.23E+07($\approx$) & 5.94E+06$\pm$2.15E+06($\approx$) \\\hline
F13 & \textbf{6.32E+11$\pm$1.15E+10} & 6.62E+11$\pm$7.23E+09(+) & 6.57E+11$\pm$7.58E+09(+) & 6.67E+11$\pm$1.23E+10(+) & 6.68E+11$\pm$1.20E+10(+) & 6.63E+11$\pm$1.23E+10(+) & 6.55E+11$\pm$9.47E+09(+) \\\hline
F14 & \textbf{2.22E+11$\pm$1.82E+10} & 2.40E+11$\pm$6.40E+09($\approx$) & 2.43E+11$\pm$6.36E+09($\approx$) & 2.47E+11$\pm$1.23E+10($\approx$) & 2.47E+11$\pm$1.52E+10($\approx$) & 2.47E+11$\pm$1.11E+10($\approx$) & 2.38E+11$\pm$4.30E+09($\approx$) \\\hline
F15 & \textbf{1.72E+04$\pm$1.81E+02} & 1.73E+04$\pm$2.09E+02($\approx$) & 1.72E+04$\pm$2.31E+02($\approx$) & 1.73E+04$\pm$1.08E+02($\approx$) & 1.73E+04$\pm$1.69E+02($\approx$) & 1.73E+04$\pm$1.29E+02($\approx$) & 1.73E+04$\pm$1.42E+02($\approx$) \\\hline
F16 & \textbf{4.20E+02$\pm$5.17E-01} & 4.20E+02$\pm$4.03E-01($\approx$) & 4.20E+02$\pm$4.44E-01($\approx$) & 4.20E+02$\pm$5.25E-01($\approx$) & 4.20E+02$\pm$4.40E-01($\approx$) & 4.20E+02$\pm$1.83E-01($\approx$) & 4.21E+02$\pm$3.78E-01($\approx$) \\\hline
F17 & 4.34E+02$\pm$2.94E-01 & 4.34E+02$\pm$2.47E-01($\approx$) & 4.34E+02$\pm$2.91E-01($\approx$) & 4.34E+02$\pm$2.83E-01($\approx$) & \textbf{4.34E+02$\pm$4.33E-01($\approx$)} & 4.34E+02$\pm$2.10E-01($\approx$) & 4.34E+02$\pm$2.99E-01($\approx$) \\\hline
F18 & \textbf{1.39E+12$\pm$7.90E+09} & 1.42E+12$\pm$6.67E+09(+) & 1.43E+12$\pm$6.67E+09(+) & 1.43E+12$\pm$1.01E+10(+) & 1.44E+12$\pm$1.38E+10(+) & 1.44E+12$\pm$6.10E+09(+) & 1.43E+12$\pm$1.62E+10(+) \\\hline
F19 & 7.93E+11$\pm$1.19E+12 & 5.05E+11$\pm$3.63E+11($\approx$) & 1.71E+12$\pm$1.20E+12($\approx$) & \textbf{3.49E+11$\pm$3.41E+11($\approx$)} & 3.71E+11$\pm$4.17E+11($\approx$) & 5.96E+11$\pm$4.61E+11($\approx$) & 9.30E+11$\pm$1.13E+12($\approx$) \\\hline
F20 & \textbf{1.58E+12$\pm$1.05E+10} & 1.61E+12$\pm$1.05E+10($\approx$) & 1.60E+12$\pm$4.18E+09($\approx$) & 1.62E+12$\pm$1.31E+10(+) & 1.62E+12$\pm$1.79E+10(+) & 1.61E+12$\pm$8.09E+09(+) & 1.65E+12$\pm$3.56E+10(+) \\\hline
+/$\approx$/- & NA & 3/16/1 & 2/17/1 & 7/12/1 & 6/13/1 & 5/14/1 & 9/10/1 \\\hline
Average Rank & 2.70 & 3.65 & 3.40 & 4.60 & 4.40 & 4.35 & 4.90 \\\hline
\end{tabular}

}
\label{tab:ablation}
\end{table}

The experimental results are shown in \autoref{tab:ablation}, and our analysis is as follows:

\begin{itemize}
  \item DSKT-DDEA performs significantly better than other variants, indicating the clear advantage of our DSKT-DDEA in solving large-scale problems. 
    \item Compared to DSKT-DDEA, the performance of DSKT-DDEA-w/o-H is somewhat inferior. This indicates that diverse surrogate models have the capacity to guide subpopulations in different directions, thereby avoiding premature convergence. 

  \item DSKT-DDEA-w/o-F is completely inferior to DSKT-DDEA, indicating that the semi-supervised model fine-tuning is an effective approach for solving large-scale and complex multimodal problems. It not only increases the amount of data but also enhances the ability of the surrogate model to 
  fit the local problem landscape.

  \item In most cases, DSKT-DDEA-w/o-M performs worse than DSKT-DDEA. The results suggest that our attractiveness-based periodic knowledge transfer mechanism can transfer excellent gene fragments that are conducive to convergence, facilitating global exploration and exploitation.
  
  \item When calculating migration probabilities, removing the calculation of attractiveness (DSKT-DDEA-w/o-a) and diversity (DSKT-DDEA-w/o-d) will have an impact on the results of the algorithm. When the attractiveness is removed, the migration probability is completely dominated by the diversity of the island, so the algorithm cannot gain experience from historical migrations. When the diversity factor is removed, islands with more historical migrations will be more attractive, leading to a loss of diversity.
  
 \item DSKT-DDEA-blank performs the worst, as can be expected. 
  
\end{itemize}

\subsubsection{Surrogate Accuracy}
\begin{figure}
    \centering
    \begin{tabular}{ccccc}
        \includegraphics[height=0.15\textwidth,keepaspectratio]{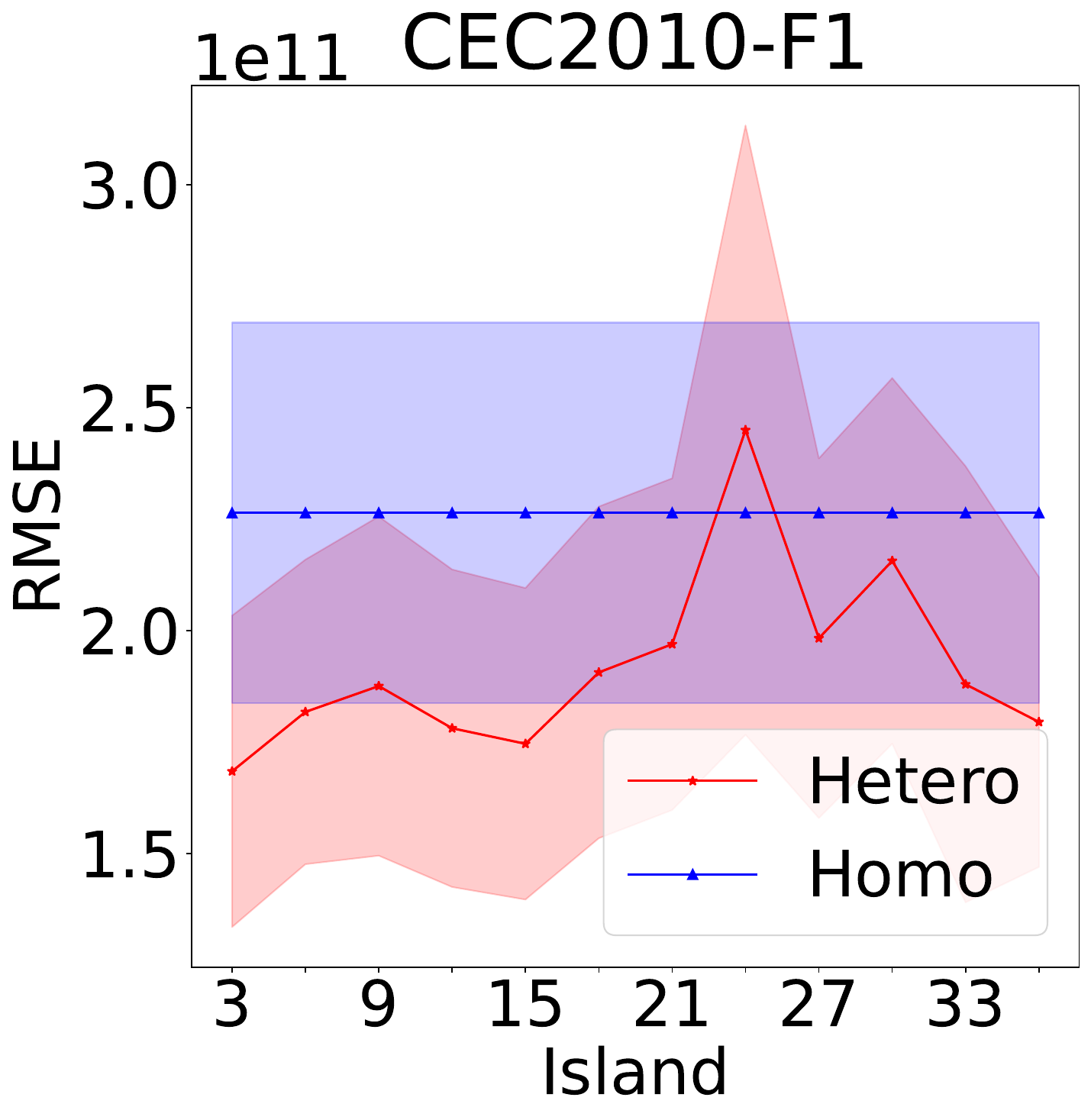}&
        \includegraphics[height=0.15\textwidth, keepaspectratio]{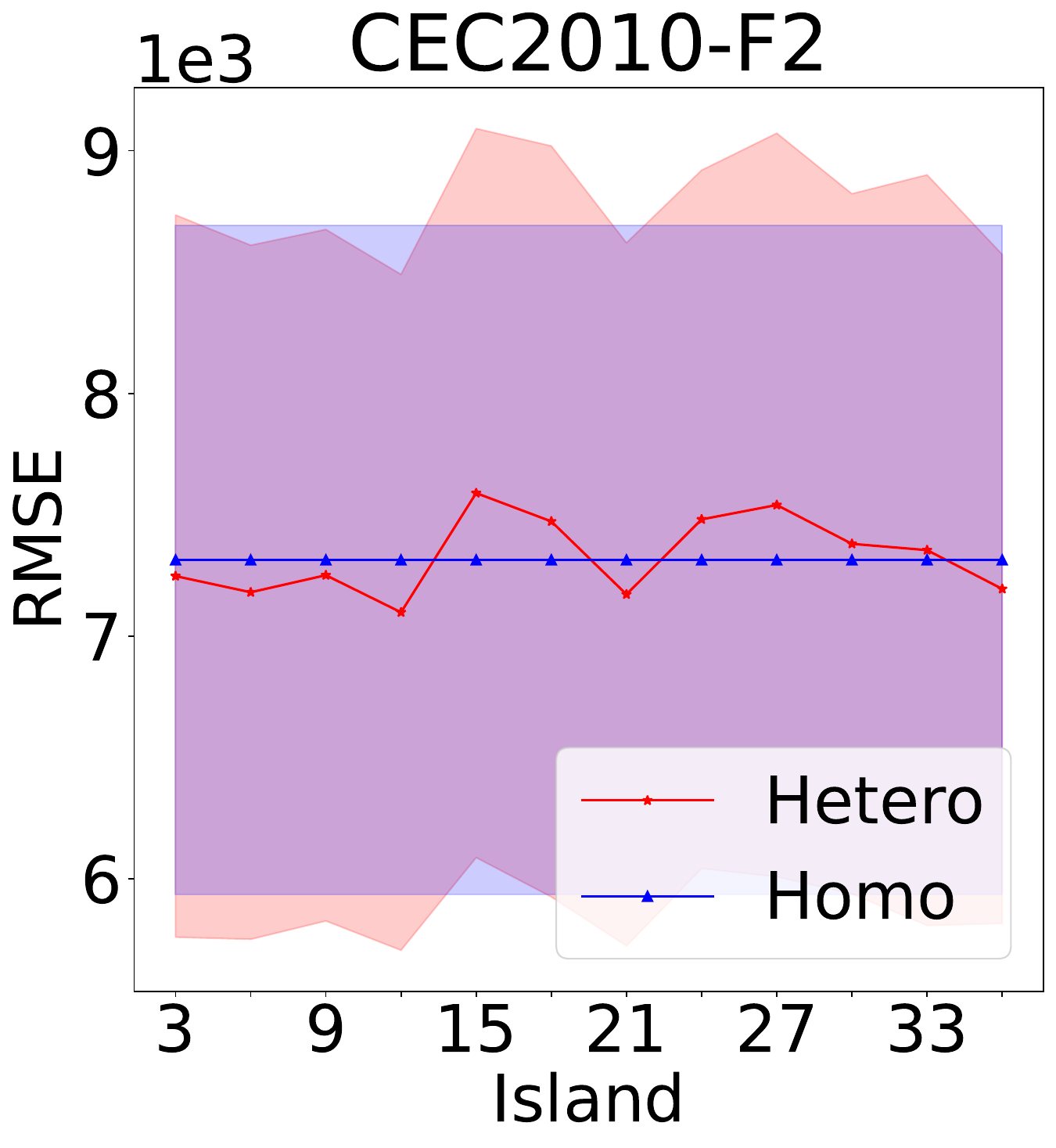}&
        \includegraphics[height=0.15\textwidth, keepaspectratio]{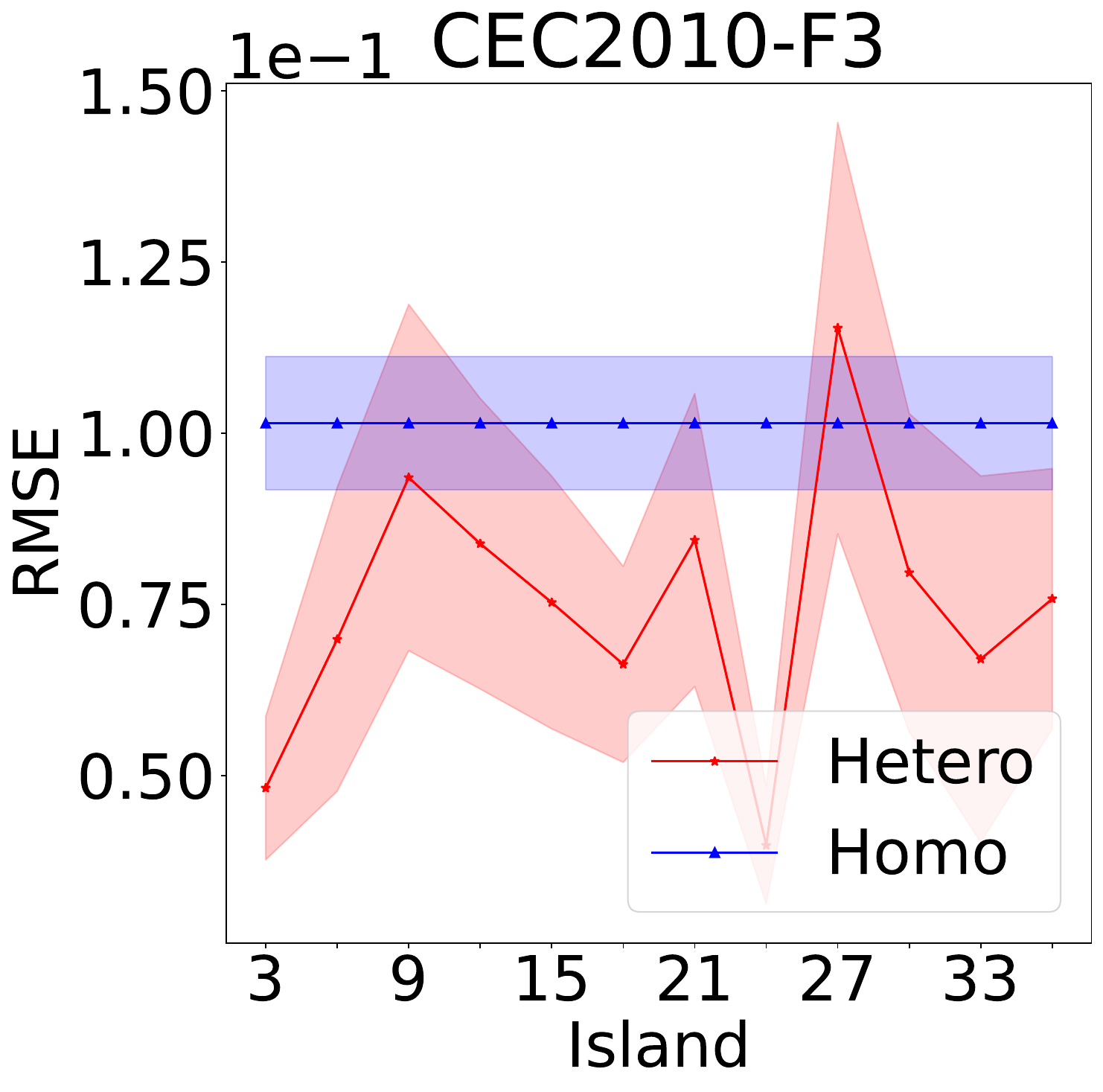}&
        \includegraphics[height=0.15\textwidth, keepaspectratio]{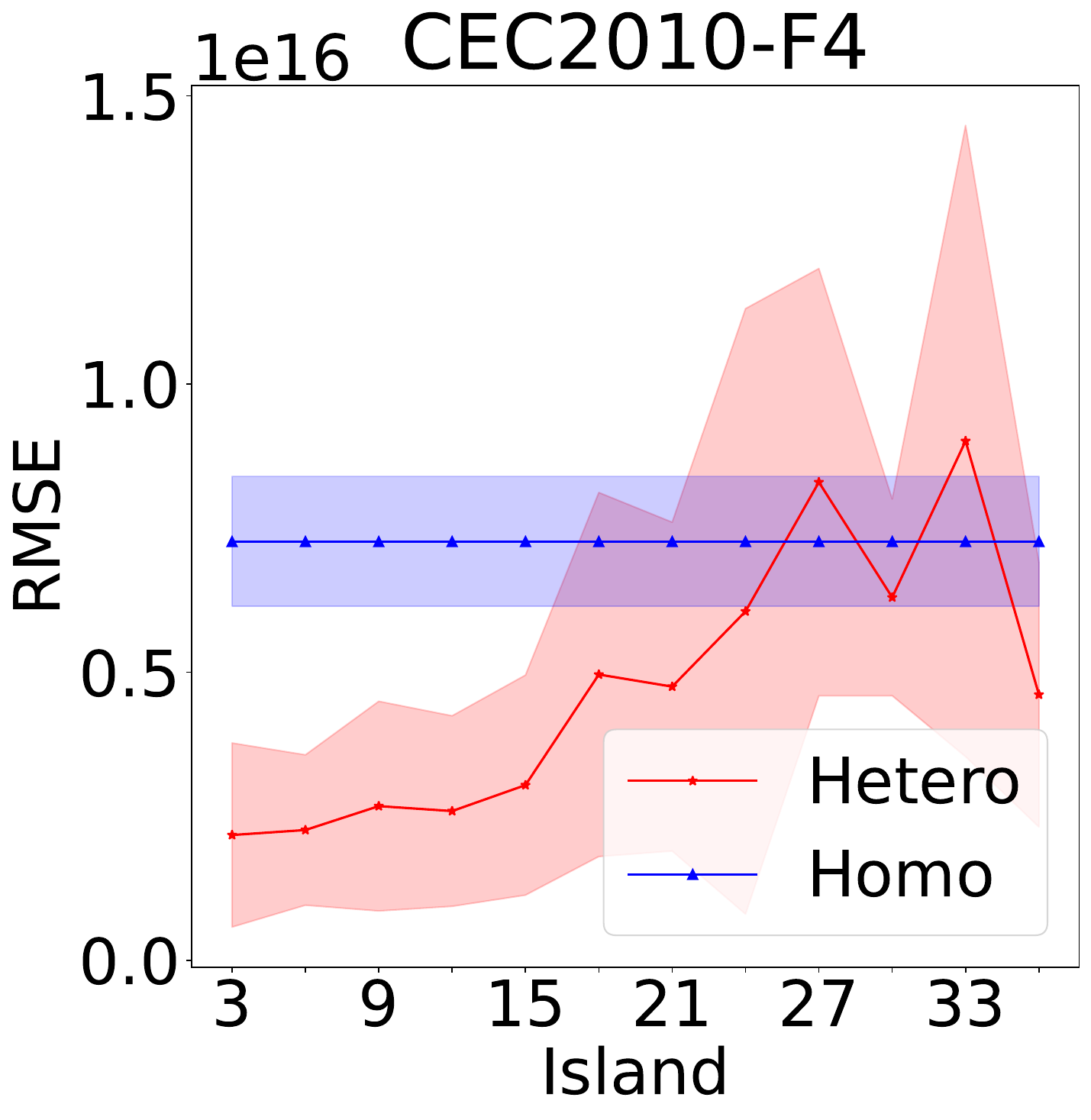}&
        \includegraphics[height=0.15\textwidth, keepaspectratio]{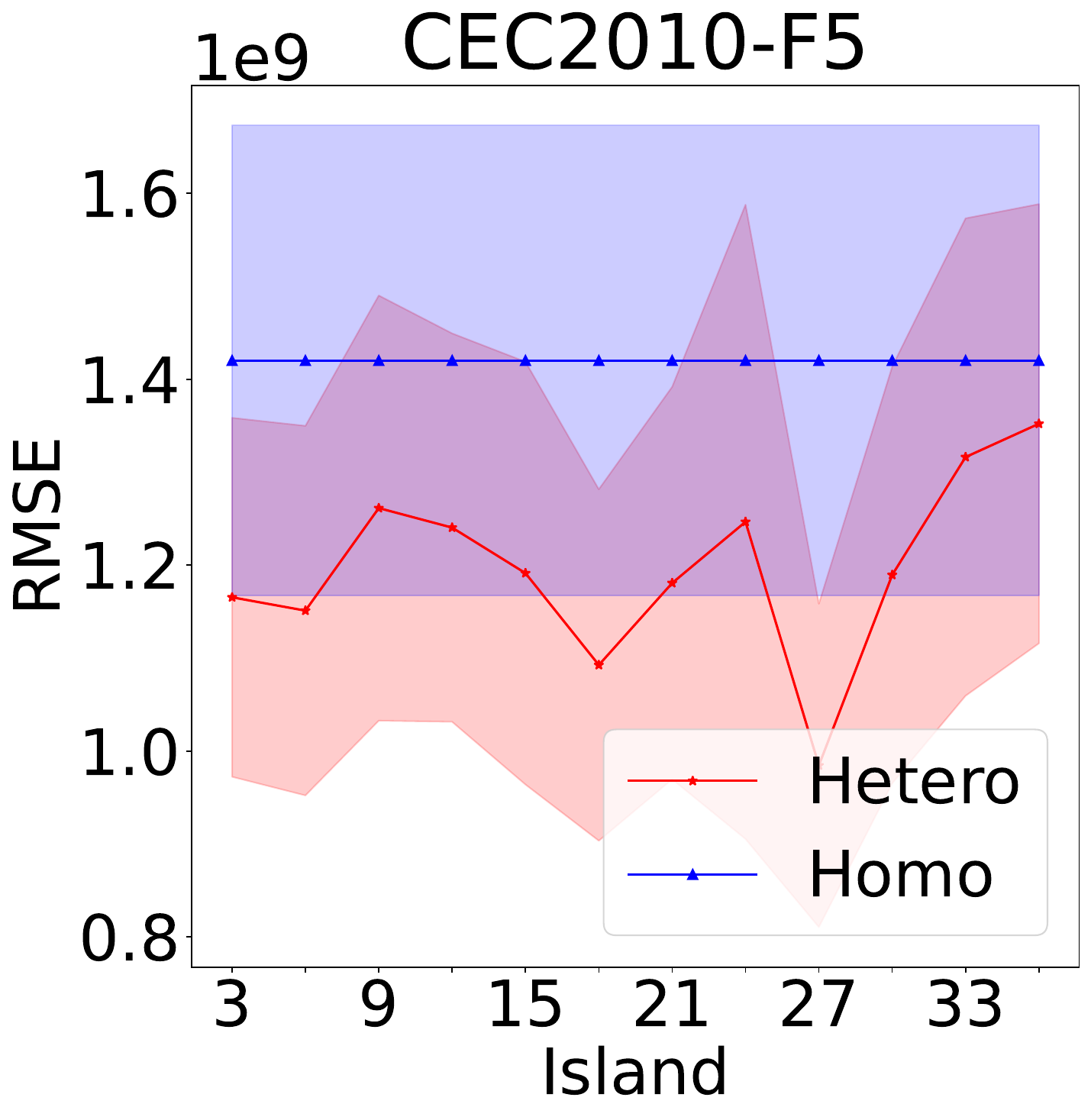}\\
        \includegraphics[height=0.15\textwidth, keepaspectratio]{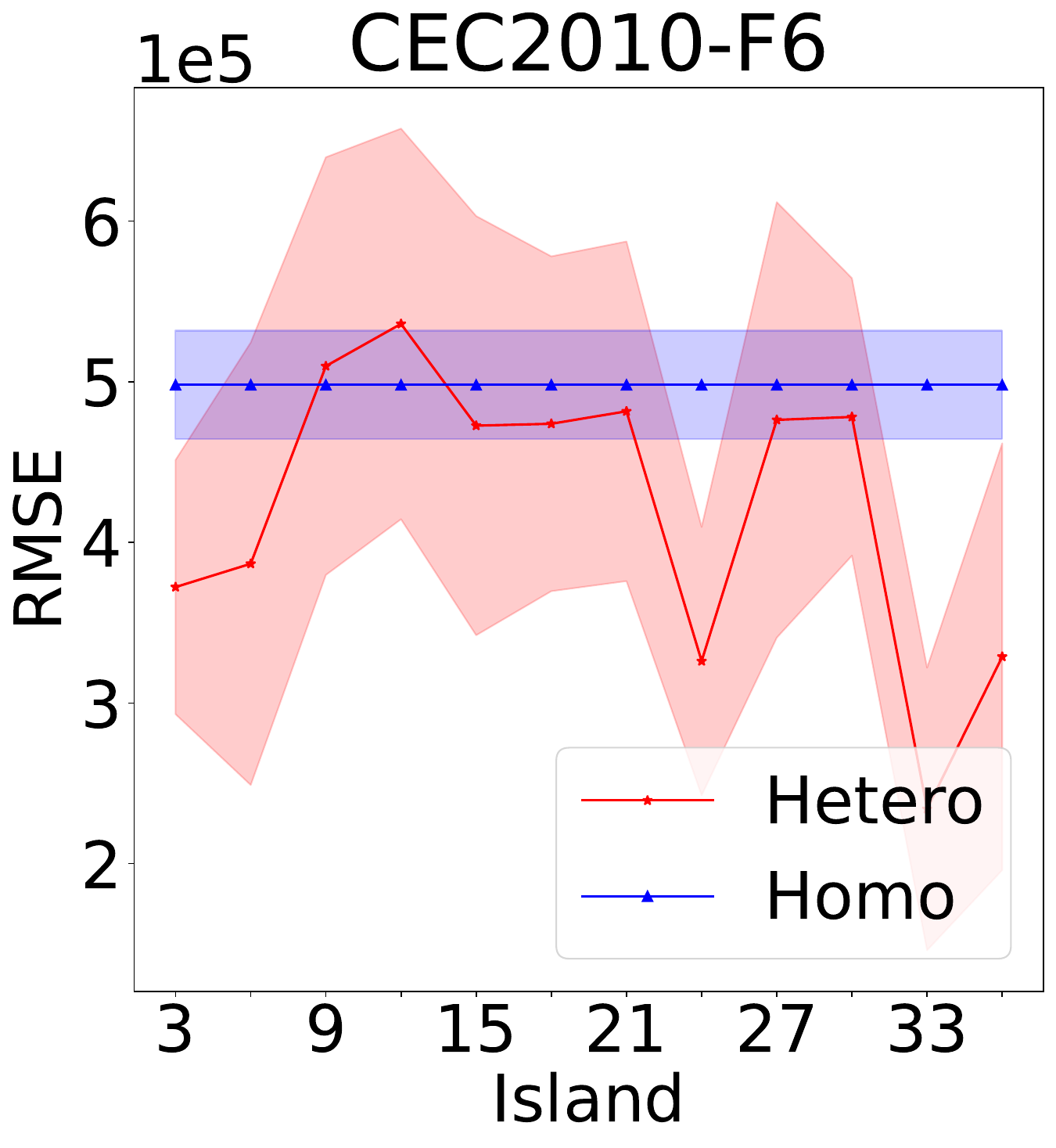}&
        \includegraphics[height=0.15\textwidth, keepaspectratio]{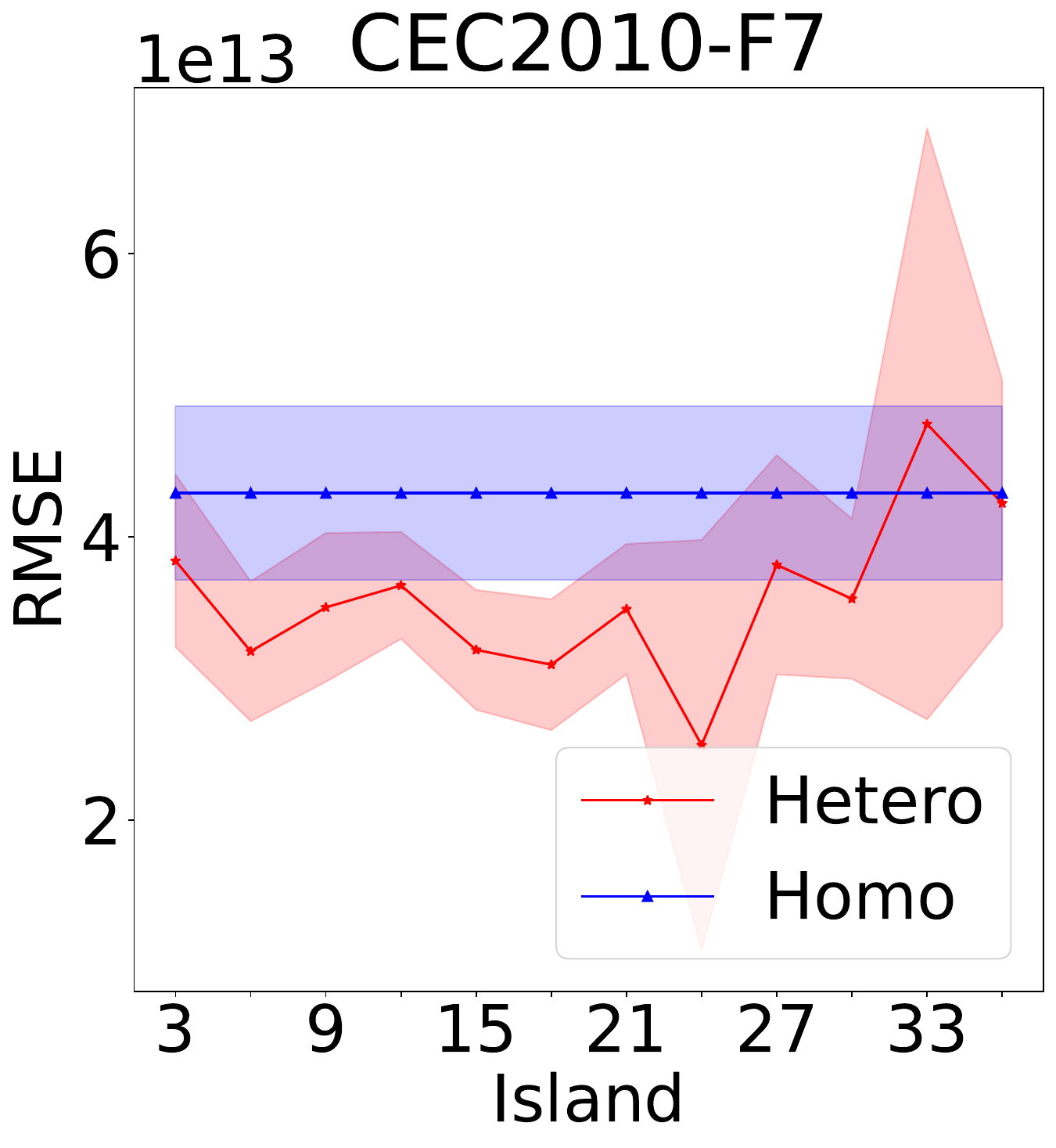}&
        \includegraphics[height=0.15\textwidth, keepaspectratio]{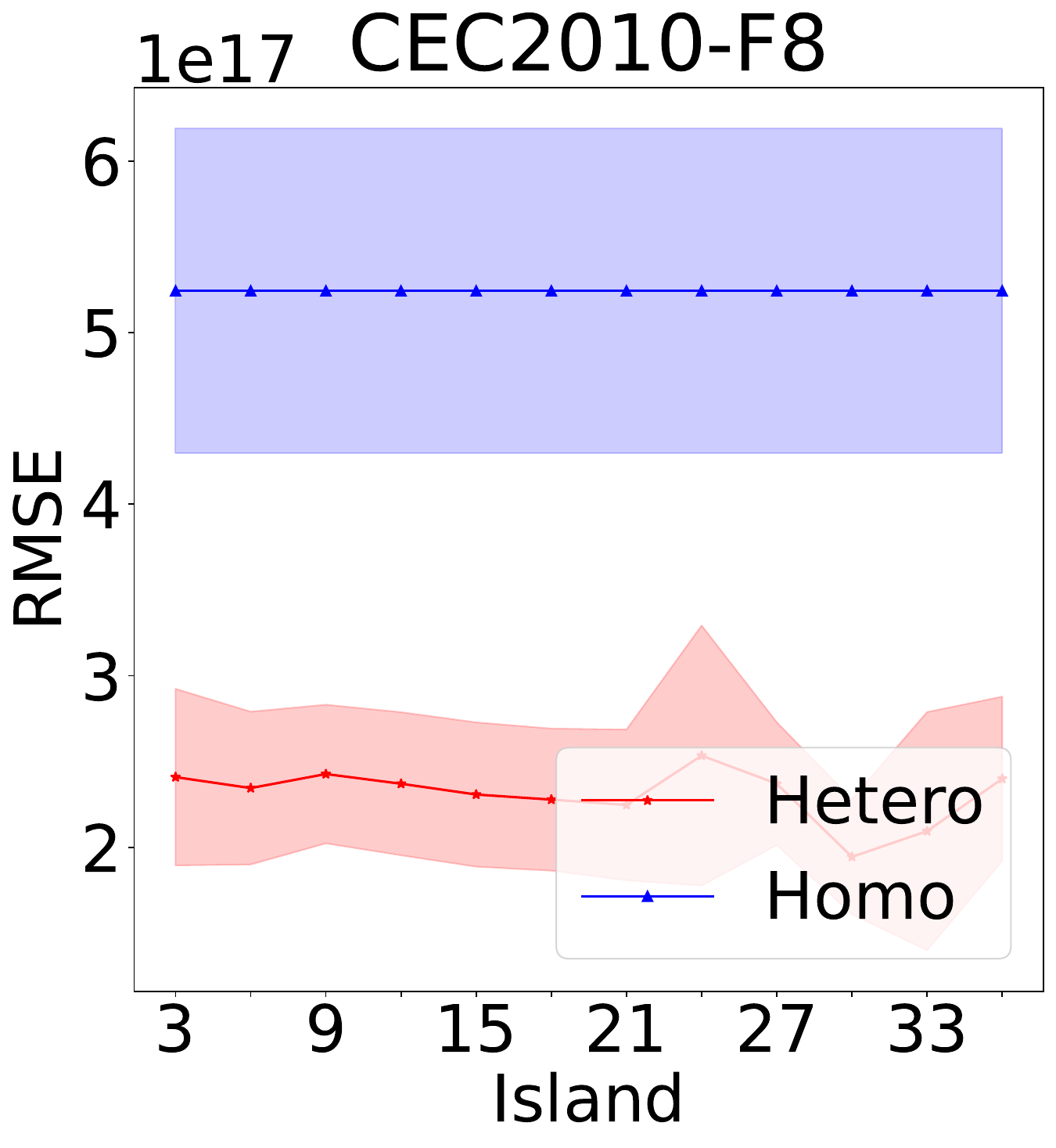}&
        \includegraphics[height=0.15\textwidth, keepaspectratio]{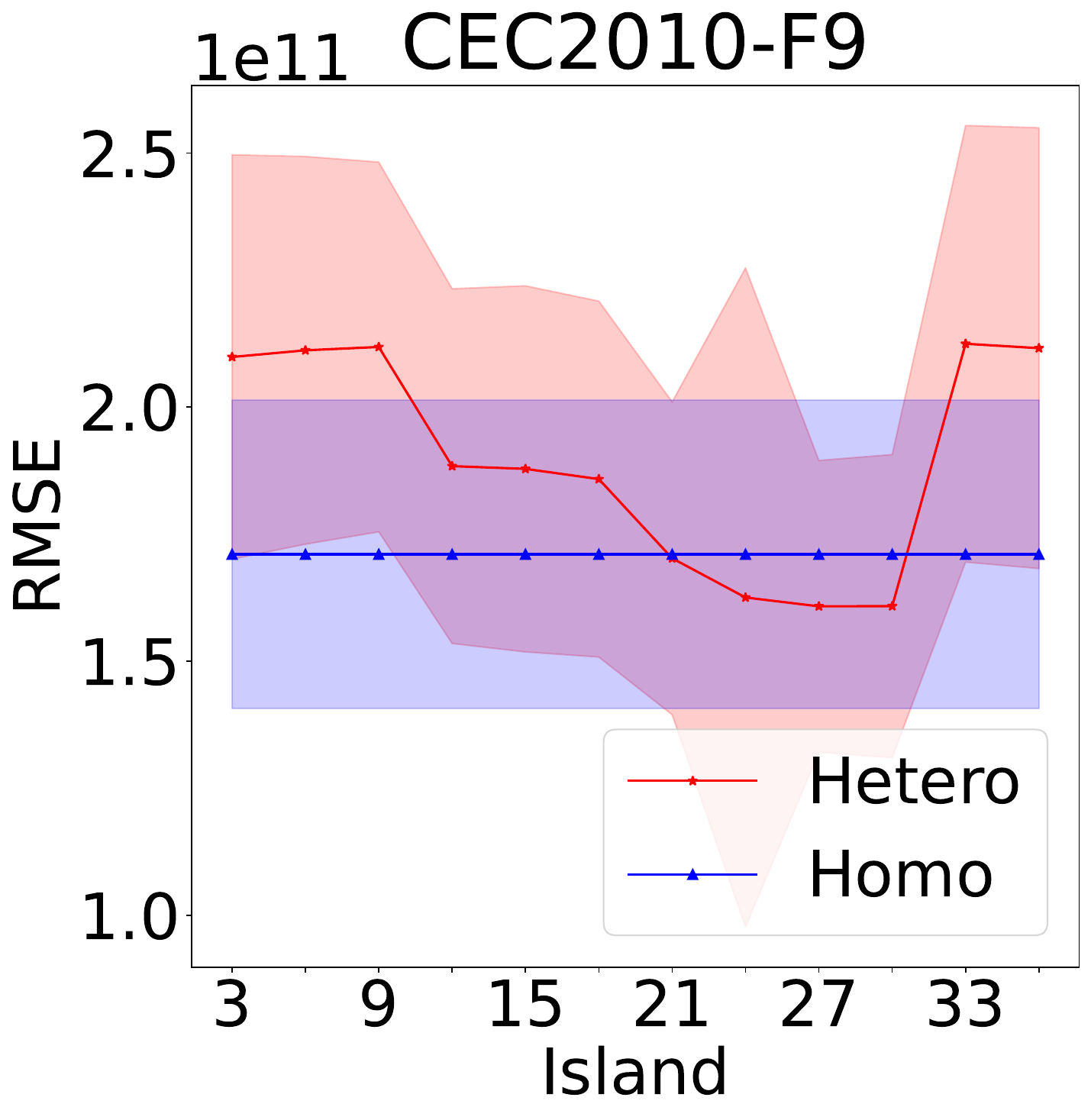}&
        \includegraphics[height=0.15\textwidth, keepaspectratio]{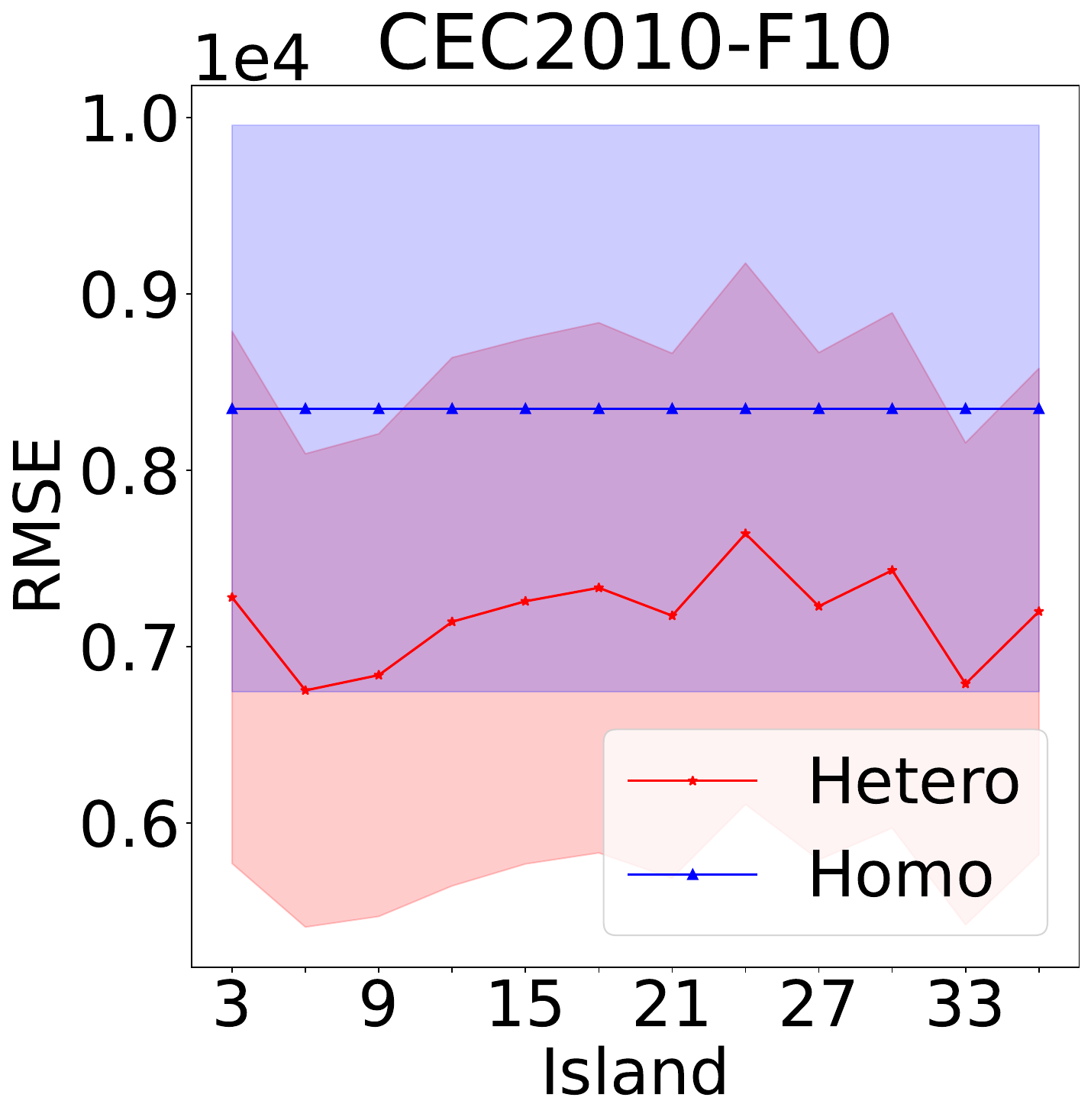}\\
        \includegraphics[height=0.15\textwidth, keepaspectratio]{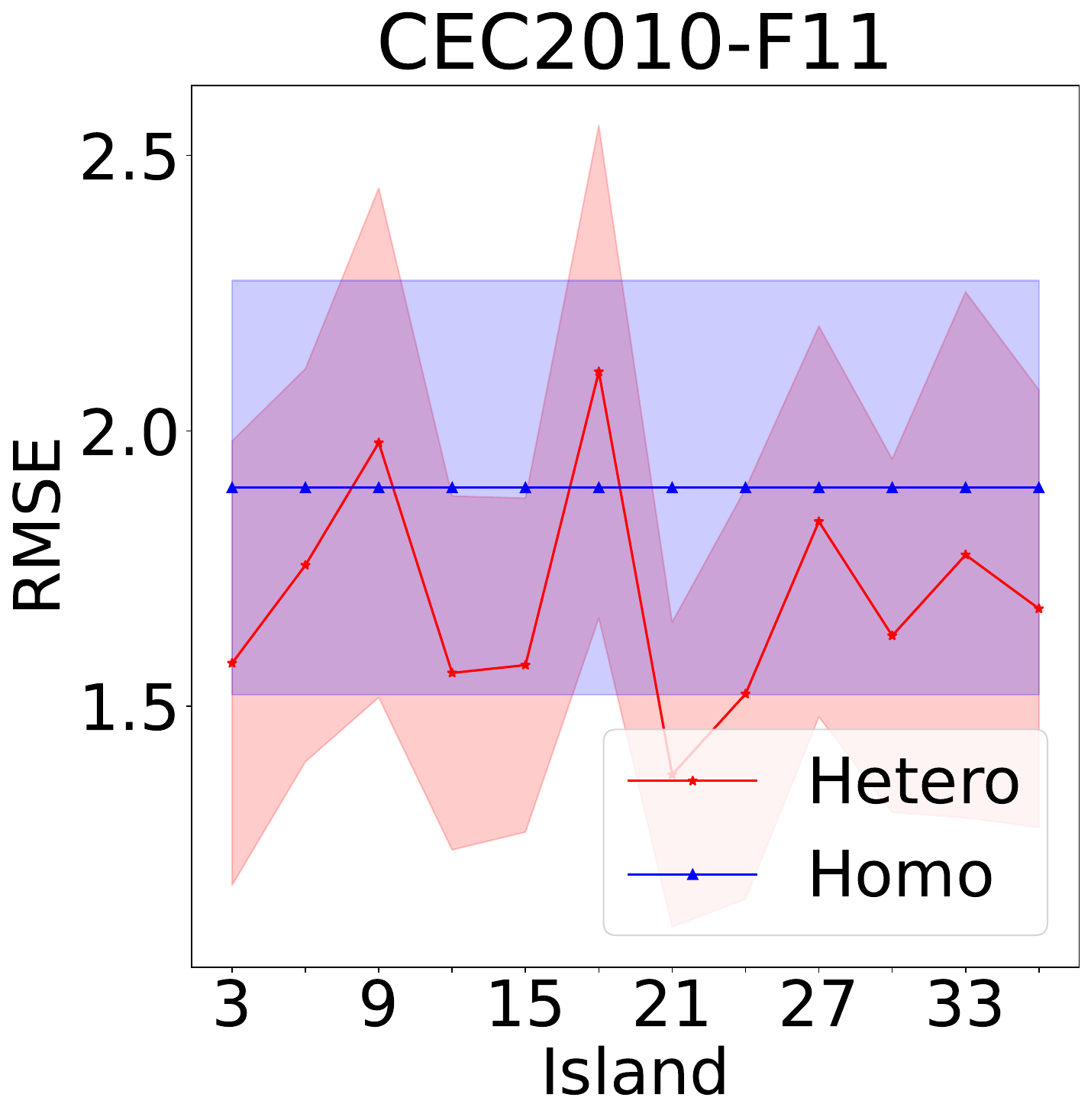}&
        \includegraphics[height=0.15\textwidth, keepaspectratio]{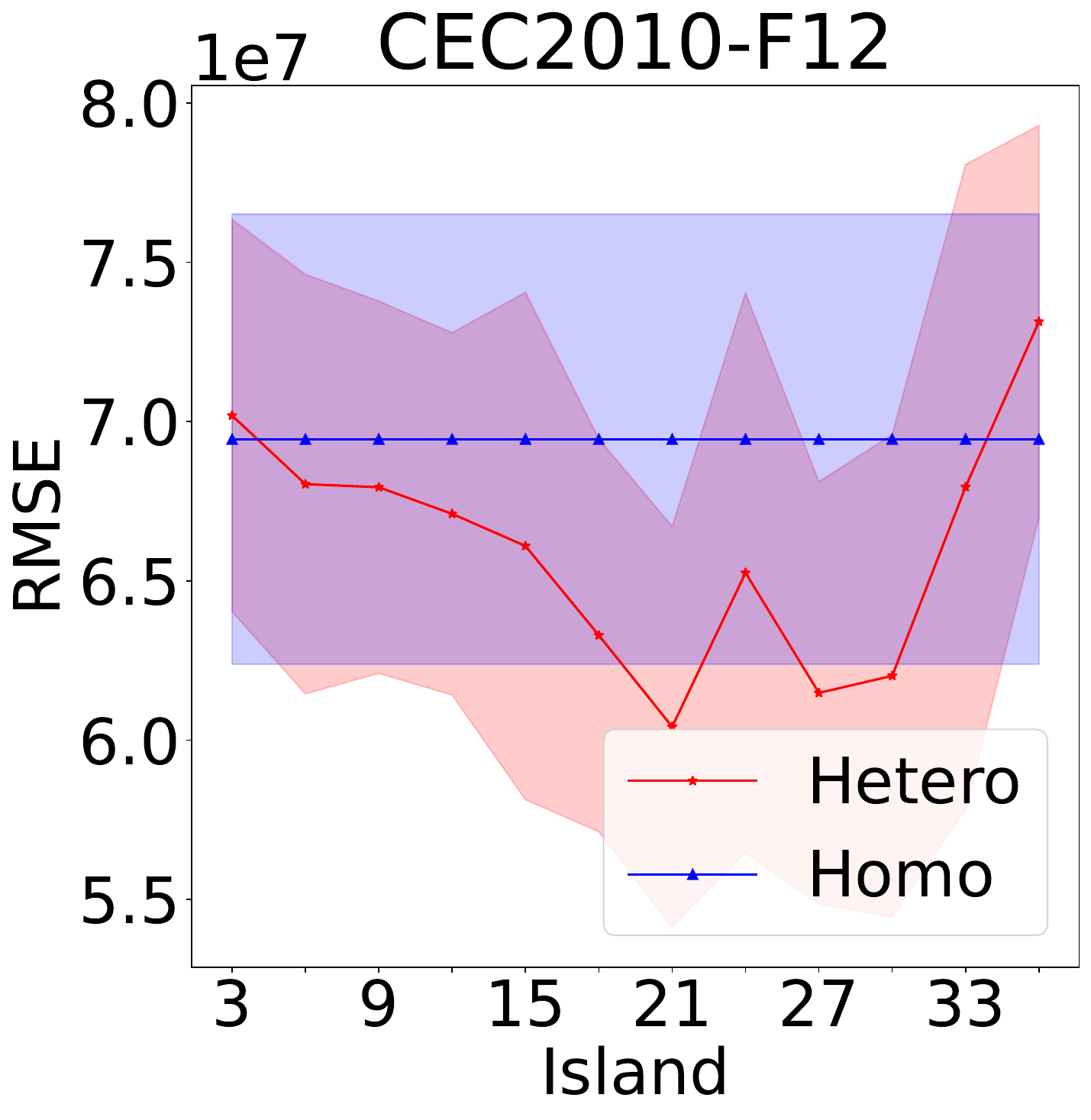}&
        \includegraphics[height=0.15\textwidth, keepaspectratio]{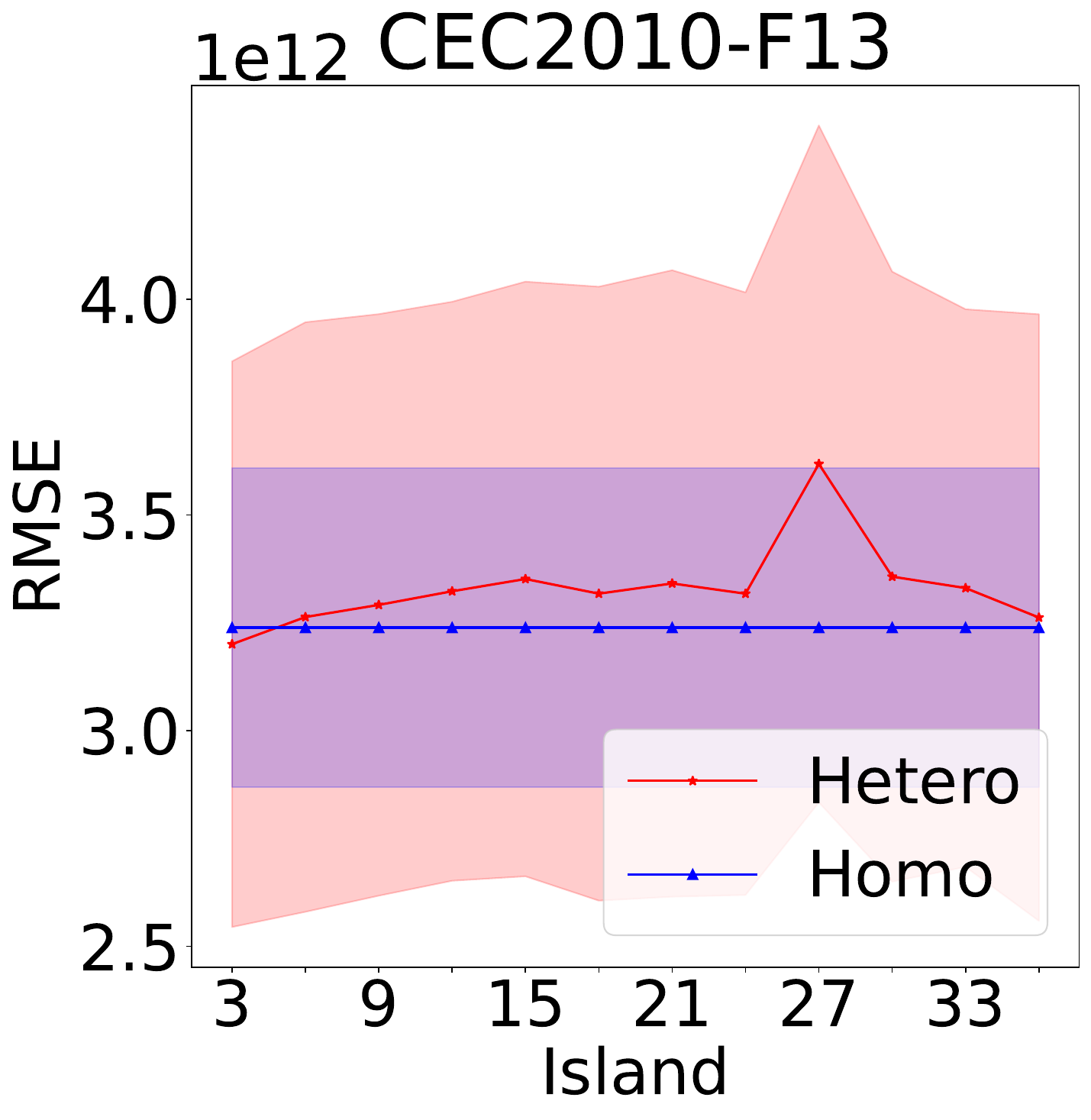}&
        \includegraphics[height=0.15\textwidth, keepaspectratio]{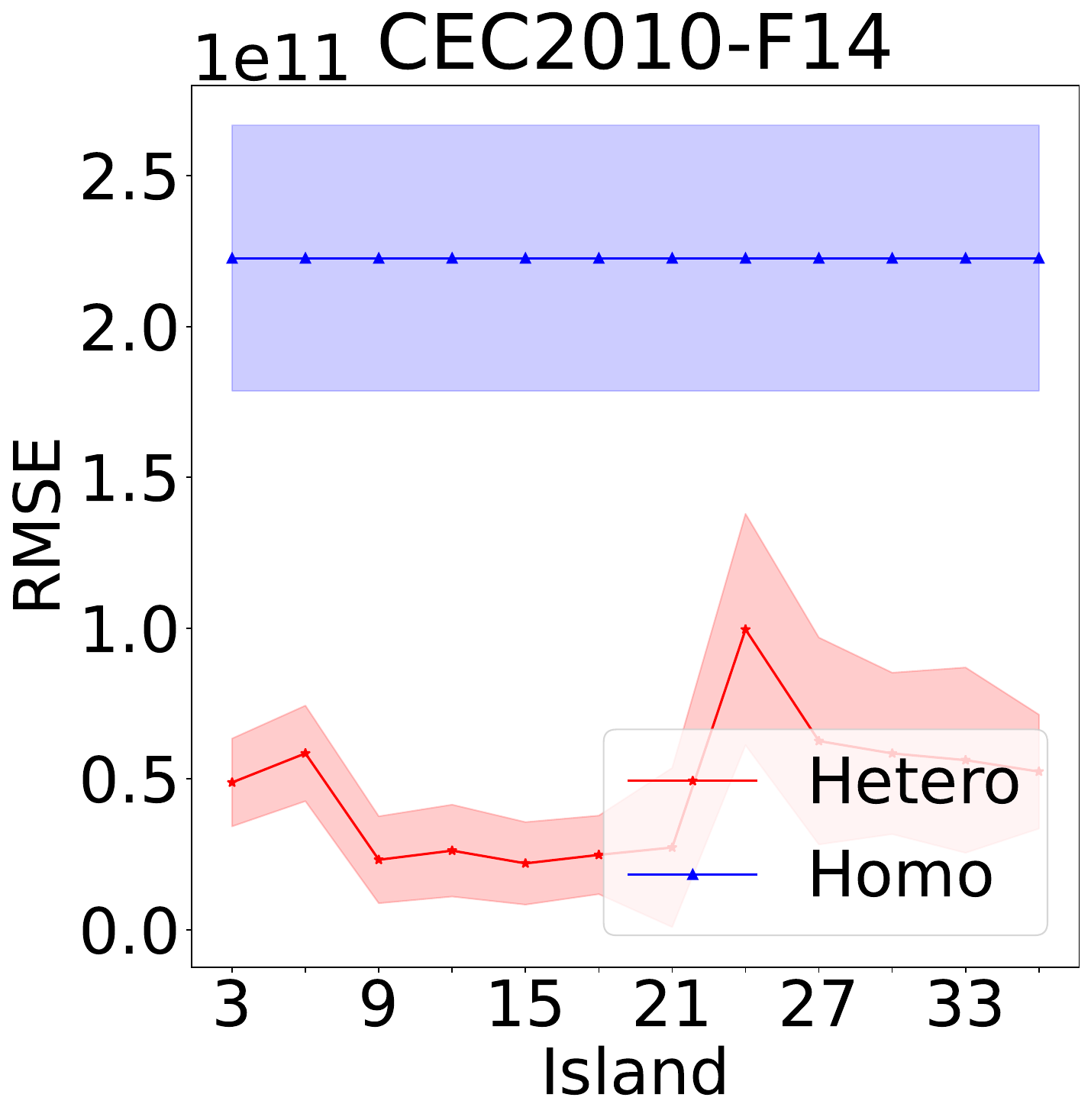}&
        \includegraphics[height=0.15\textwidth, keepaspectratio]{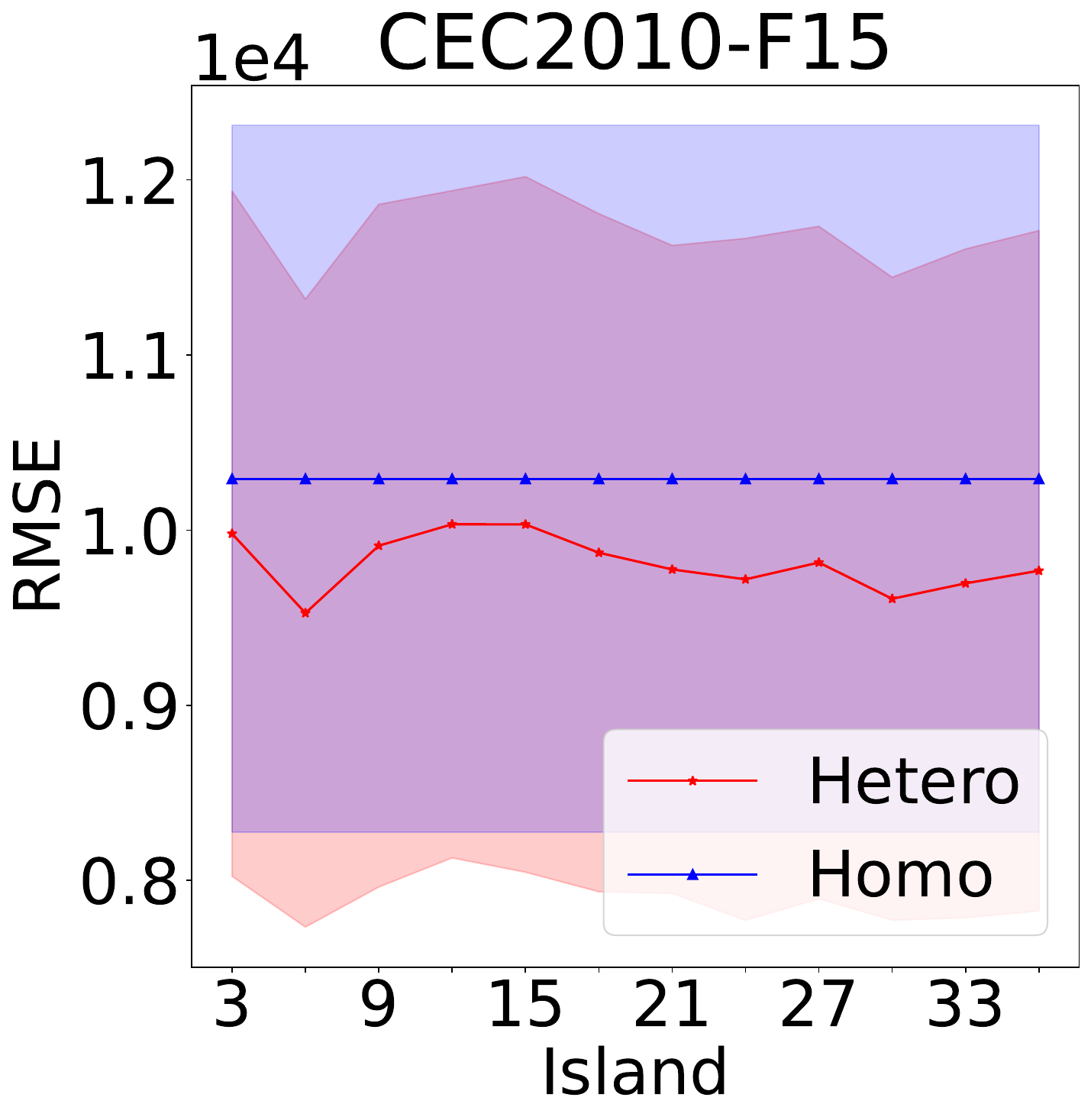}\\
        \includegraphics[height=0.15\textwidth, keepaspectratio]{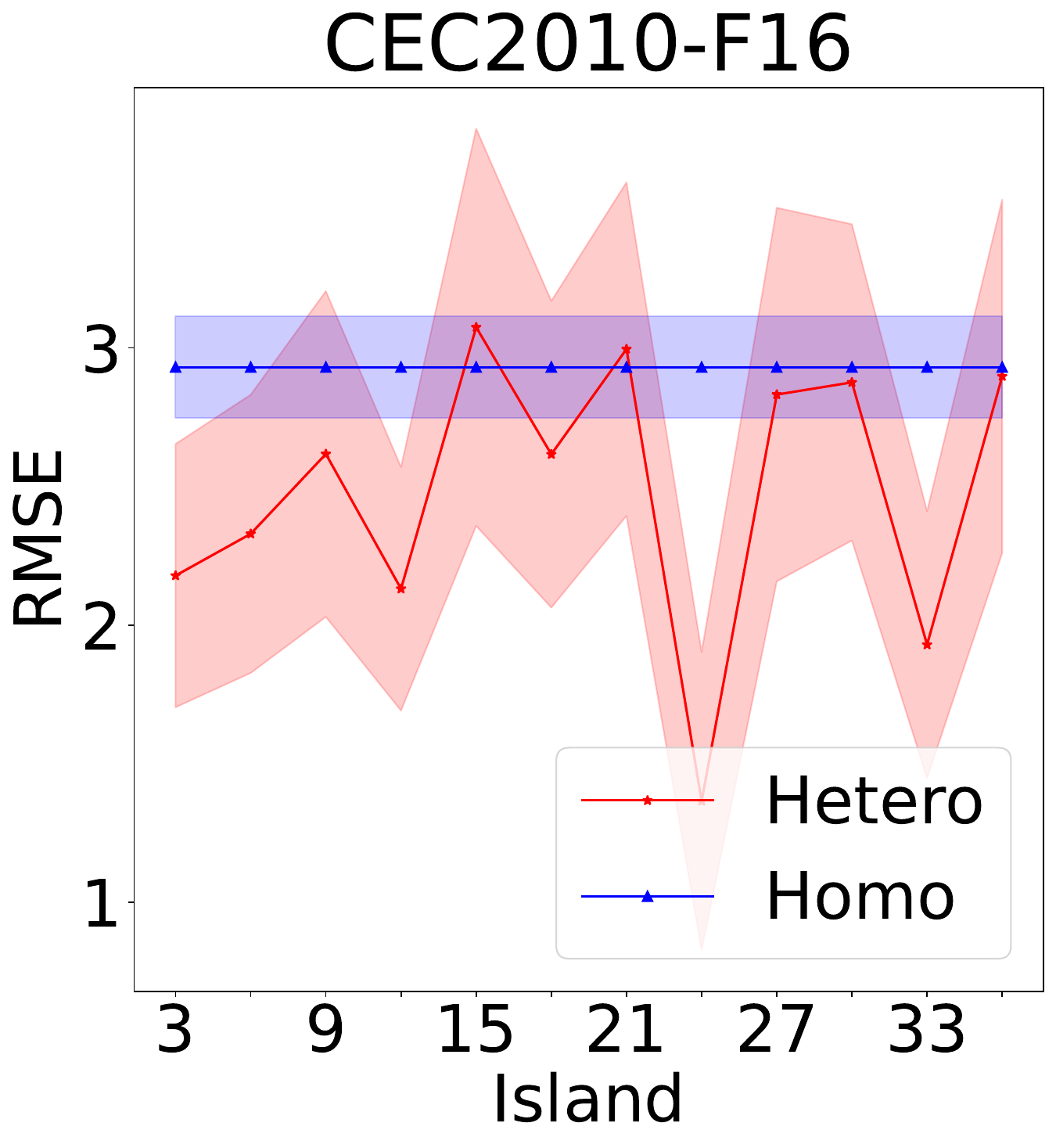}&
        \includegraphics[height=0.15\textwidth, keepaspectratio]{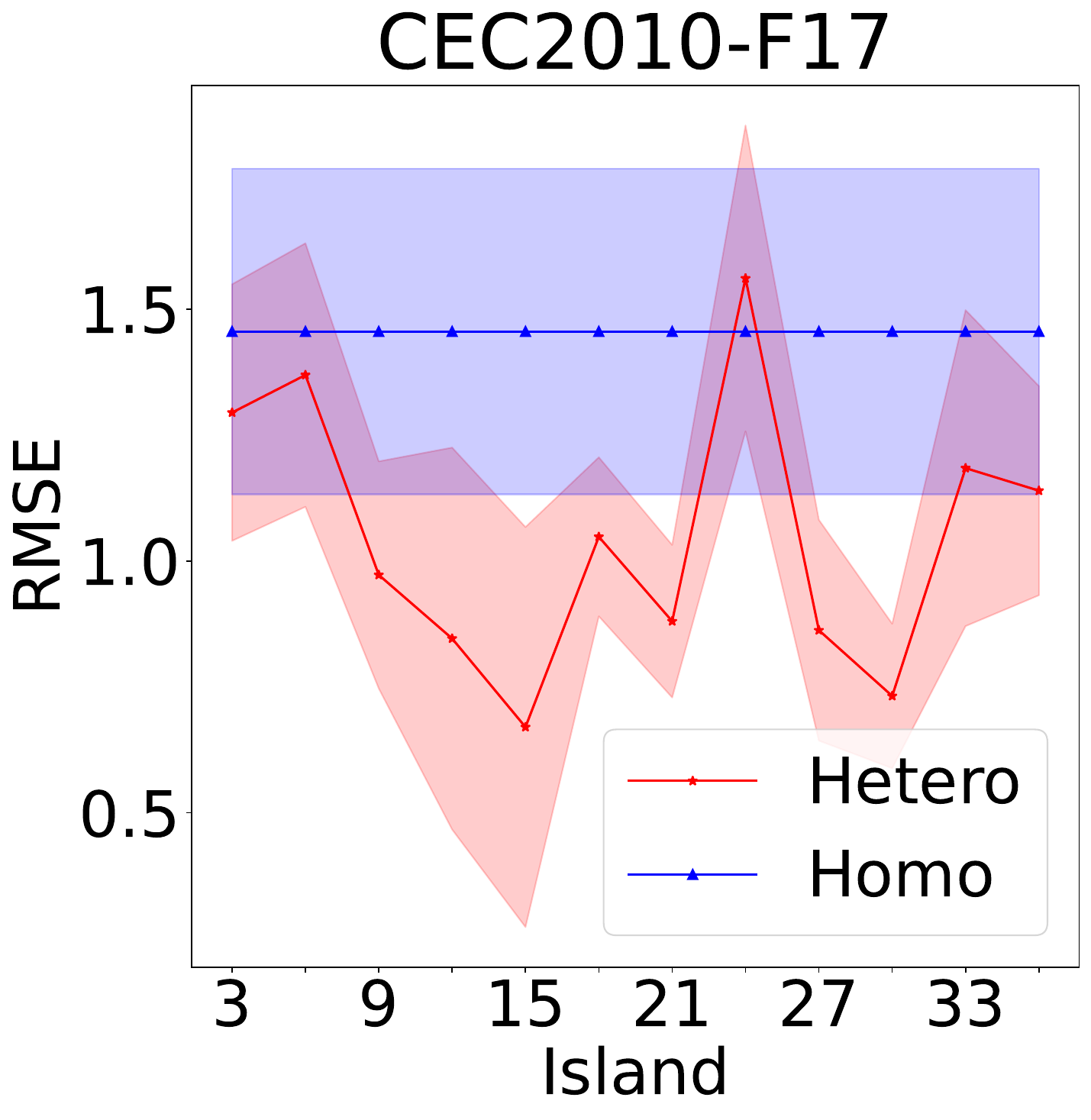}&
        \includegraphics[height=0.15\textwidth, keepaspectratio]{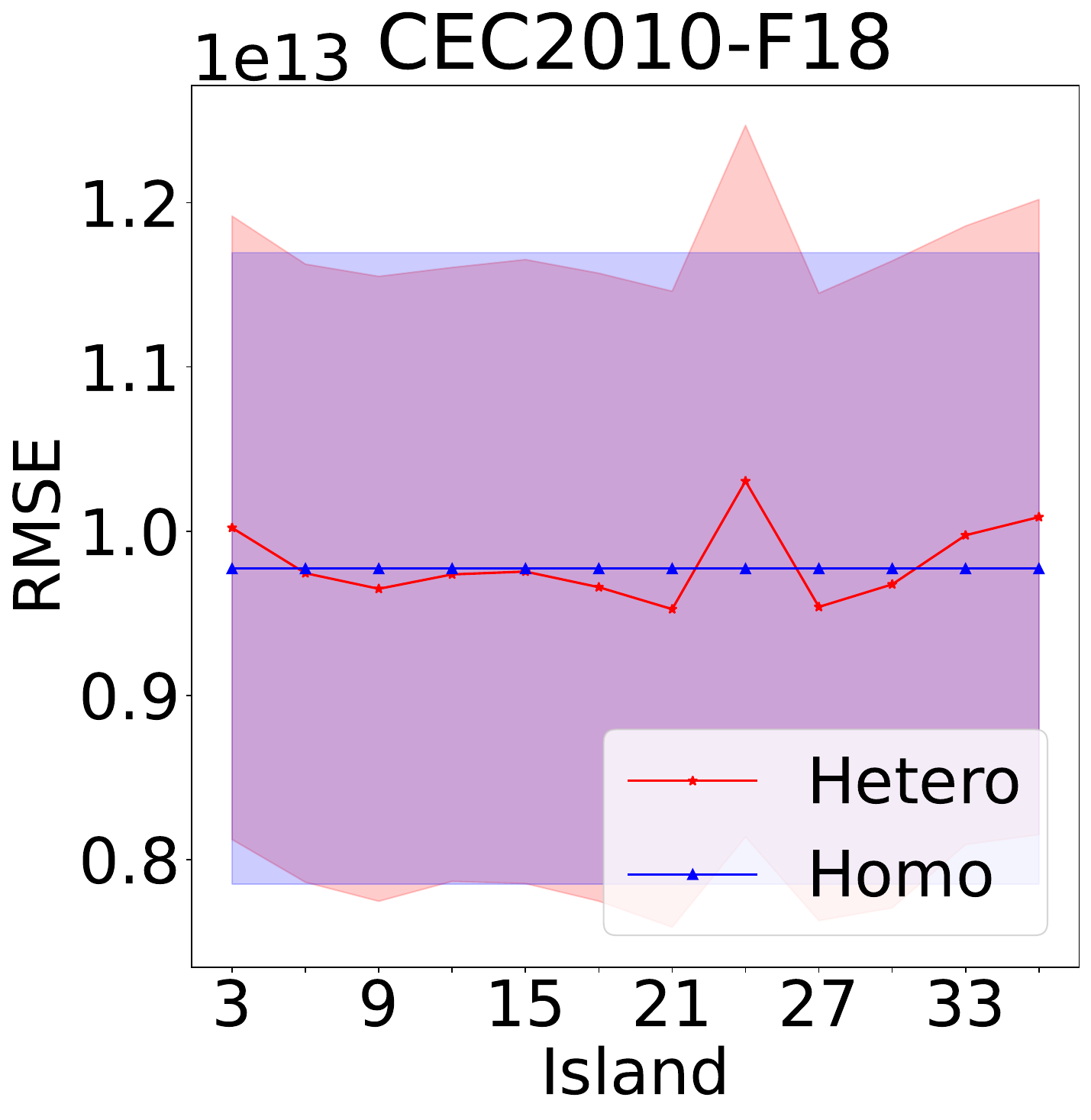}&
        \includegraphics[height=0.15\textwidth, keepaspectratio]{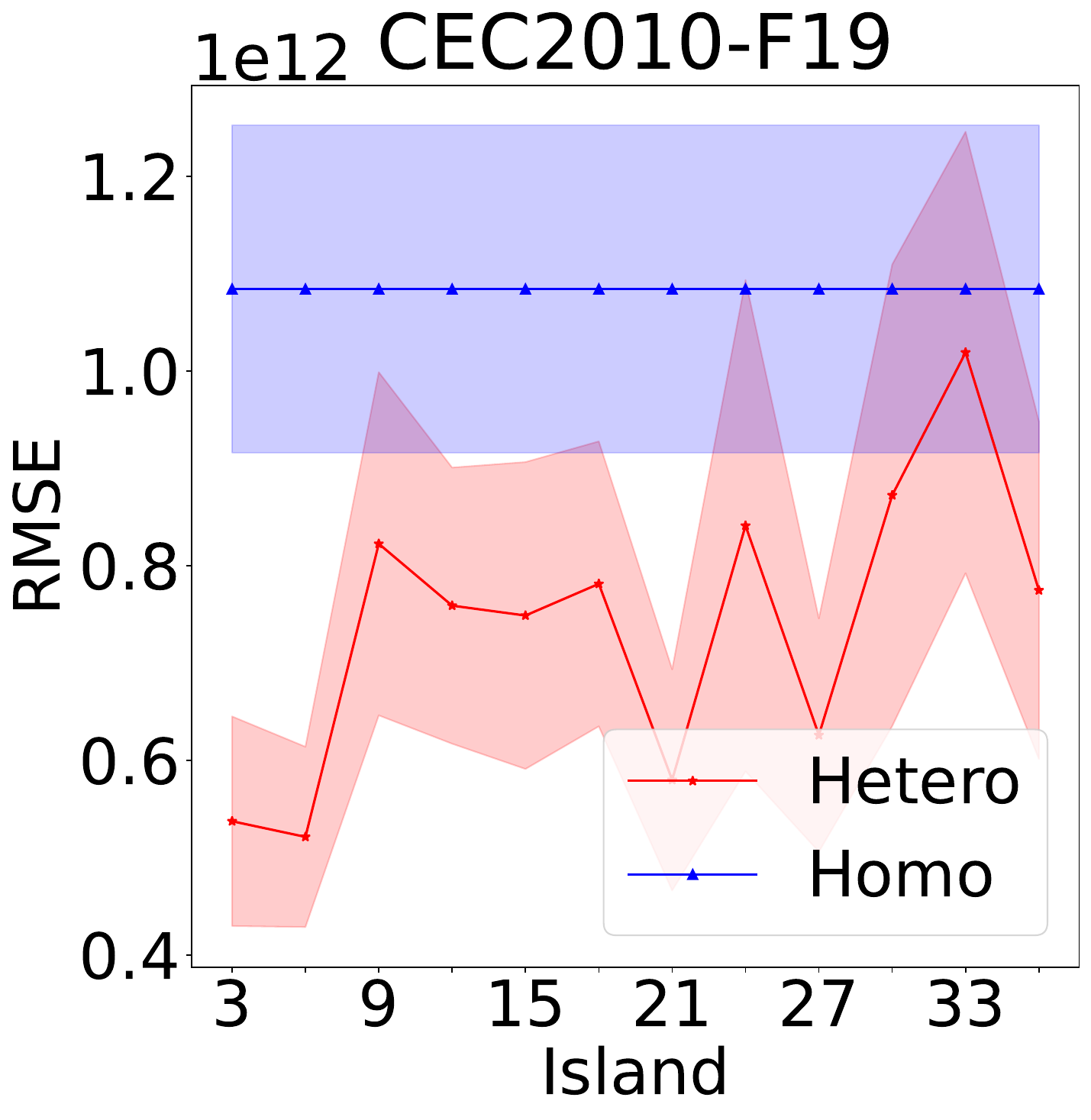}&
        \includegraphics[height=0.15\textwidth, keepaspectratio]{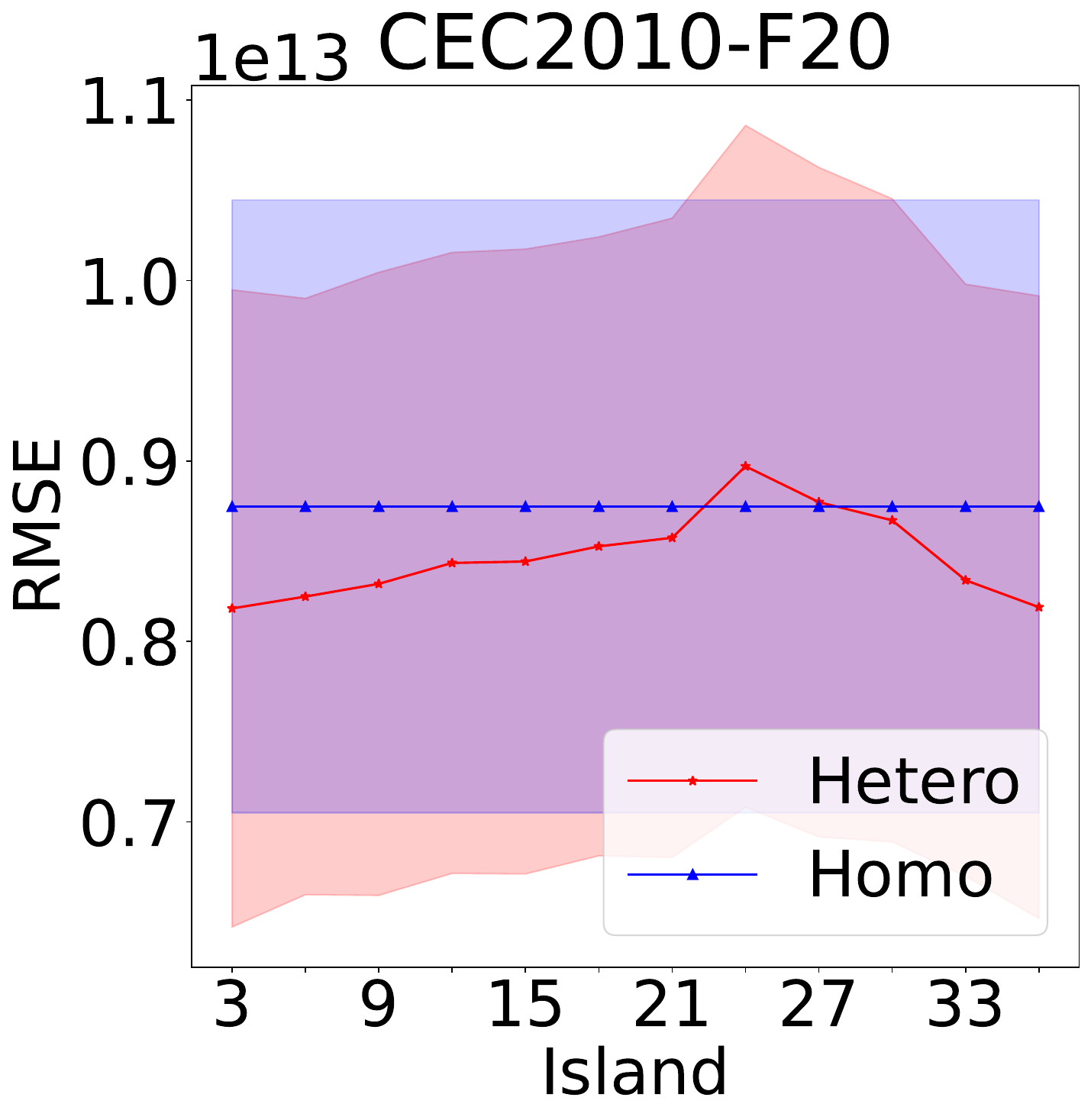}\\
    \end{tabular}
    \caption{
    {
Analyzing the accuracy of our diverse surrogates.}
    }
    \label{fig:homo}
\end{figure}

In our work, to enhance the diversity of searches, we designed diverse surrogate models, each trained on 2/3 of the dataset. However, the reduction of data for each surrogate model inevitably leads to a decline in the performance of individual island's surrogate model. To mitigate this issue, we integrate each island model with models from neighboring islands in a simple ensemble manner. For comparison, the homogeneous model refers to each island training with the full dataset. 
The comparative results, illustrated in \autoref{fig:homo}, underscore the efficacy of our diverse ensemble approach. Although the individual surrogates work with smaller data subsets, the ensemble strategy ensures that the surrogates not only preserve but, in certain cases, even enhance model performance. This approach realizes the dual objectives of maintaining high model accuracy while benefiting from the enriched search diversity.

\subsubsection{Fine-tuning Effectiveness}
\begin{figure}
    \centering
    \includegraphics[width=0.8\linewidth]{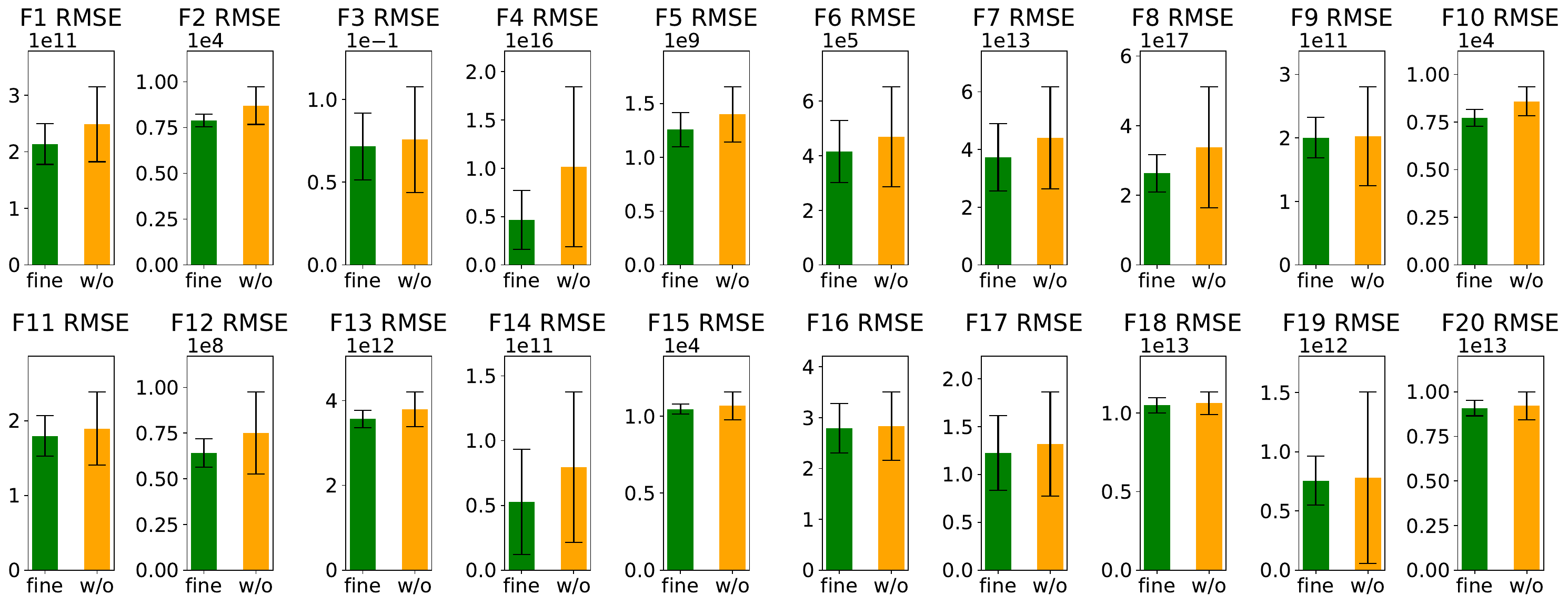}
    \caption{
    {
        Fine-tuning vs. no fine-tuning, comparison of RMSE of all surrogate models.
    }
}
    \label{fig:fintune}
\end{figure}

We compared the RMSE of surrogate models with and without fine-tuning on 20 test functions of CEC2010. Specifically, we took individuals on each island at the 50th generation as the test set. Each surrogate model on 36 islands is associated with an RMSE, and the mean and standard deviation of all RMSEs are reported, represented by a bar graph with error bars. From the experimental results shown in \autoref{fig:fintune}, the RMSE of models without fine-tuning is higher than that of models with fine-tuning, validating the efficacy of our design. 

\subsubsection{Topological Structure Analysis}
\begin{figure}
    \centering
    \includegraphics[width=0.8\linewidth]{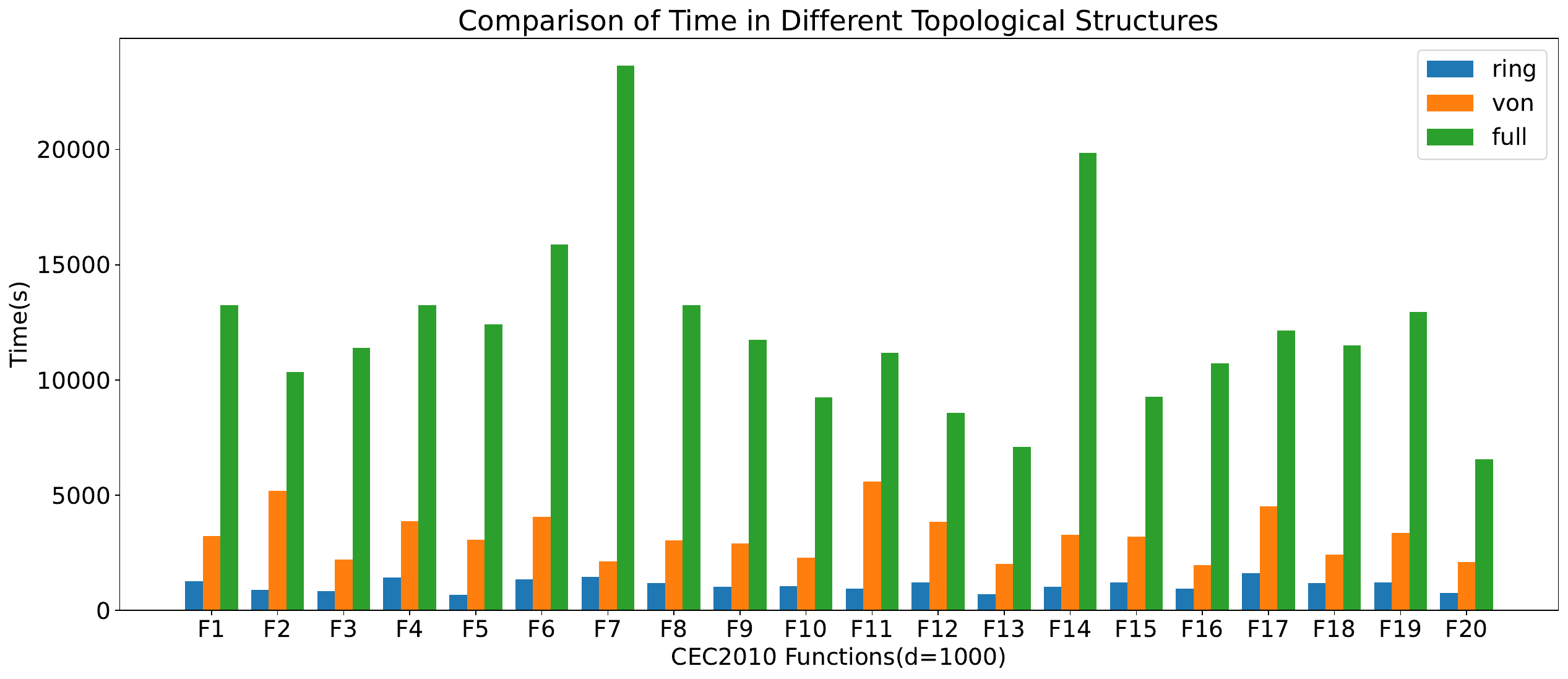}
    \caption{
    {
           Time consumption of three different topological structures on CEC2010 functions for 1000-dimensional problems.
    }}
    \label{fig:topo_time}
\end{figure}

To investigate the impact of topological structures on our algorithm, we compared the performance of algorithms based on three different topological structures: ring, Von Neumann, and fully connected. Next, we analyze from the perspectives of optimization efficacy and computational efficiency. From \autoref{tab:Fit},  it can be seen that under the three topological structures, the ring structure performs the worst, primarily due to its slower information exchange which hampers population convergence. Both the Von Neumann and fully connected typologies exhibit competitive performance, and their differences are not so significant. As shown in \autoref{fig:topo_time}, as expected, the Von Neumann topology demonstrates superior time efficiency in reaching optimal solutions compared to the fully connected topology. Given these findings, we adopt the Von Neumann topology for its balanced performance in both optimization results and computational efficiency. 

 \begin{table}
\centering
\caption{
Comparison of the optimization performance of three different topological structures on CEC2010 functions
}
\resizebox{0.45\linewidth}{!}{

\begin{tabular}{|c|c|c|c|}
\hline
Benchmark & HSKT-DDEA-ring & HSKT-DDEA-von & HSKT-DDEA-full \\\hline
F1 & 1.91E+11$\pm$5.32E+08($\approx$) & 1.61E+11$\pm$8.47E+09 & \textbf{1.59E+11$\pm$8.06E+09($\approx$)} \\\hline
F2 & 1.71E+04$\pm$1.87E+02($\approx$) & \textbf{1.70E+04$\pm$1.16E+02} & 1.70E+04$\pm$1.77E+02($\approx$) \\\hline
F3 & 2.10E+01$\pm$1.93E-02($\approx$) & 2.10E+01$\pm$2.69E-02 & \textbf{2.10E+01$\pm$9.47E-03($\approx$)} \\\hline
F4 & \textbf{1.71E+16$\pm$1.45E+16($\approx$)} & 2.08E+16$\pm$1.64E+16 & 2.98E+16$\pm$3.10E+16($\approx$) \\\hline
F5 & 8.35E+08$\pm$4.36E+07($\approx$) & \textbf{7.59E+08$\pm$8.02E+07} & 7.64E+08$\pm$7.63E+07($\approx$) \\\hline
F6 & 2.07E+07$\pm$3.20E+05($\approx$) & 2.06E+07$\pm$4.13E+05 & \textbf{2.06E+07$\pm$5.79E+05($\approx$)} \\\hline
F7 & 1.78E+14$\pm$1.17E+14($\approx$) & 2.16E+14$\pm$6.87E+13 & \textbf{1.36E+14$\pm$1.90E+13($\approx$)} \\\hline
F8 & 2.71E+16$\pm$4.16E+15($\approx$) & 2.44E+16$\pm$4.20E+15 & \textbf{2.36E+16$\pm$3.81E+15($\approx$)} \\\hline
F9 & 2.06E+11$\pm$9.00E+09($\approx$) & 2.02E+11$\pm$8.68E+09 & \textbf{2.00E+11$\pm$7.06E+09($\approx$)} \\\hline
F10 & \textbf{1.74E+04$\pm$1.73E+02($\approx$)} & 1.74E+04$\pm$1.51E+02 & 1.74E+04$\pm$2.26E+02($\approx$) \\\hline
F11 & 2.31E+02$\pm$3.06E-01($\approx$) & \textbf{2.31E+02$\pm$2.76E-01} & 2.31E+02$\pm$1.36E-01($\approx$) \\\hline
F12 & 2.07E+07$\pm$3.47E+07($\approx$) & 2.70E+07$\pm$4.74E+07 & \textbf{3.83E+06$\pm$1.13E+06($\approx$)} \\\hline
F13 & 1.64E+12$\pm$1.20E+09(+) & \textbf{6.32E+11$\pm$1.15E+10} & 6.33E+11$\pm$9.18E+09($\approx$) \\\hline
F14 & 2.29E+11$\pm$9.92E+09($\approx$) & 2.22E+11$\pm$1.82E+10 & \textbf{2.17E+11$\pm$1.48E+10($\approx$)} \\\hline
F15 & 1.17E+05$\pm$4.60E-01($\approx$) & 1.72E+04$\pm$1.81E+02 & \textbf{1.70E+04$\pm$9.85E+01($\approx$)} \\\hline
F16 & \textbf{4.20E+02$\pm$1.54E-01($\approx$)} & 4.20E+02$\pm$5.17E-01 & 4.20E+02$\pm$6.62E-01($\approx$) \\\hline
F17 & 4.34E+02$\pm$1.76E-01($\approx$) & 4.34E+02$\pm$2.94E-01 & \textbf{4.34E+02$\pm$1.45E-01($\approx$)} \\\hline
F18 & 1.40E+12$\pm$6.48E+09($\approx$) & 1.39E+12$\pm$7.90E+09 & \textbf{1.38E+12$\pm$4.89E+09($\approx$)} \\\hline
F19 & 1.27E+12$\pm$1.38E+12($\approx$) & 7.93E+11$\pm$1.19E+12 & \textbf{2.78E+11$\pm$1.48E+11($\approx$)} \\\hline
F20 & 1.16E+13$\pm$3.93E+09(+) & 1.58E+12$\pm$1.05E+10 & \textbf{1.57E+12$\pm$8.43E+08($\approx$)} \\\hline
+/$\approx$/- & 2/18/0 & NA & 0/20/0 \\\hline
Average Rank & 2.50 & 2.00 & 1.50 \\\hline
\end{tabular}

}

\label{tab:Fit}
\end{table}

\subsubsection{Convergence Analysis}
\begin{figure}
\centering
\subfigure{\includegraphics[height=0.15\textheight]{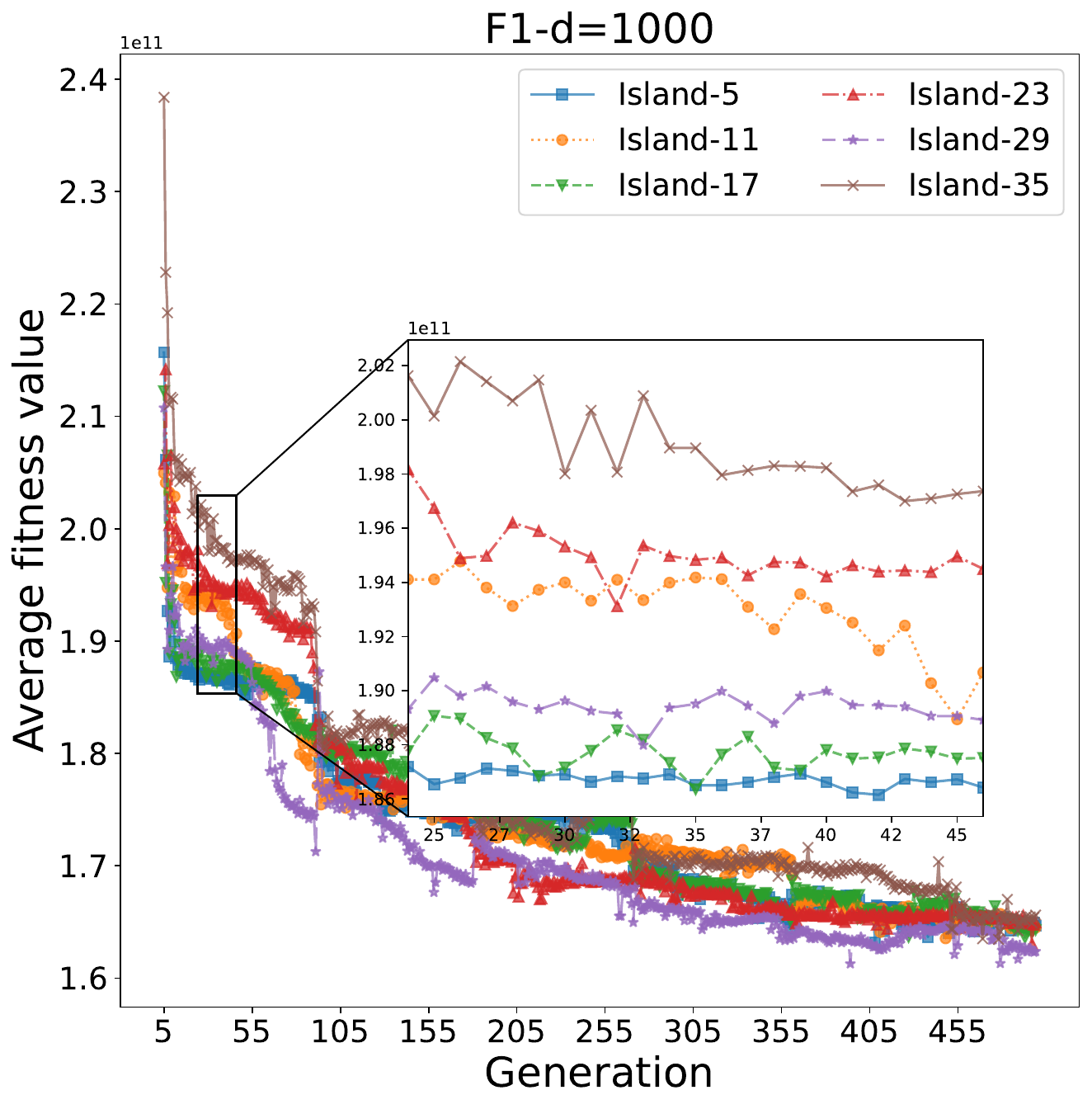}}
\subfigure{\includegraphics[height=0.15\textheight]{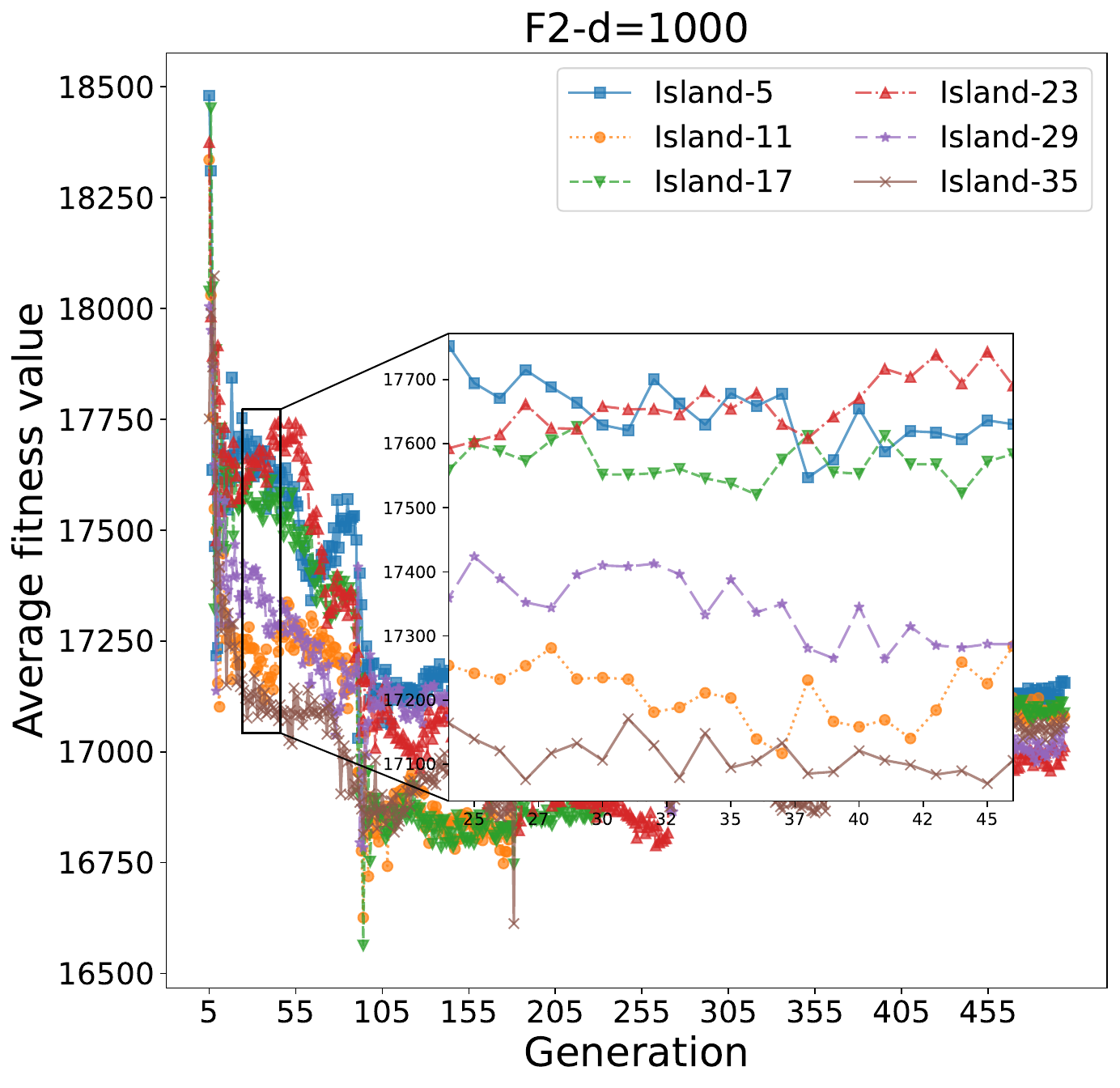}}
\subfigure{\includegraphics[height=0.15\textheight]{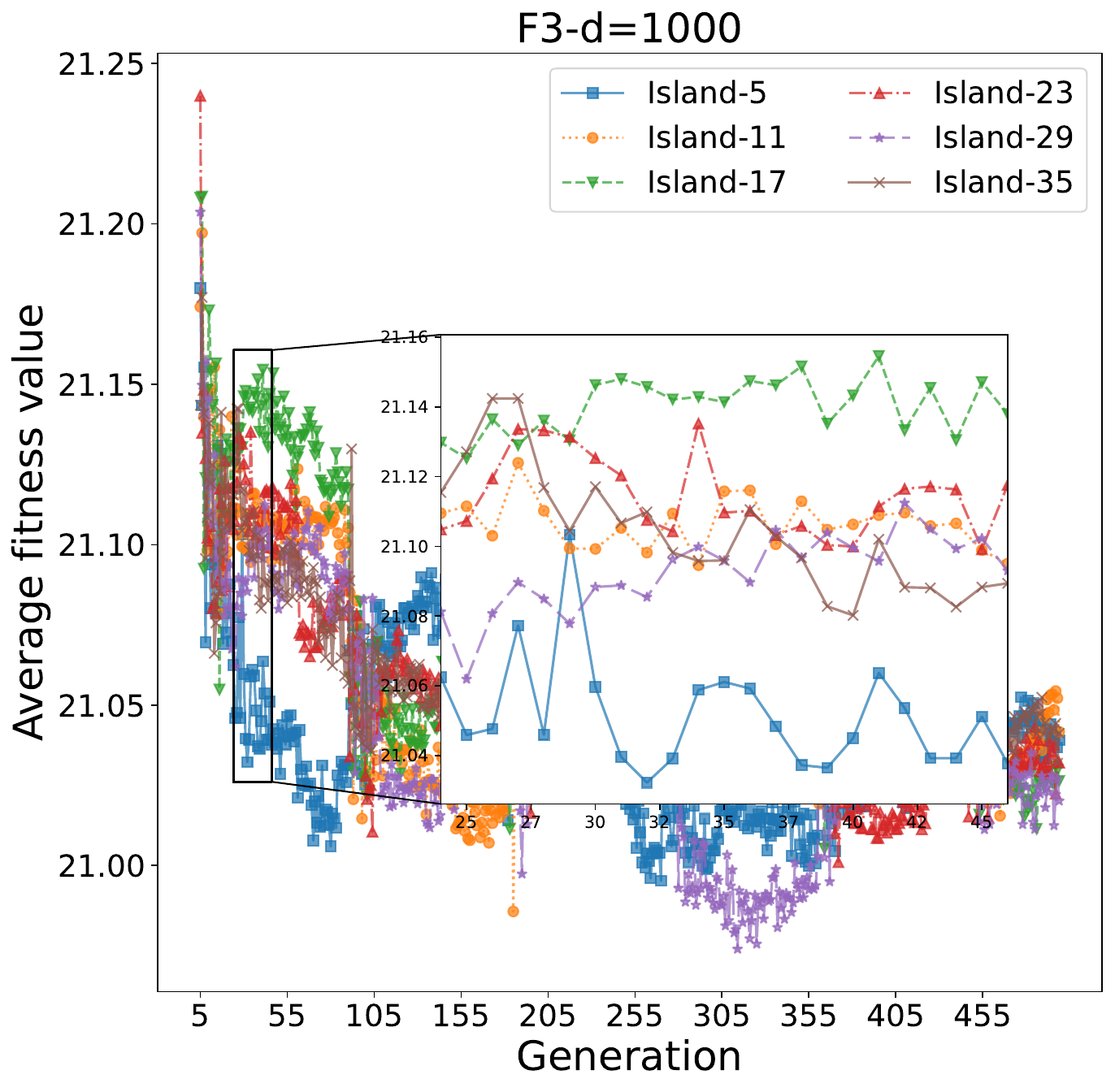}}
\subfigure{\includegraphics[height=0.15\textheight]{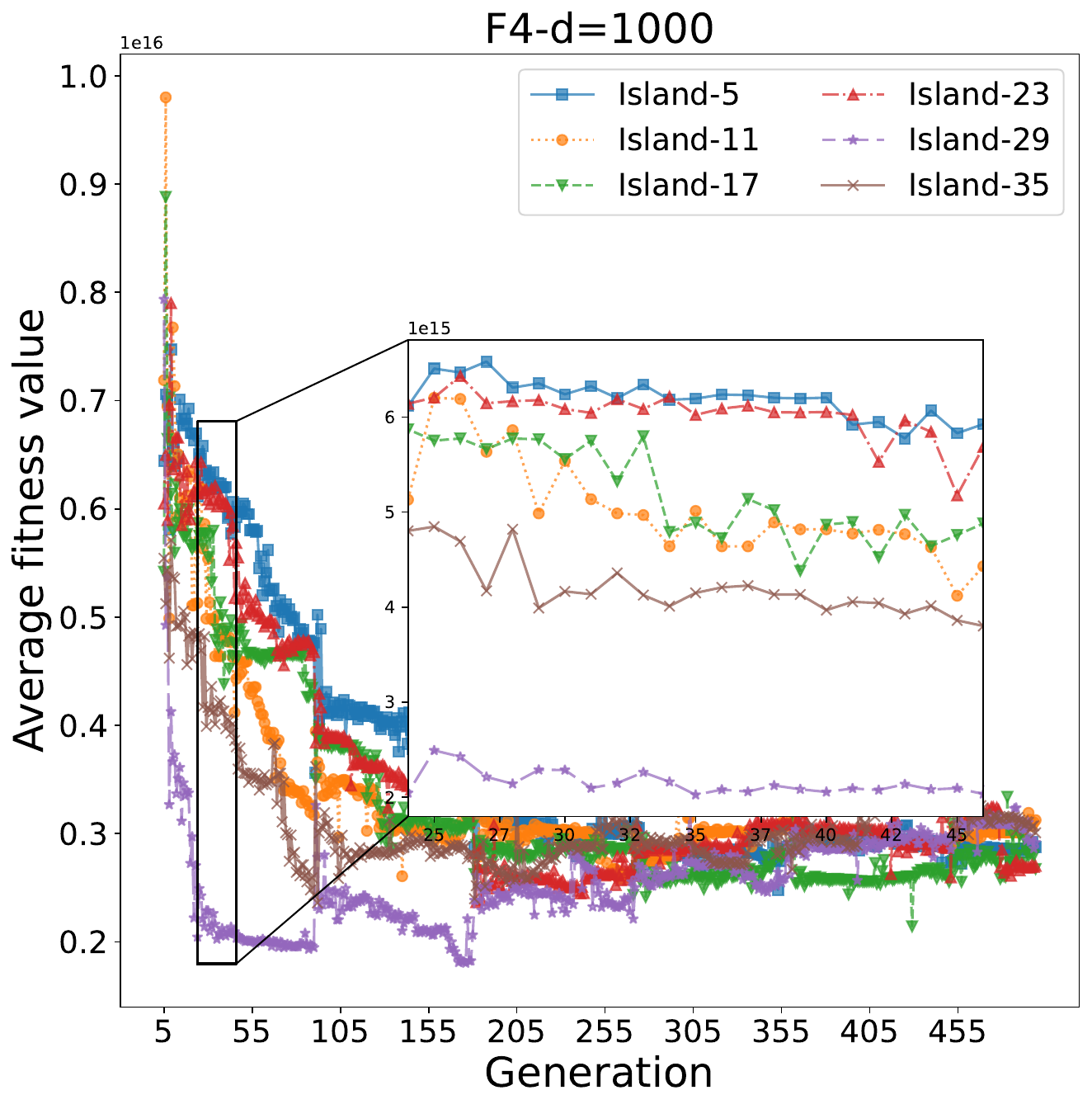}}
\subfigure{\includegraphics[height=0.15\textheight]{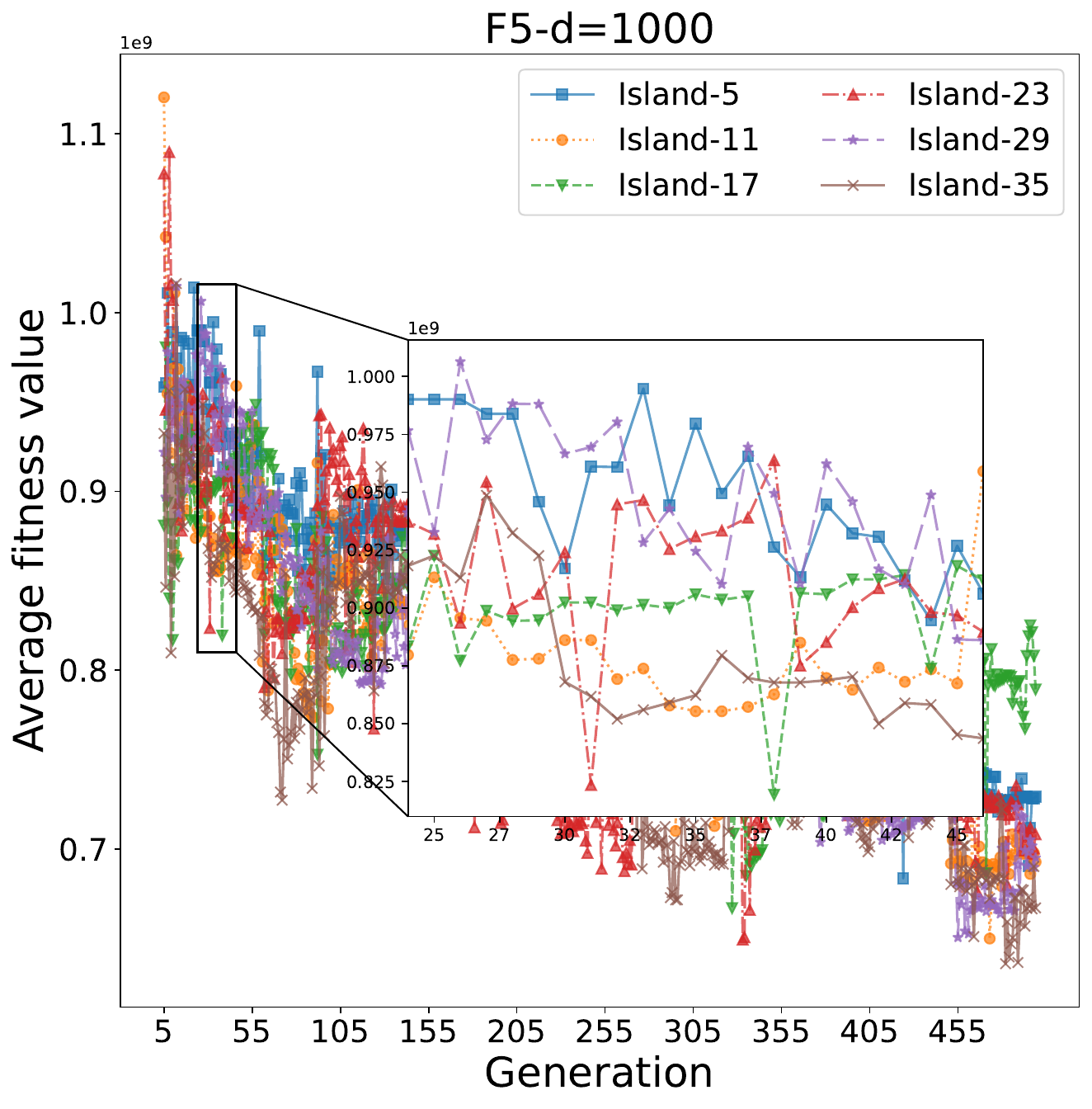}}
\subfigure{\includegraphics[height=0.15\textheight]{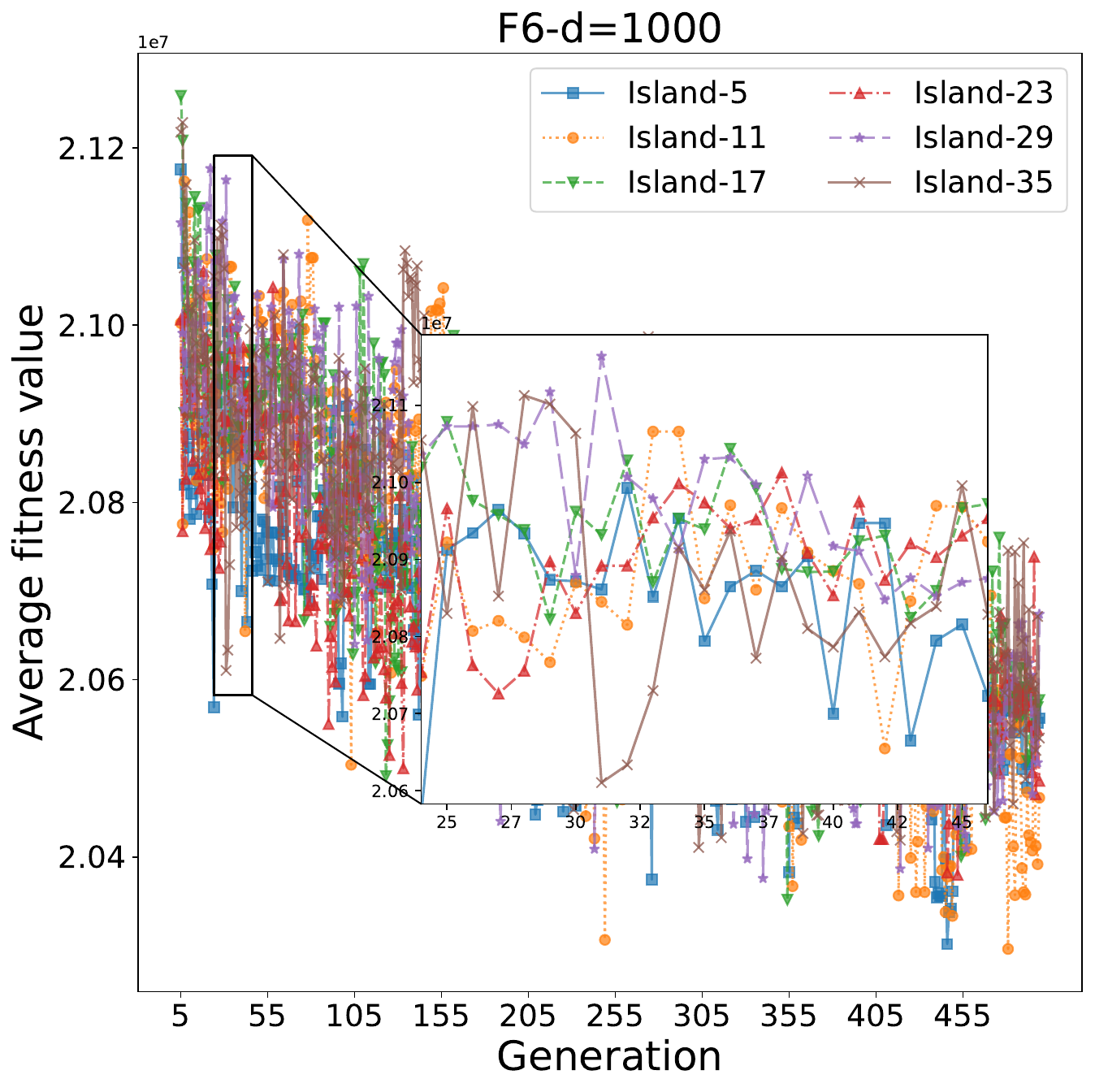}}
\subfigure{\includegraphics[height=0.15\textheight]{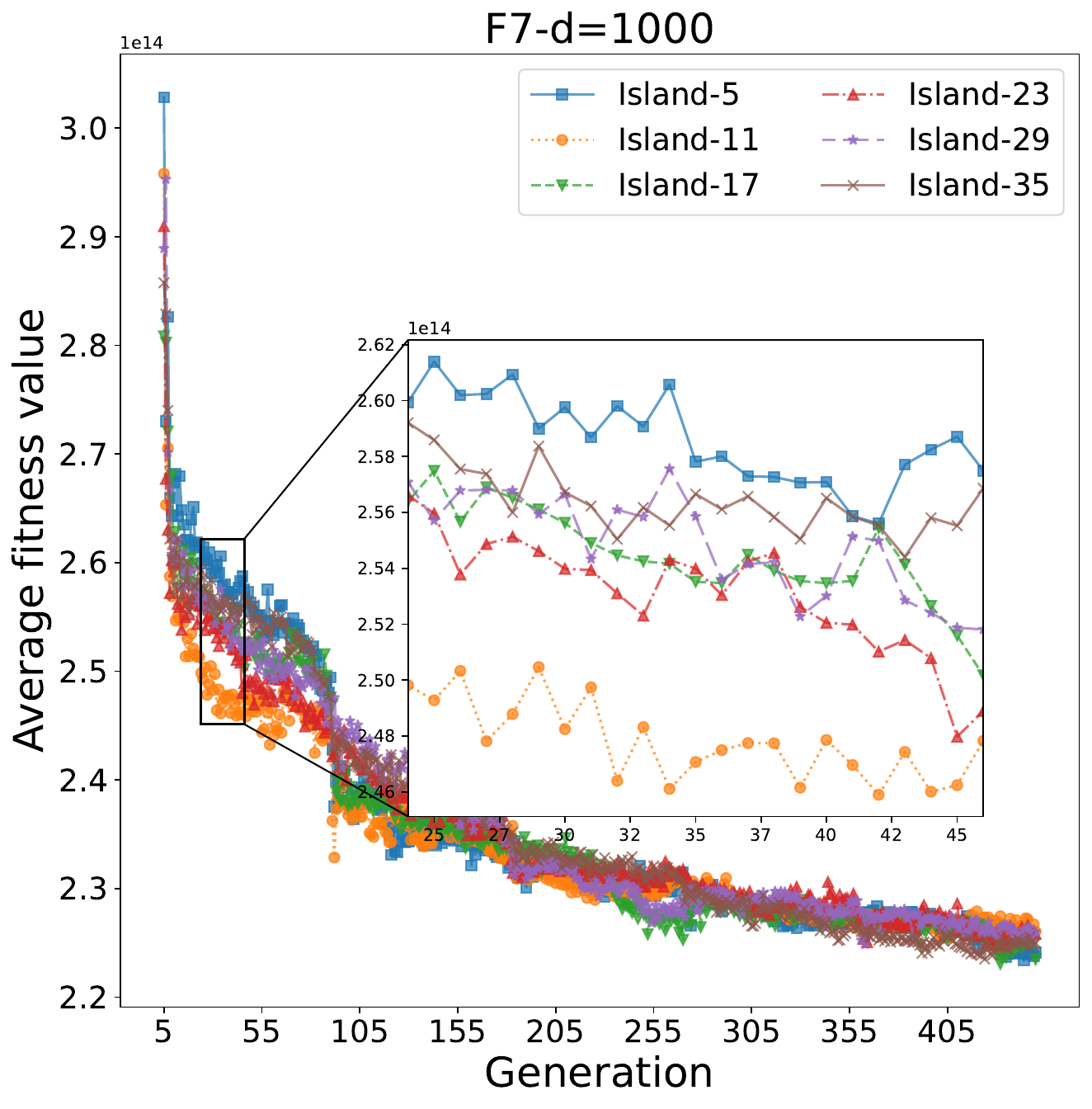}}
\subfigure{\includegraphics[height=0.15\textheight]{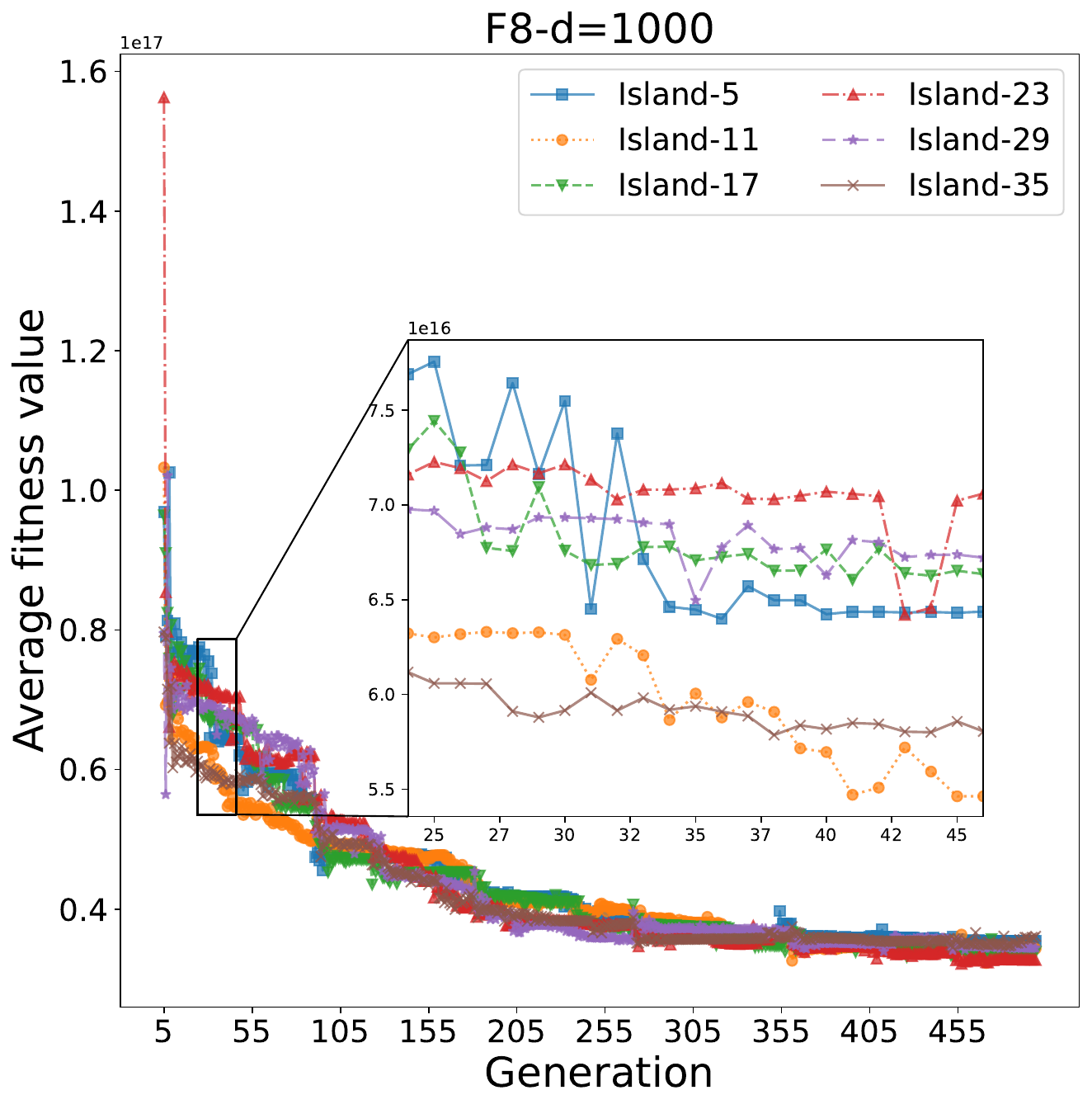}}
\subfigure{\includegraphics[height=0.15\textheight]{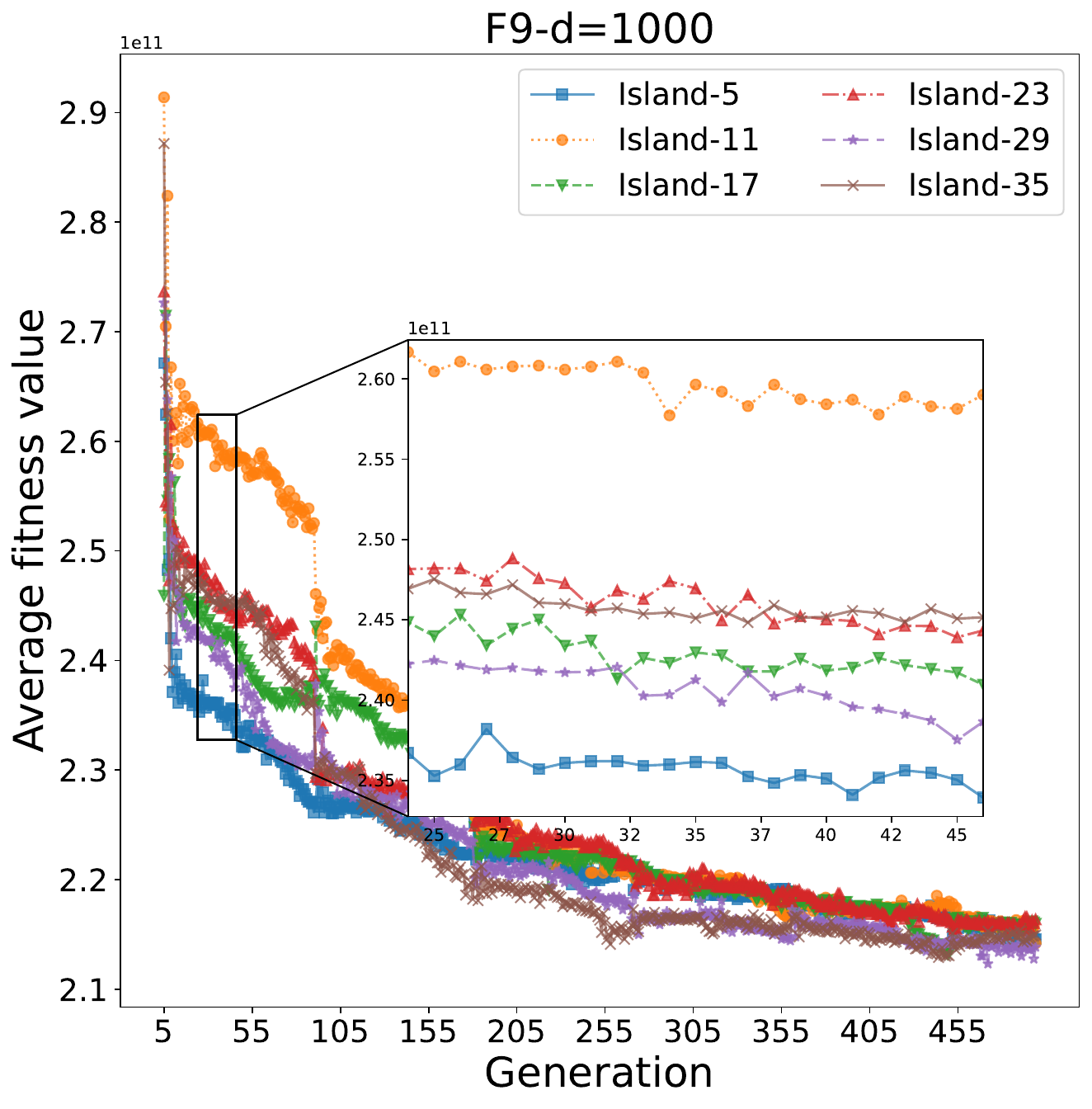}}
\subfigure{\includegraphics[height=0.15\textheight]{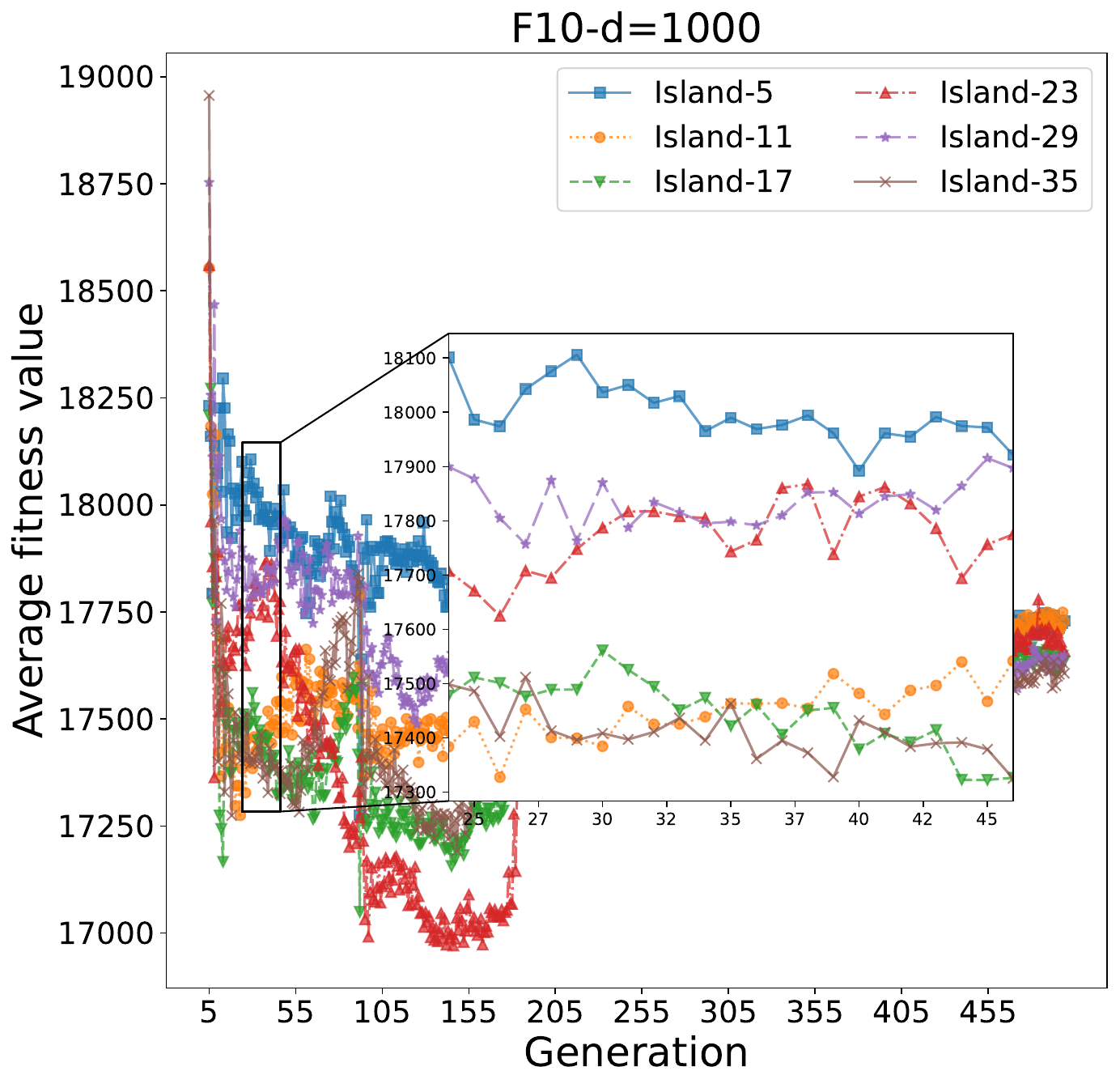}}
\subfigure{\includegraphics[height=0.15\textheight]{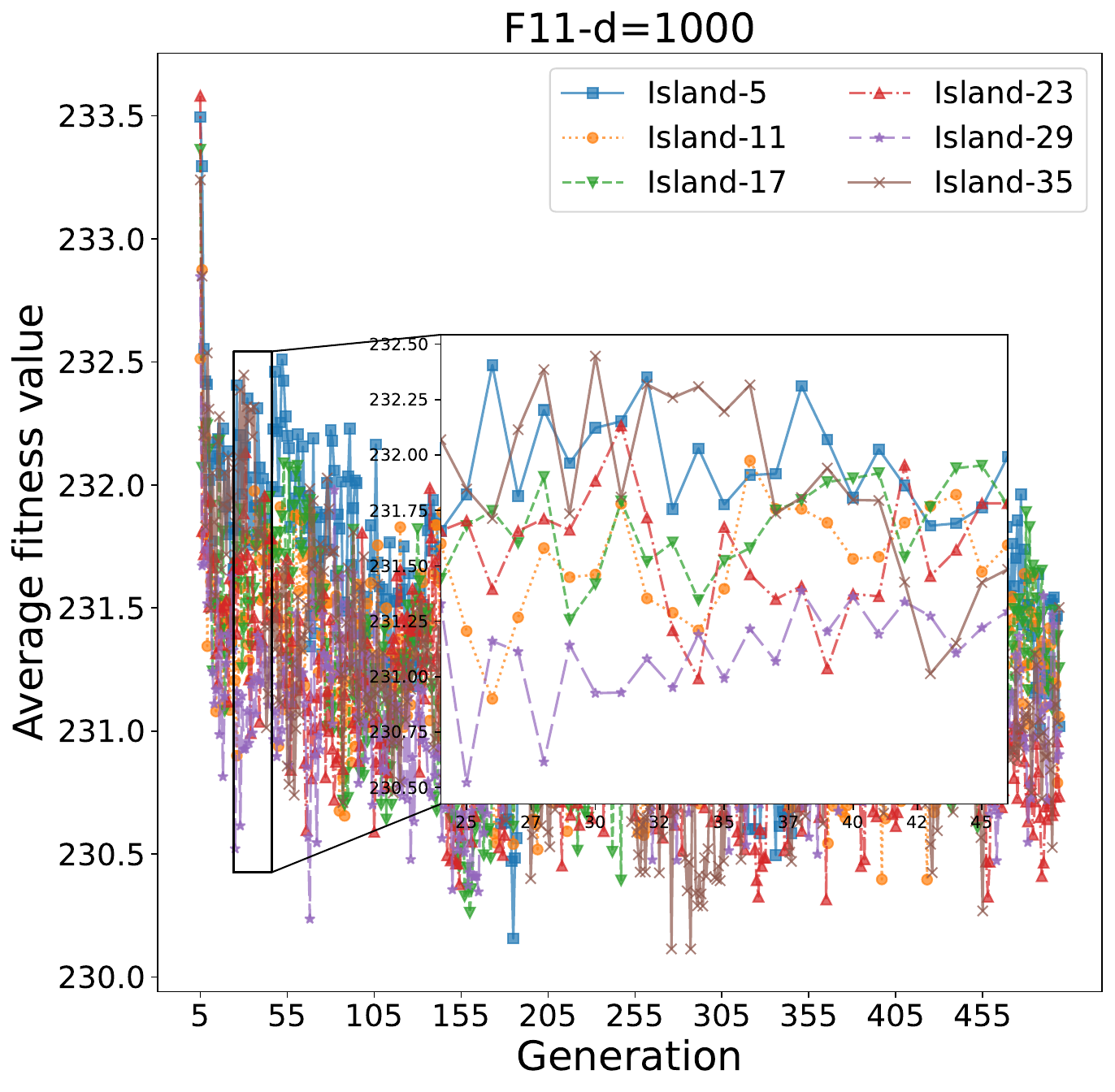}}
\subfigure{\includegraphics[height=0.15\textheight]{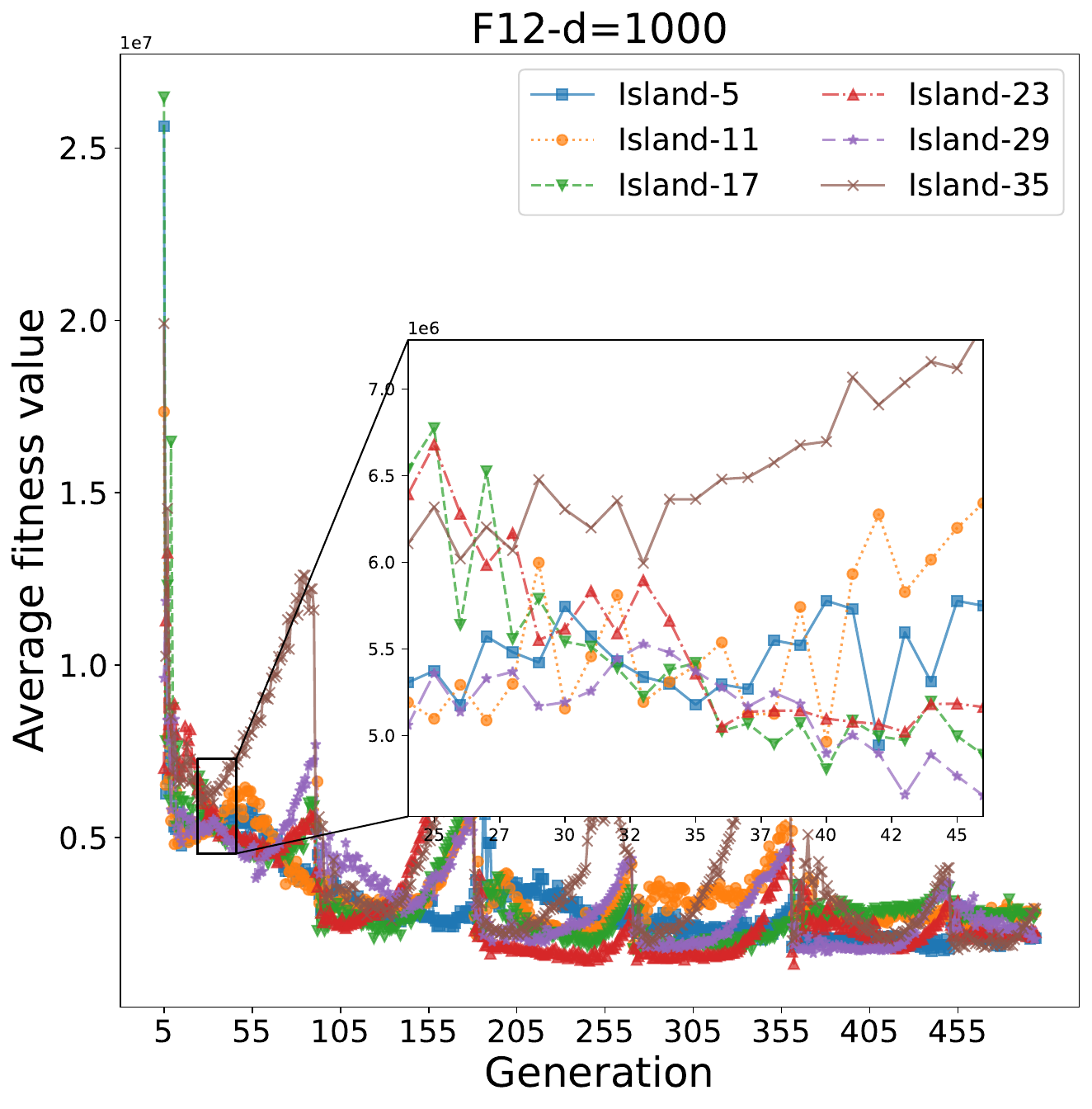}}
\subfigure{\includegraphics[height=0.15\textheight]{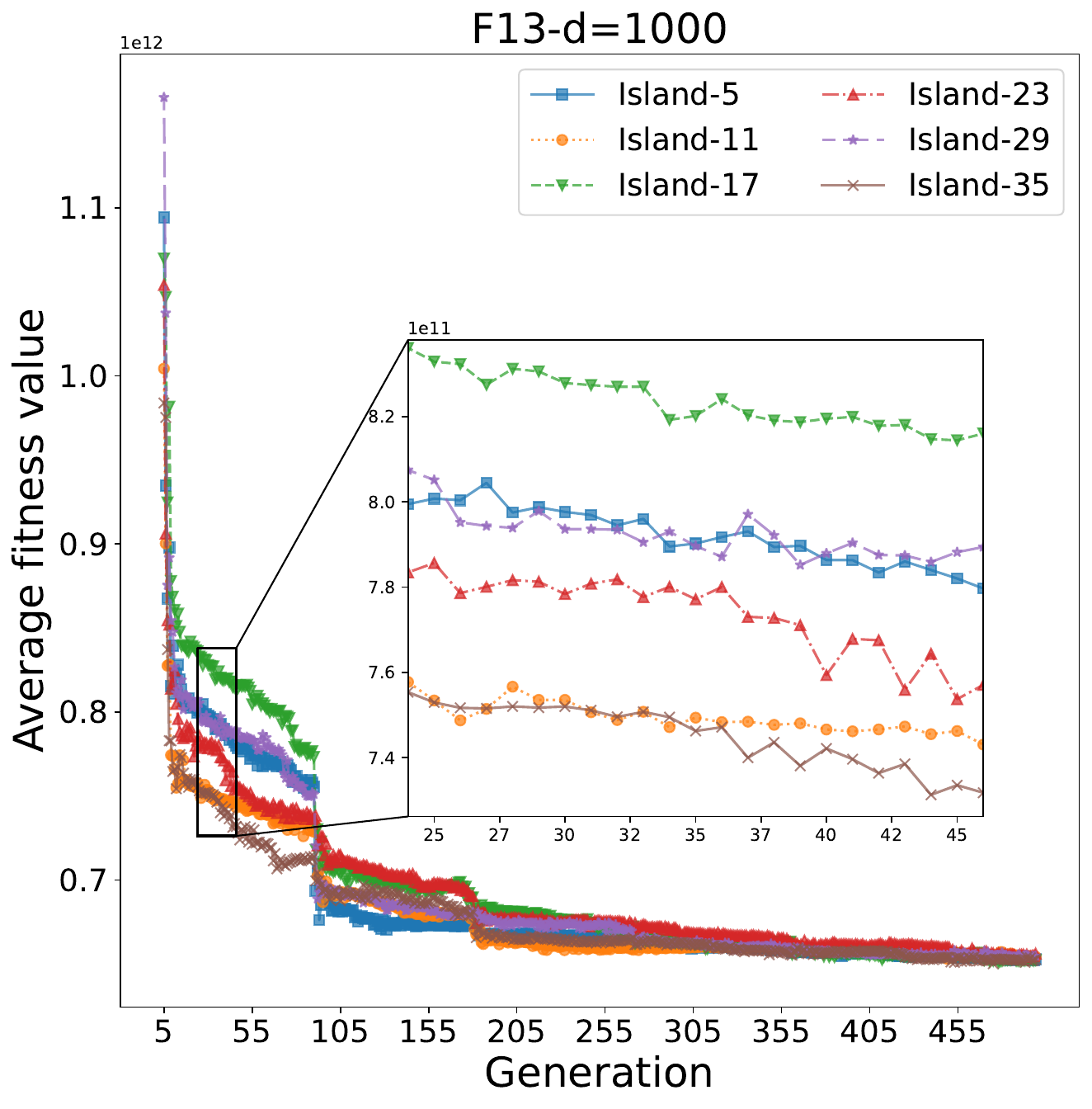}}
\subfigure{\includegraphics[height=0.15\textheight]{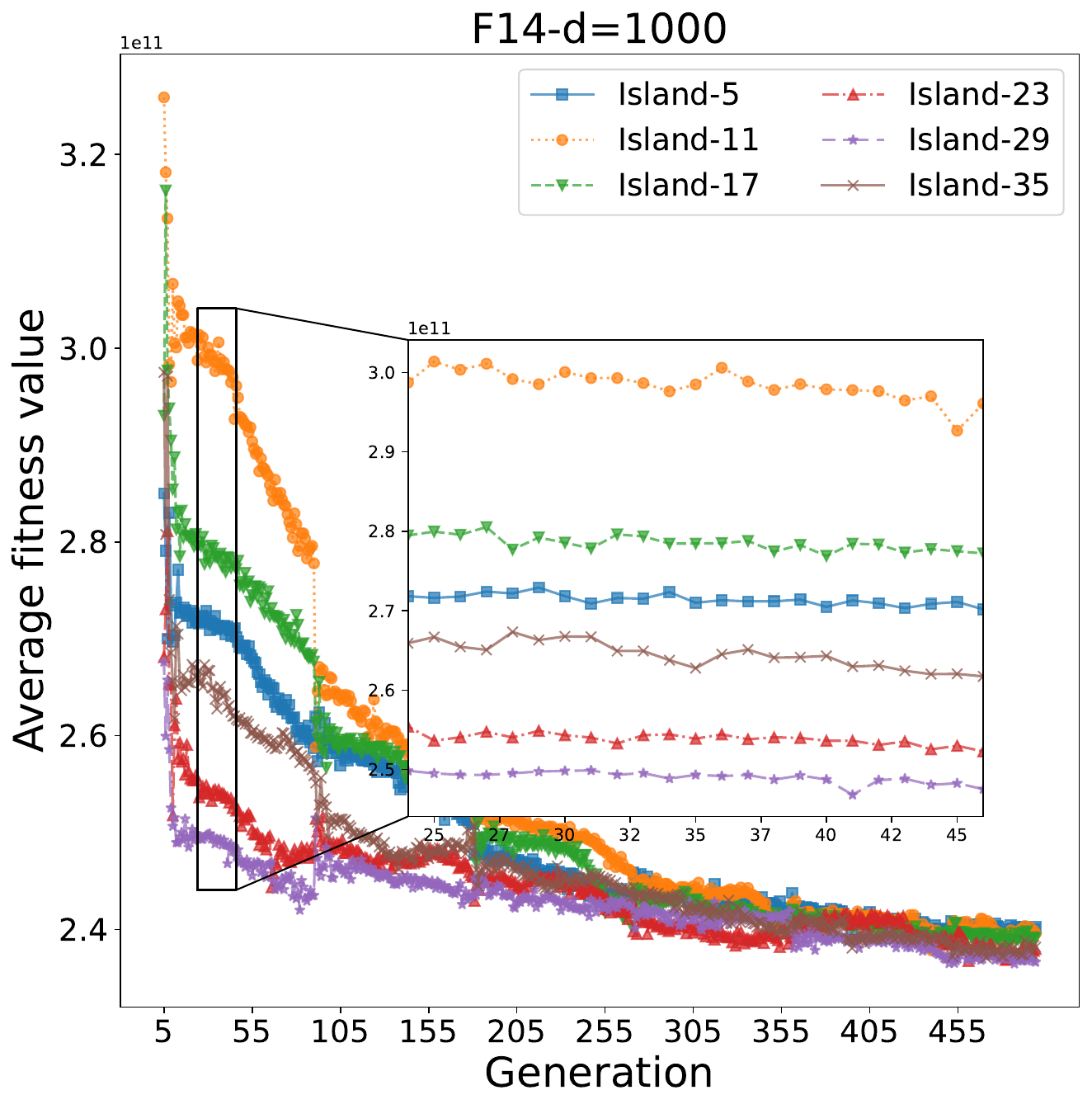}}
\subfigure{\includegraphics[height=0.15\textheight]{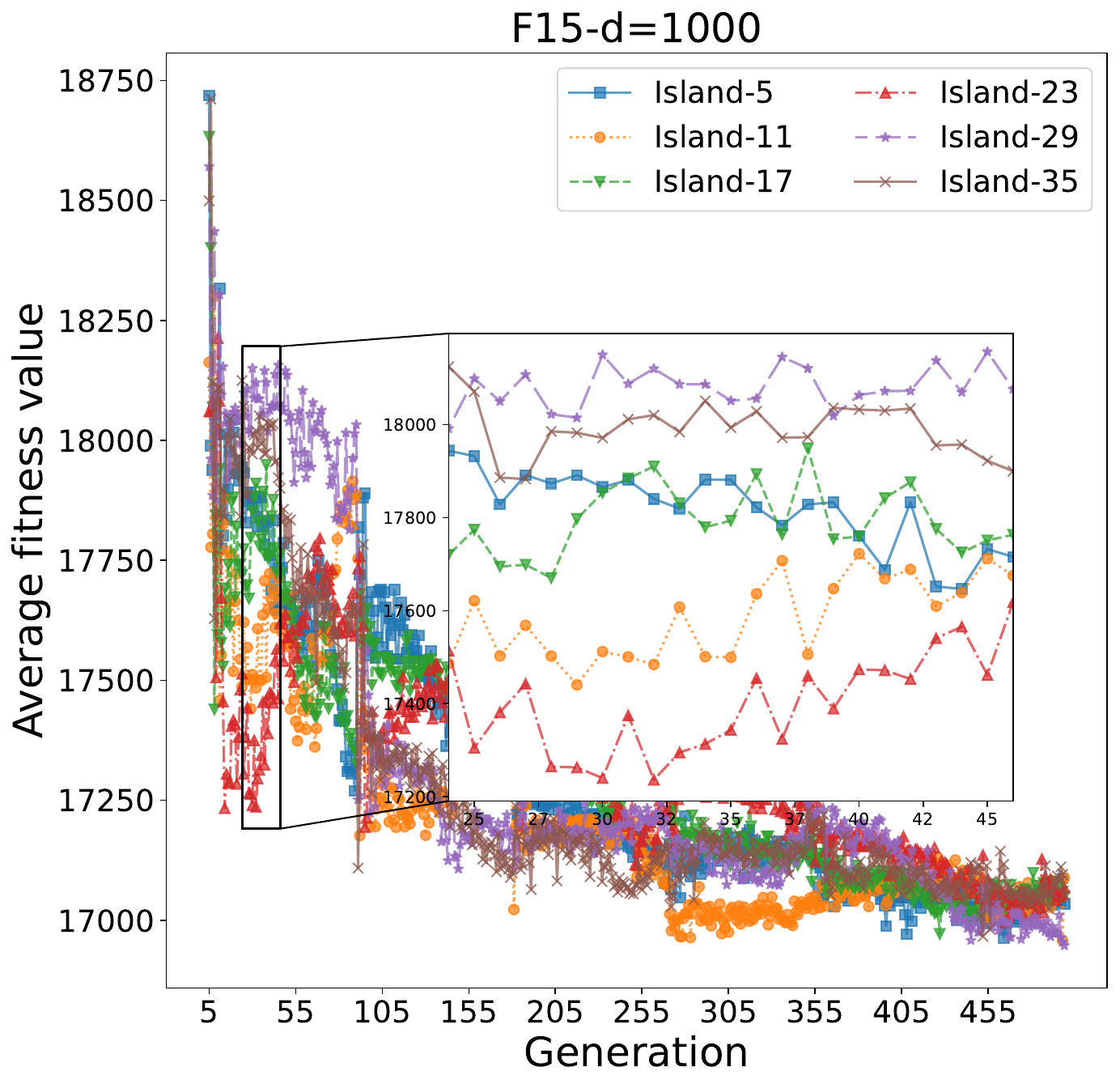}}
\subfigure{\includegraphics[height=0.15\textheight]{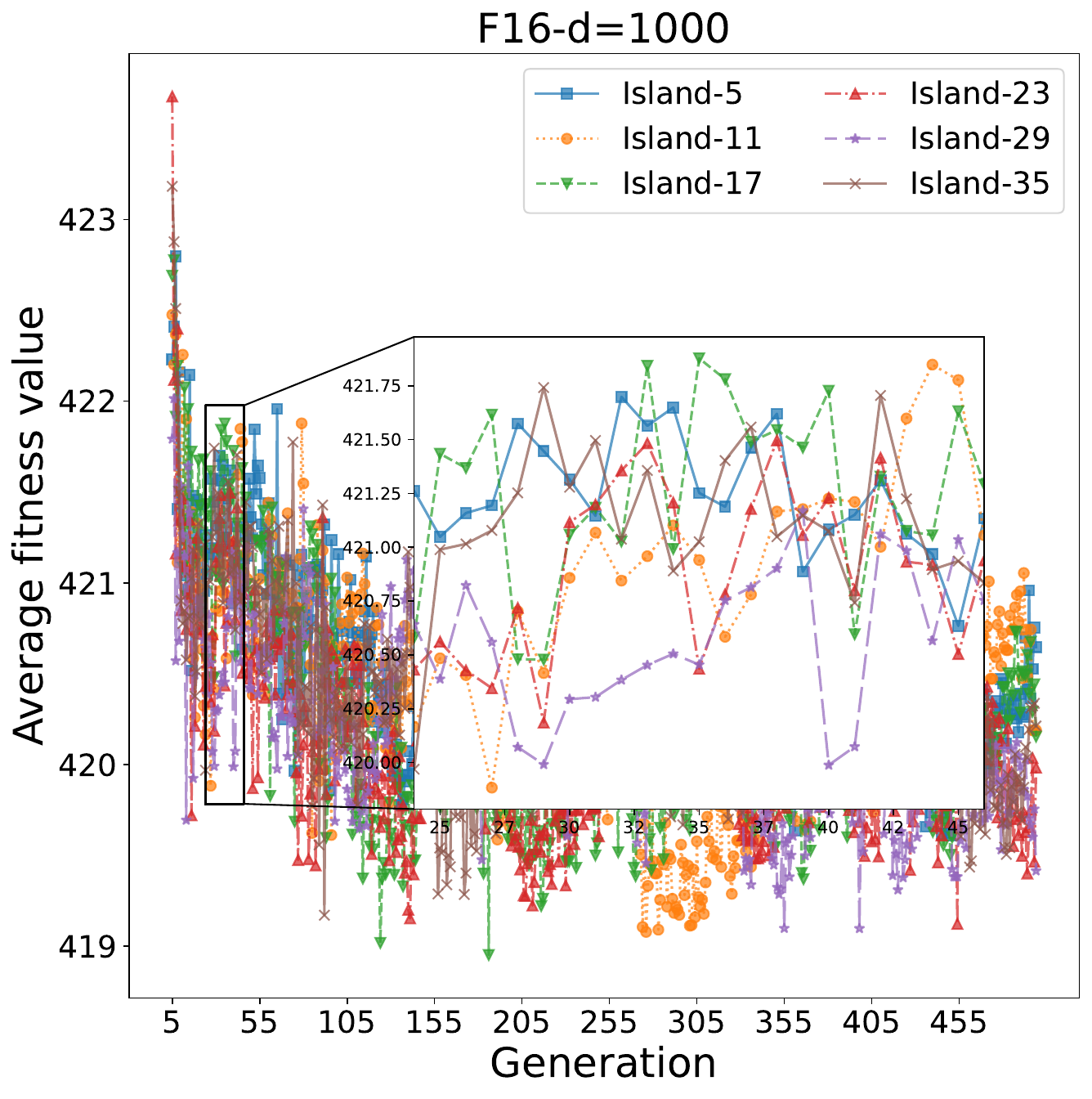}}
\subfigure{\includegraphics[height=0.15\textheight]{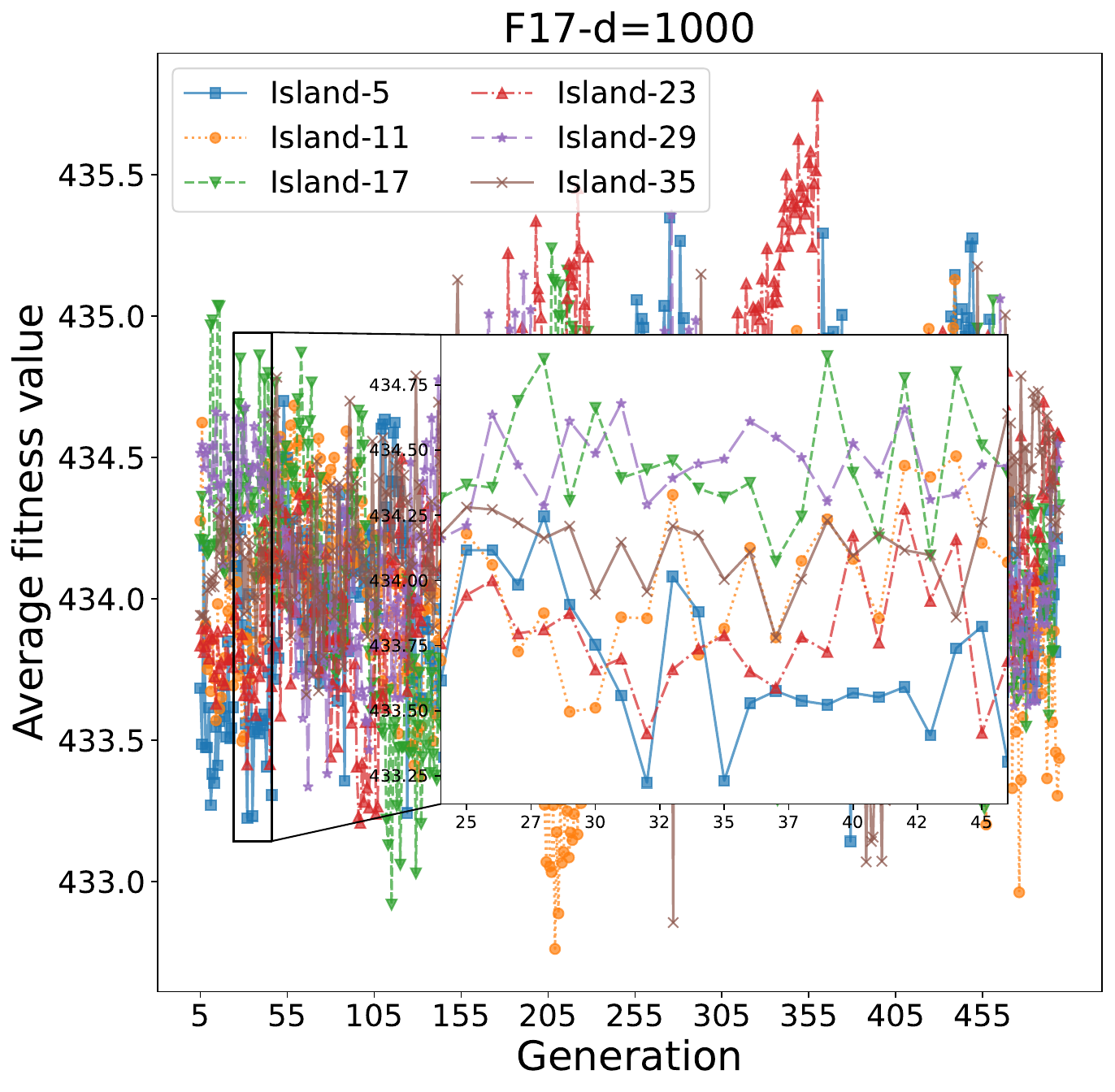}}
\subfigure{\includegraphics[height=0.15\textheight]{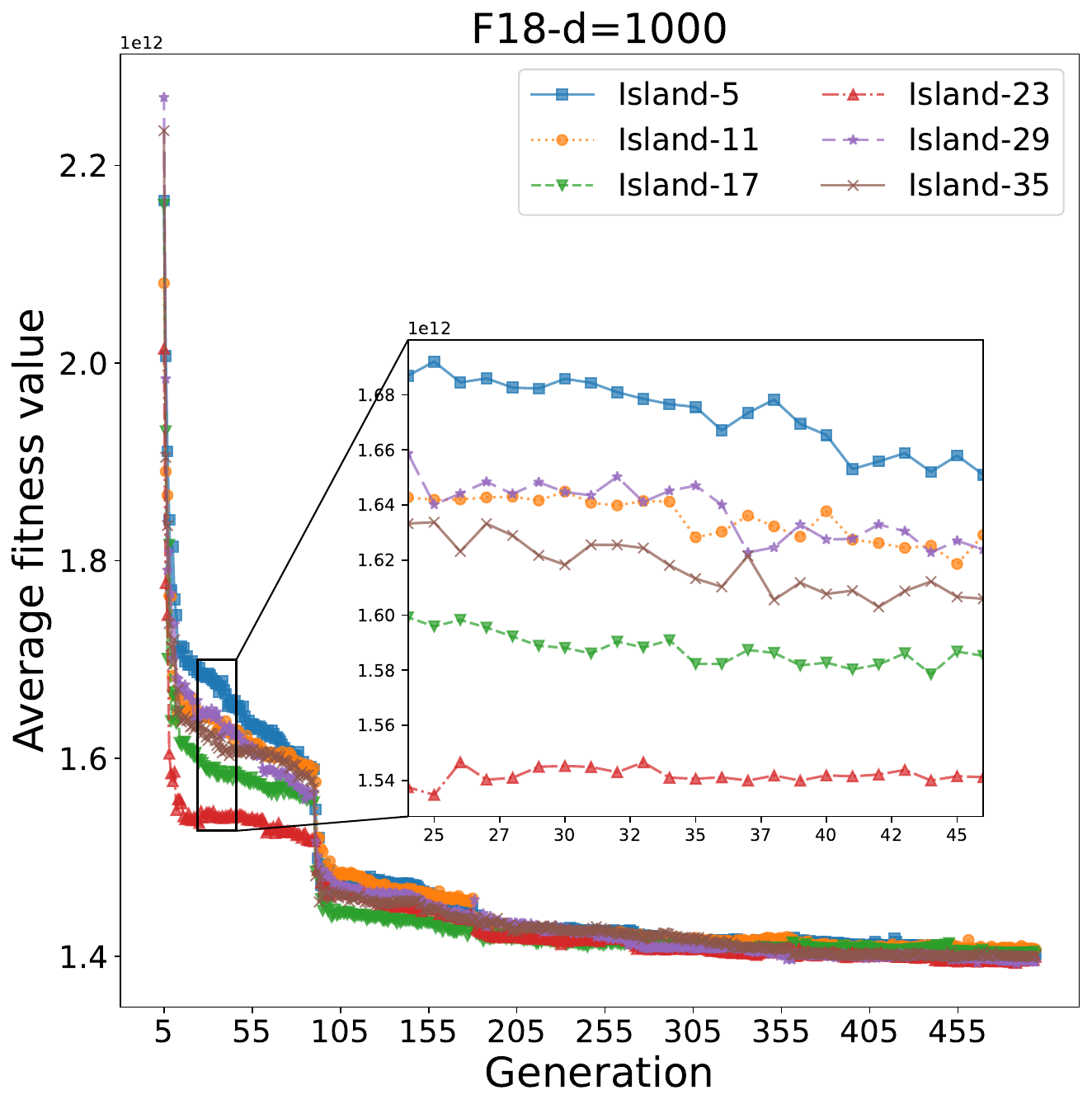}}
\subfigure{\includegraphics[height=0.15\textheight]{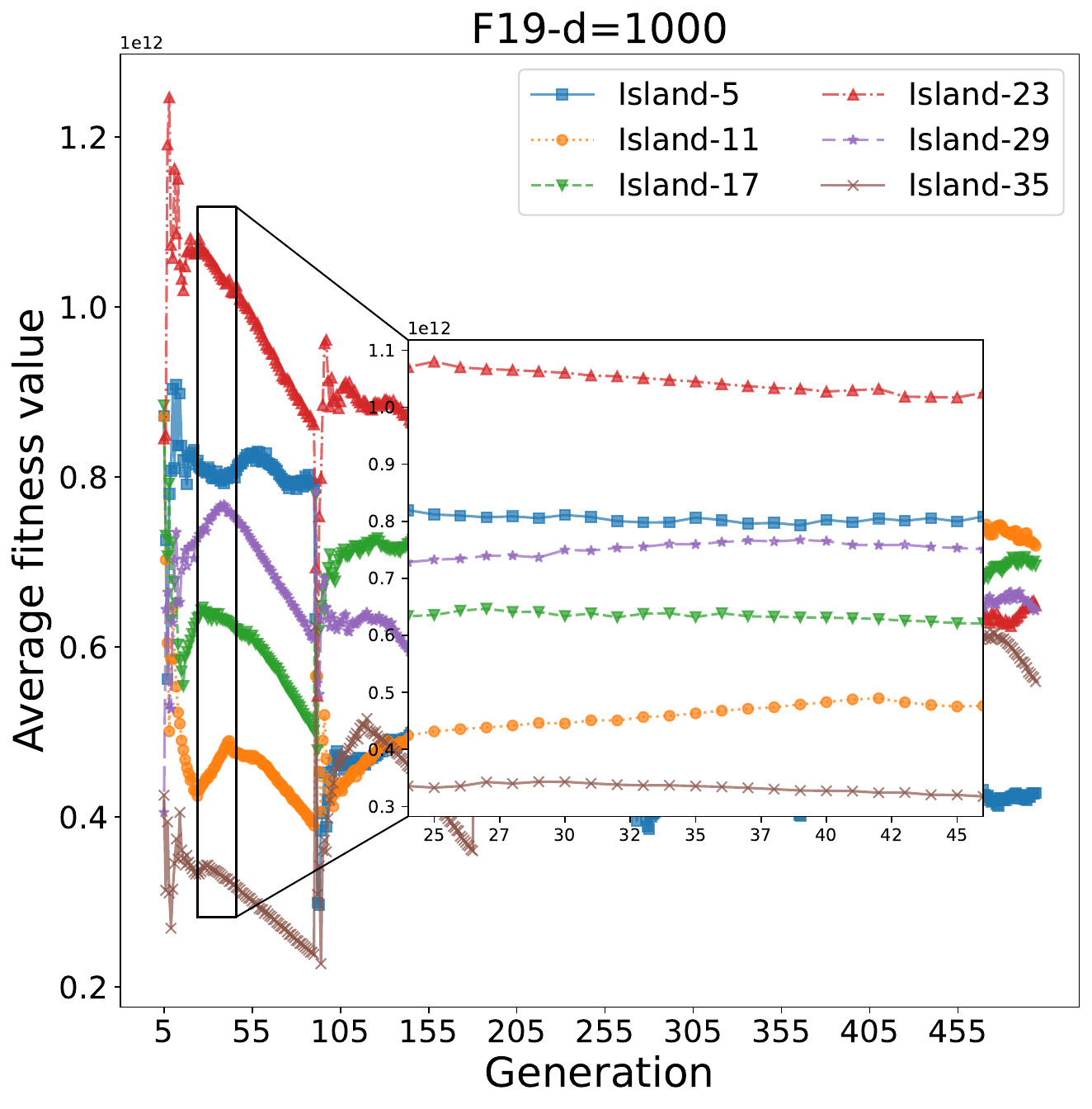}}
\subfigure{\includegraphics[height=0.15\textheight]{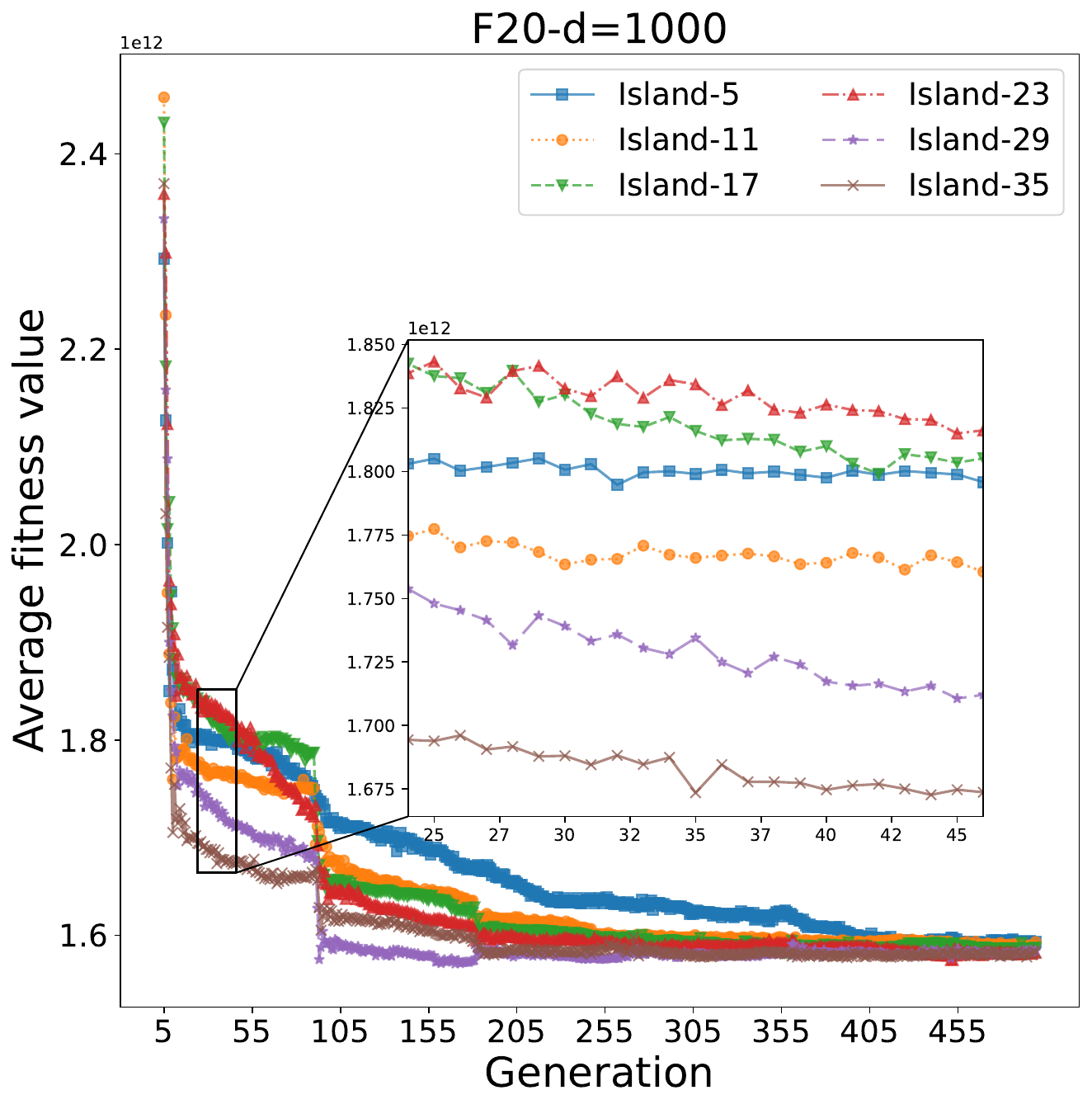}}
\caption{{The convergence curves of DSKT-DDEA on 20 problem instances (CEC2010).}}
\label{fig:covergence}
\end{figure}

Next, we discuss the advantages of DSKT-DDEA by analyzing the convergent curves.  \autoref{fig:covergence} shows the convergence curves of six islands on 20 problem instances. Migration is triggered every 90 generations, and each time it is triggered, the convergence curves exhibit a rapid drop. This phenomenon indicates that through the adaptive migration policy, after the migration and the local population recombination, they generate offsprings that are entirely different from the parent generation, taking a substantial step toward the optimal value. After the migration, each sub-population, under the guidance of the diverse surrogate model, explores the neighboring area within the island. It is a gradual and step-by-step search process. 
Consequently, the algorithm effectively combines local exploitation and global exploration: the local optimization within the island excels  at exploiting around a certain optimum, yielding more accurate result, and the migration process helps the evolutionary algorithm perform global optimization and escape from local optima. 
On the ``F12" and ``F17" cases, the convergence curves of some islands seem to show a rising fluctuation trend. Our early stop mechanism effectively prevents the deterioration of the optimal solution. When the elite solution is not improved, the algorithm stops iterating.

\subsubsection{Parameter Sensitivity Analysis} \label{subsec:psa}
In this section, we investigate the parameter sensitivity of the attraction decay factor $\rho$, migration gap, pseudo-label counts, and early stopping parameter $es$.

For simplicity of presentation, here we utilize a parameter sensitivity analysis method called performance profiles~\cite{dolan2002benchmarking}, which compares algorithms using the cumulative distribution function of performance indicators. Suppose there are $n_\mathrm{s}$ solvers and $n_\mathrm{p}$ problems, and define $
    result_{p, s} \text{as mean result of 20 runs by solver $s$ for problem $p$}. 
    $    
    Then compute the ``performance ratio'' as $$r_{p,s}=\frac{result_{p,s}}{\min\{result_{p,s}:s\in\mathcal{S}\}},$$ 
    which measures the relative performance of solver $s$ against the best performance for problem $p$.
    To obtain an overall assessment of the performance of the solver, define 
    $$
    \operatorname{CDF}_s(\tau)=\frac1{n_p}\text{size}\{p\in\mathcal{P}:r_{p,s}\leq\tau\},
    $$ 
    then 
    $\operatorname{CDF}_s(\tau)$ is essentially a measure that indicates the proportion of problems for which a particular solver $s\in S$ achieves a performance ratio that is within a factor $\tau\in\mathbb{R}$ of the best observed performance across all solvers. 
    By varying $\tau$, we can adjust our tolerance for what we consider to be ``comparably good'' performance. The function $\operatorname{CDF}_s(\tau)$ is the cumulative distribution function of the performance ratio. In particular,  
    we consider the solver whose probability $\operatorname{CDF}_s(\tau)$ reaches 1 as the first as the most preferred.
    
\begin{figure}
    \centering
    \begin{tabular}{cc}
        \includegraphics[width=0.25\linewidth]{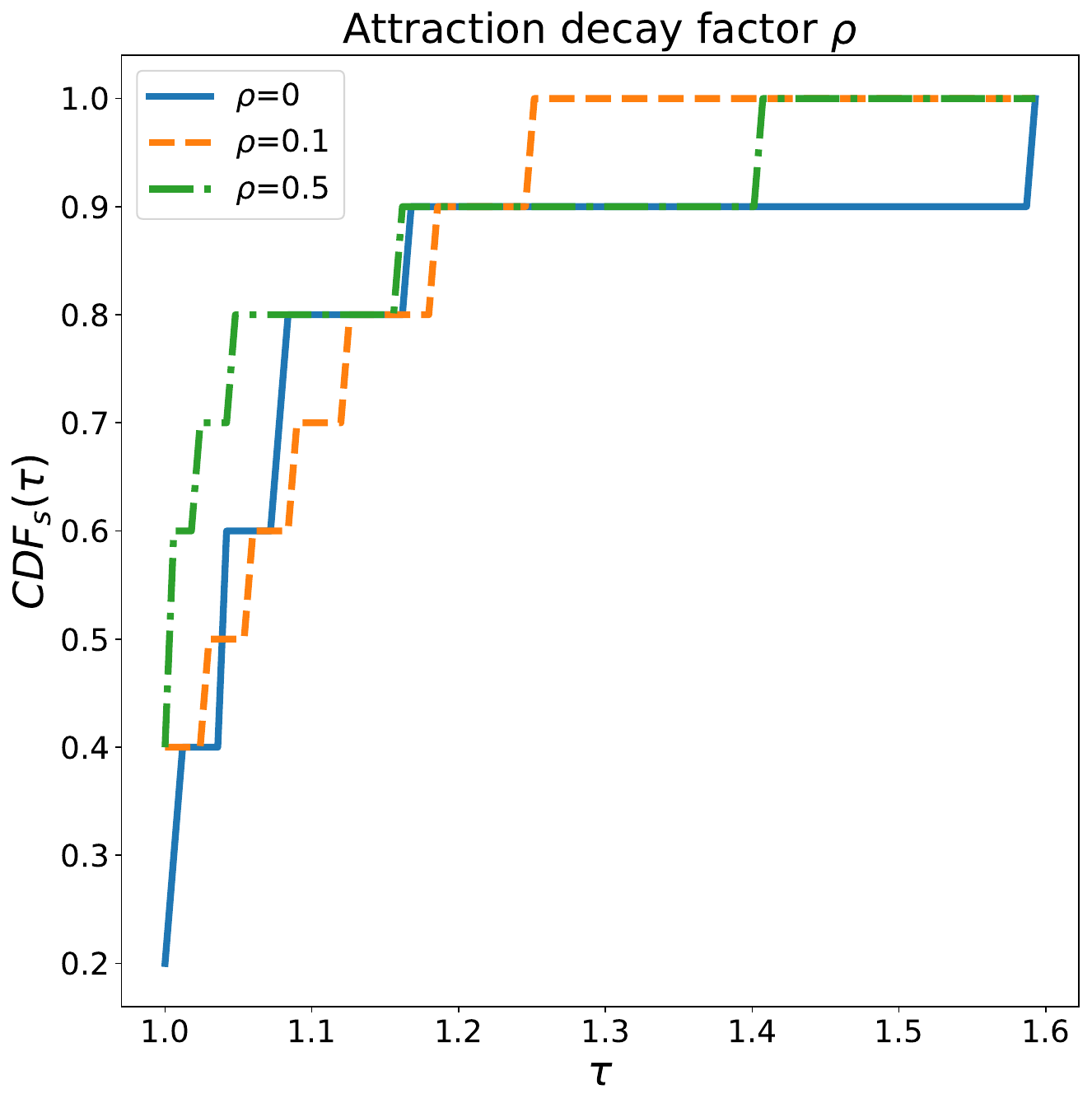}&
        \includegraphics[width=0.25\textwidth]{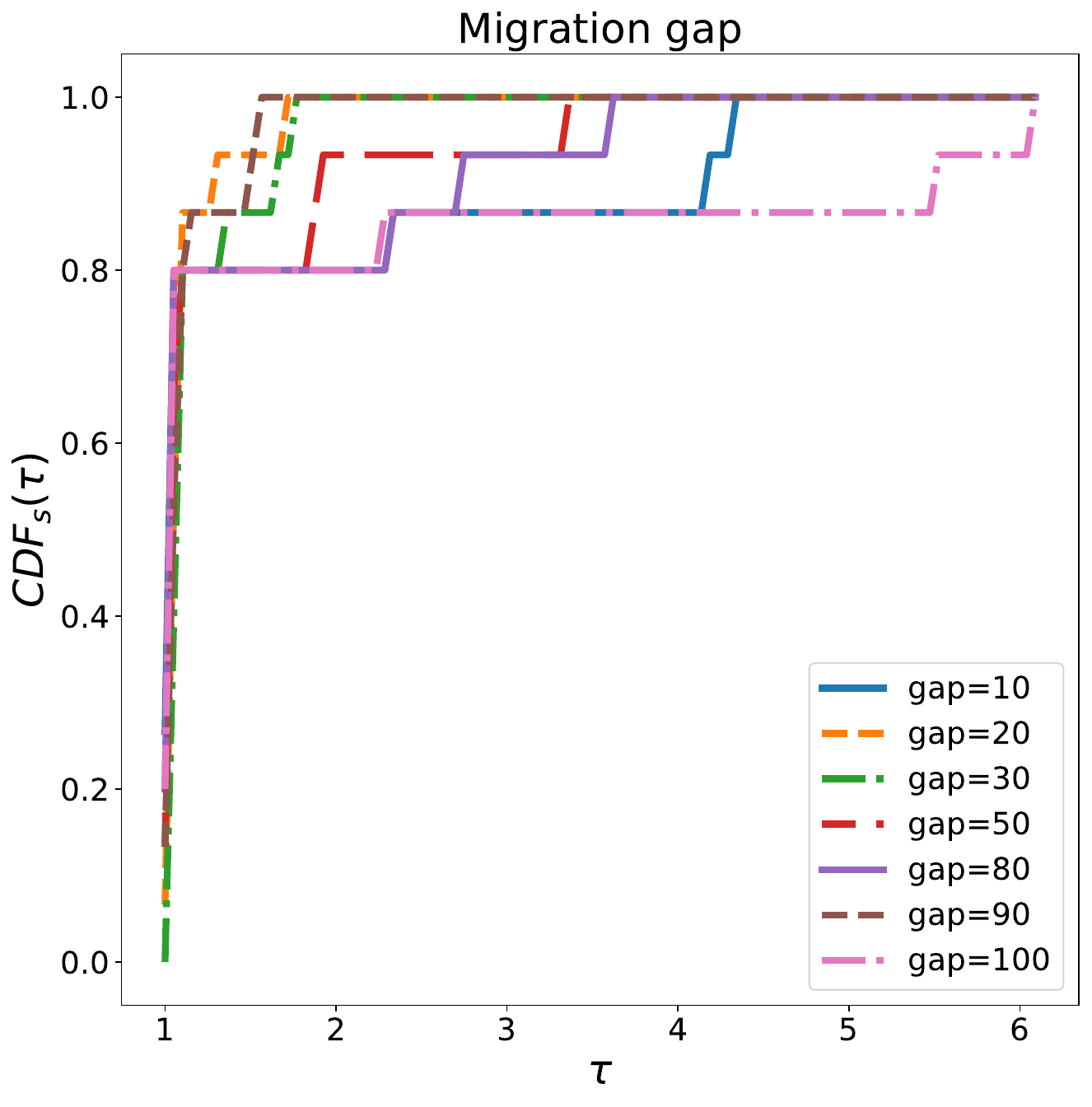} \\
       \footnotesize{(a)} & \footnotesize{(b)} \\ 
        \includegraphics[width=0.25\textwidth]{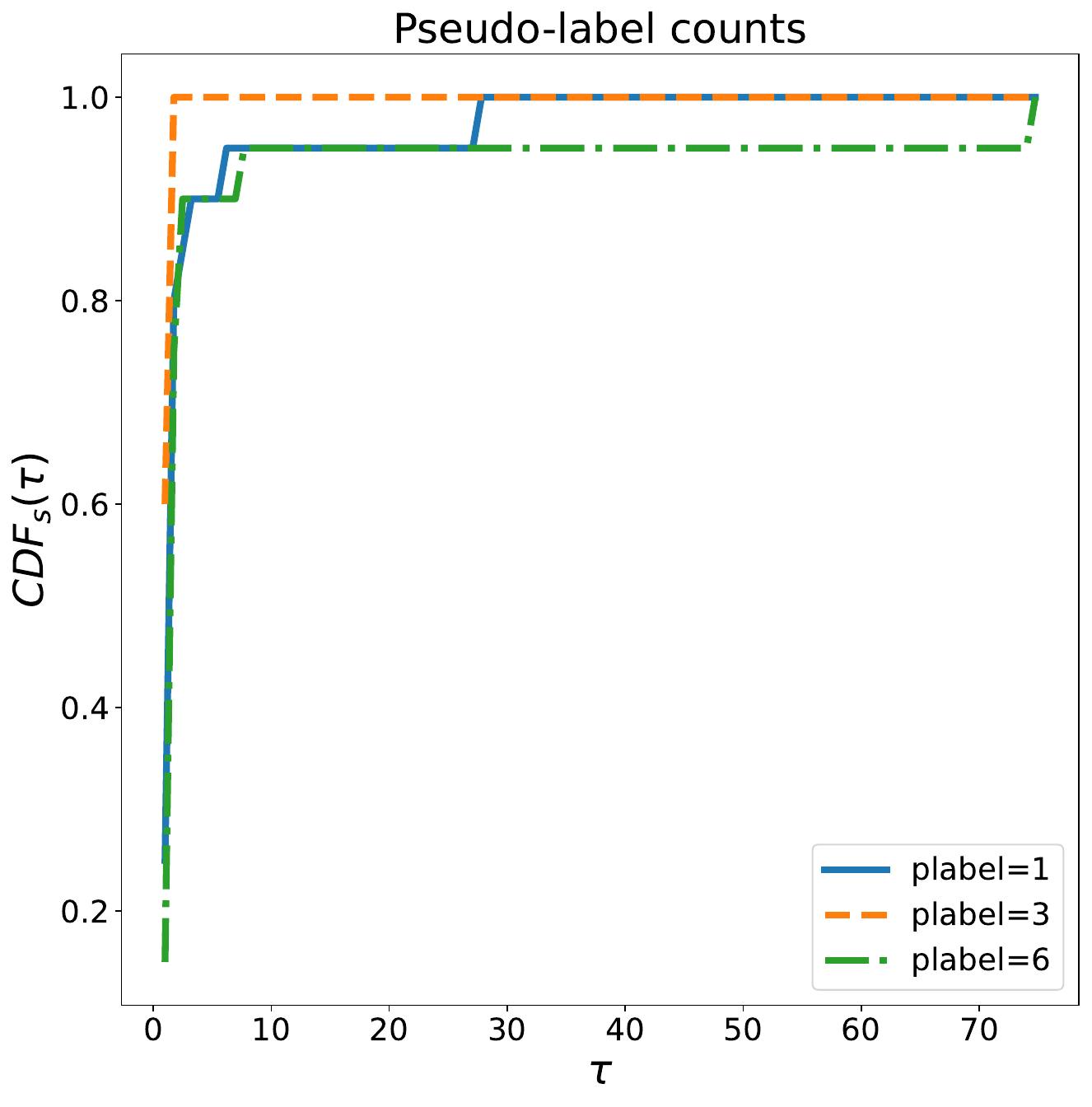} &
        \includegraphics[width=0.25\textwidth]{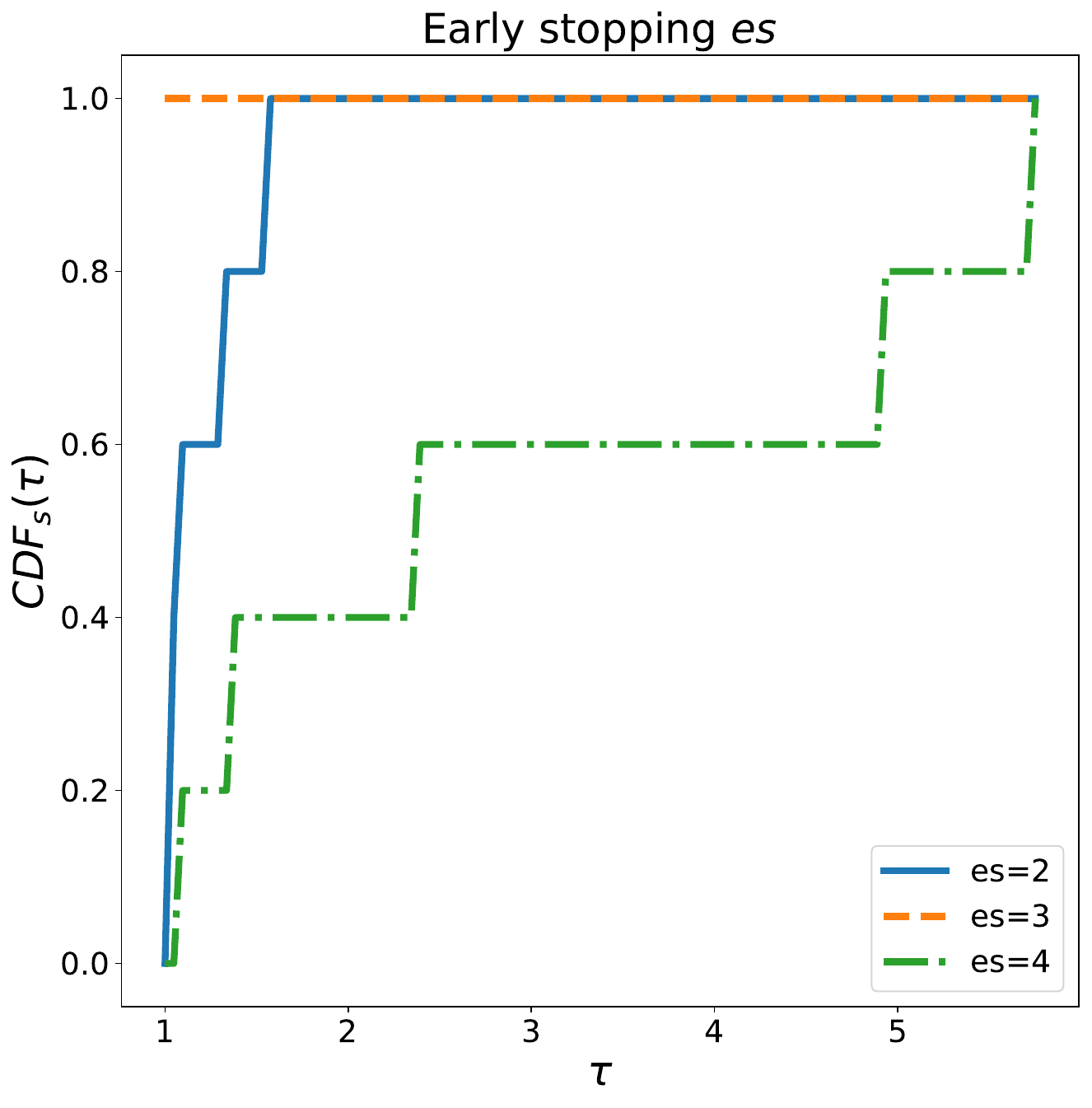} \\
      \footnotesize{(c)} & \footnotesize{(d)} \\ 
        
    \end{tabular}
    \caption{
    Parameter sensitivity analysis: 
    (a) We compare $\rho=0, 0.1, 0.5$, the solver with $\rho=0.1$ first reaches 1 for the performance profile measure. (b) The performance profile for different migration gaps shows that when the migration gap is 90, $CDF_s(\tau)$ reaches 1 the earliest. (c) The performance profile for different pseudo-label counts indicates that the performance is the best when the pseudo-label count is 3. (d) The performance profile for different early stopping criteria ($es$) demonstrates that the algorithm performs the best when $es=3$.}
    \label{fig:para}
\end{figure}

Parameter sensitivity analysis was conducted on
500-dimensional and 1000-dimensional problems across CEC2010 functions mentioned earlier. The results are summarized below.
\begin{itemize}
    \item \textbf{Attraction decay factor $\rho$: }As an investigation, we compare $\rho=0, 0.1, 0.5$. From the \autoref{fig:para}(a), it can be seen that the solver with $\rho=0.1$ first reaches 1 for the performance profile measure. The value of 0.1 ensures that the algorithm effectively retains historical island attraction data without excessively accumulating or ``forgetting'' it. This balance is crucial for maintaining the adaptive capability of the approach. 

    \item \textbf{Migration gap: }In assessing the migration gap, we considered intervals [10, 20, 30, 50,80, 90,100]. The migration gap refers to how often migration occurs in terms of generations. 
    According to \autoref{fig:para}(b), We recommend 90.

    \item \textbf{Pseudo-label counts: }To evaluate the impact of varying augmented data volumes (pseudo-label counts), we compared the numbers [1, 3, 6 ]. According to \autoref{fig:para}(c), the best performance is achieved when the number of pseudo-labels is~3. 

   \item \textbf{Early stopping $es$: }For the early stopping criterion, denoted as $es$, we experimented with values $[2,3,4]$. Here, $e s=3$ implies that the algorithm halts if there is no improvement in the global optimum over three successive migration intervals. \autoref{fig:para}(d) illustrates that an $es$ value of 3 yields the optimal performance.

\end{itemize}

\subsection{Parallel Performance Analysis }
\label{subsec:para}
The proposed DSKT-DDEA employs an island-based evolutionary framework, inherently designed for parallelization. For instance, we can execute parallel local optimizations on each island across multiple processors and subsequently initiate migration once the tasks are completed. In this section, we delve into the parallel performance analysis of DSKT-DDEA.

Parallel evaluation can be measured using the parallel speedup ratio ($R_s$), defined as:
$$R_s = \frac{t_s}{t_p}$$
where $t_s$ is the time required to complete the DSKT-DDEA using serial programming, and $t_p$ is the time required for parallel execution.

\begin{figure}
\subfigure{\includegraphics[width=0.3\textwidth]{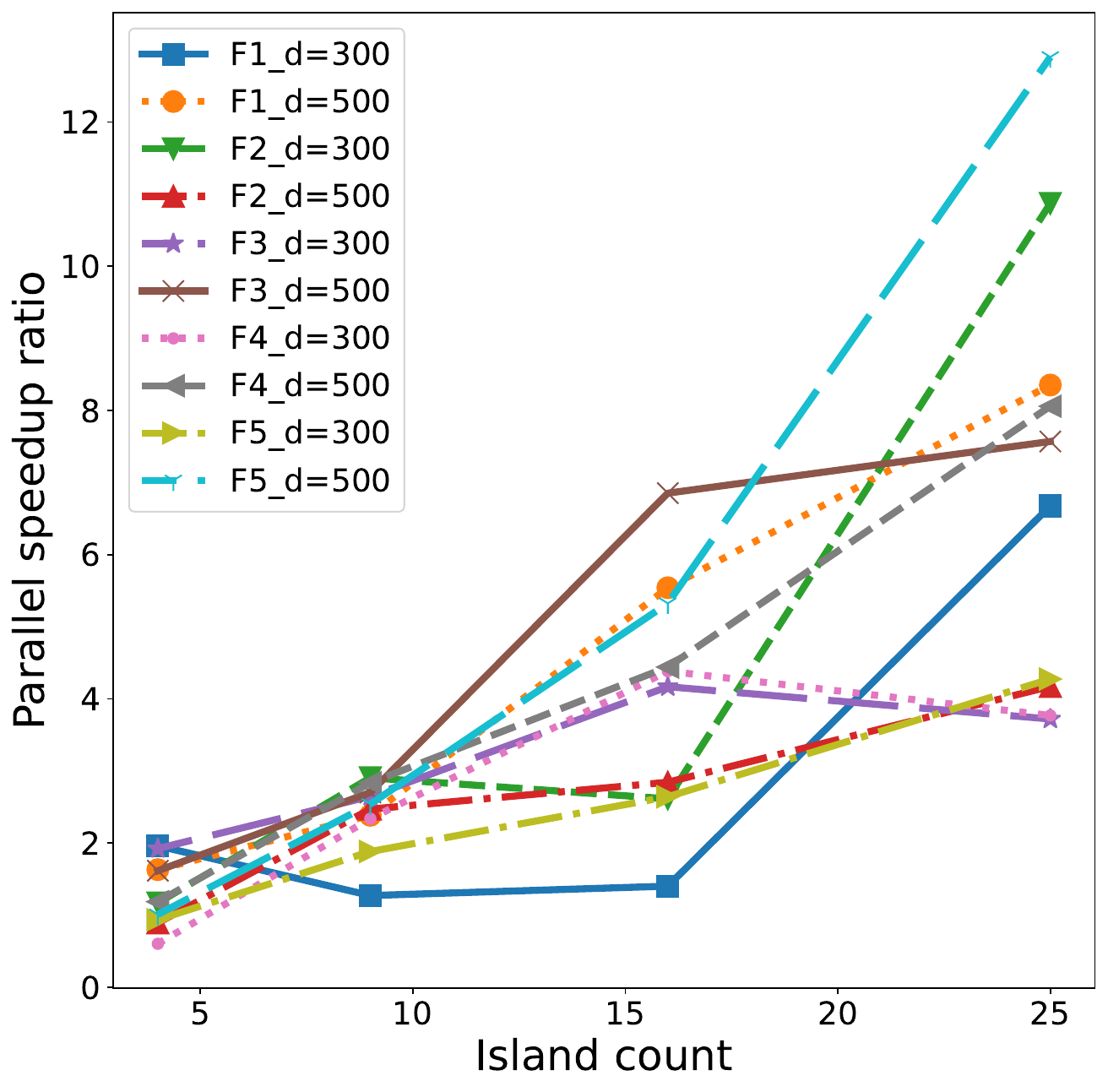}}
\caption{Parallel speedup ratio on CEC2010.}
\label{fig:parallel}
\end{figure}

As shown in \autoref{fig:parallel}, the parallel speedup is presented. We conducted an analysis considering scenarios where the number of islands is [4, 9, 16, 25]. Note that, since the islands in DSKT-DDEA adhere to a von Neumann topology, we choose the number of islands as an integer capable of being square rooted, which ensures the a balanced distribution of islands. 
In general, the algorithm demonstrates a sub-linear speedup as the number of deployed islands increases. But in some cases, the speedup fluctuates, 
which may be caused by the synchronous communication process. 
Specifically, when an island updates its own model, it needs to lock the model, consequently inducing waiting times for other islands that necessitate the same model. The speedup is also influenced by the inefficiency of the Python library multiprocess.Manager regarding shared data management. Various enhancement methods, such as implementing shared memory or adopting asynchronous communication, are available. Nevertheless, we consider these methods to be beyond the scope of this paper.

\section{Conclusion}\label{sec:conclusion_future_work}
Offline DDEA faces challenges when solving expensive large-scale problems. In this regard, we propose the DSKT-DDEA, an island-based algorithm characterized by diverse surrogates and adaptive knowledge transfer. 
Specifically, DSKT-DDEA builds multiple fitness-approximation models based on locally sampled data, guiding diverse subpopulations for optimization. During the optimization process, a semi-supervised learning method is utilized to effectively fine-tune the surrogate models, leveraging different distribution information generated during the population search. The surrogate models are progressively refined through population feedback to better fit the fitness landscape, thereby enhancing their ability to address complex problems such as multimodal and large-scale ones. Subsequently, after a certain number of generations of independent optimization for each subpopulation, the inter-island knowledge transfer is performed, with individual migration probabilities adaptively adjusted based on historical transfer effectiveness. 
The outcome is an algorithm that effectively overcomes the common optimization challenges.

We conducted comprehensive experiments to assess the performance of DSKT-DDEA, comparing it with state-of-the-art DDEAs in both offline and online settings. The results show that DSKT-DDEA excels in solving high-dimensional problems. The ablation study also validate the effectiveness of our design choices. Additionally, DSKT-DDEA exhibits good parallelism and scalability.
For future work, 
we may explore the utilization of unsupervised pre-training to further enhance performance.  
Additionally, given the strong performance of DSKT-DDEA, extending the algorithm to different scenarios, such as solving multi-task and multi-objective problems, holds great promise.



\begin{acks}
This work was supported in part by the Guangdong Natural Science Funds for Distinguished Young Scholars No. 2022B1515020049, in part by the National Natural Science Foundation of China No. 62276100, in part by the Guangdong Regional Joint Funds for Basic and Applied Research No. 2021B1515120078, in part by the National Research Foundation of Korea No. NRF2022H1D3A 2A01093478, and in part by the TCL Young Scholars Program. 
\end{acks}

\bibliographystyle{ACM-Reference-Format}
\bibliography{sample-base}


\begin{thebibliography}{97}


\ifx \showCODEN    \undefined \def \showCODEN     #1{\unskip}     \fi
\ifx \showISBNx    \undefined \def \showISBNx     #1{\unskip}     \fi
\ifx \showISBNxiii \undefined \def \showISBNxiii  #1{\unskip}     \fi
\ifx \showISSN     \undefined \def \showISSN      #1{\unskip}     \fi
\ifx \showLCCN     \undefined \def \showLCCN      #1{\unskip}     \fi
\ifx \shownote     \undefined \def \shownote      #1{#1}          \fi
\ifx \showarticletitle \undefined \def \showarticletitle #1{#1}   \fi
\ifx \showURL      \undefined \def \showURL       {\relax}        \fi
\providecommand\bibfield[2]{#2}
\providecommand\bibinfo[2]{#2}
\providecommand\natexlab[1]{#1}
\providecommand\showeprint[2][]{arXiv:#2}

\bibitem[Abdelhafez et~al\mbox{.}(2019)]%
        {best}
\bibfield{author}{\bibinfo{person}{Amr Abdelhafez}, \bibinfo{person}{Enrique Alba}, {and} \bibinfo{person}{Gabriel Luque}.} \bibinfo{year}{2019}\natexlab{}.
\newblock \showarticletitle{Performance analysis of synchronous and asynchronous distributed genetic algorithms on multiprocessors}.
\newblock \bibinfo{journal}{\emph{Swarm and Evolutionary Computation}}  \bibinfo{volume}{49} (\bibinfo{year}{2019}), \bibinfo{pages}{147--157}.
\newblock


\bibitem[Al-Betar et~al\mbox{.}(2019)]%
        {flower}
\bibfield{author}{\bibinfo{person}{Mohammed~Azmi Al-Betar}, \bibinfo{person}{Mohammed~A Awadallah}, \bibinfo{person}{Iyad Abu~Doush}, \bibinfo{person}{Abdelaziz~I Hammouri}, \bibinfo{person}{Majdi Mafarja}, {and} \bibinfo{person}{Zaid Abdi~Alkareem Alyasseri}.} \bibinfo{year}{2019}\natexlab{}.
\newblock \showarticletitle{Island flower pollination algorithm for global optimization}.
\newblock \bibinfo{journal}{\emph{The Journal of Supercomputing}}  \bibinfo{volume}{75} (\bibinfo{year}{2019}), \bibinfo{pages}{5280--5323}.
\newblock


\bibitem[Awadallah et~al\mbox{.}(2020)]%
        {bee}
\bibfield{author}{\bibinfo{person}{Mohammed~A Awadallah}, \bibinfo{person}{Mohammed~Azmi Al-Betar}, \bibinfo{person}{Asaju~La’aro Bolaji}, \bibinfo{person}{Iyad~Abu Doush}, \bibinfo{person}{Abdelaziz~I Hammouri}, {and} \bibinfo{person}{Majdi Mafarja}.} \bibinfo{year}{2020}\natexlab{}.
\newblock \showarticletitle{Island artificial bee colony for global optimization}.
\newblock \bibinfo{journal}{\emph{Soft Computing}}  \bibinfo{volume}{24} (\bibinfo{year}{2020}), \bibinfo{pages}{13461--13487}.
\newblock


\bibitem[Chen et~al\mbox{.}(2010)]%
        {CCVIL}
\bibfield{author}{\bibinfo{person}{Wenxiang Chen}, \bibinfo{person}{Thomas Weise}, \bibinfo{person}{Zhenyu Yang}, {and} \bibinfo{person}{Ke Tang}.} \bibinfo{year}{2010}\natexlab{}.
\newblock \showarticletitle{Large-scale global optimization using cooperative coevolution with variable interaction learning}. In \bibinfo{booktitle}{\emph{Parallel Problem Solving from Nature, PPSN XI: 11th International Conference, Krak{\'o}w, Poland, September 11-15, 2010, Proceedings, Part II 11}}. Springer, \bibinfo{pages}{300--309}.
\newblock


\bibitem[Chugh et~al\mbox{.}(2017)]%
        {blast__furnace2}
\bibfield{author}{\bibinfo{person}{Tinkle Chugh}, \bibinfo{person}{Nirupam Chakraborti}, \bibinfo{person}{Karthik Sindhya}, {and} \bibinfo{person}{Yaochu Jin}.} \bibinfo{year}{2017}\natexlab{}.
\newblock \showarticletitle{A data-driven surrogate-assisted evolutionary algorithm applied to a many-objective blast furnace optimization problem}.
\newblock \bibinfo{journal}{\emph{Materials and Manufacturing Processes}} \bibinfo{volume}{32}, \bibinfo{number}{10} (\bibinfo{year}{2017}), \bibinfo{pages}{1172--1178}.
\newblock


\bibitem[Chugh et~al\mbox{.}(2020)]%
        {SA-COSO}
\bibfield{author}{\bibinfo{person}{Tinkle Chugh}, \bibinfo{person}{Chaoli Sun}, \bibinfo{person}{Handing Wang}, {and} \bibinfo{person}{Yaochu Jin}.} \bibinfo{year}{2020}\natexlab{}.
\newblock \showarticletitle{Surrogate-assisted evolutionary optimization of large problems}.
\newblock \bibinfo{journal}{\emph{High-Performance Simulation-Based Optimization}} (\bibinfo{year}{2020}), \bibinfo{pages}{165--187}.
\newblock


\bibitem[de~Jes{\'u}s Leal-Romo et~al\mbox{.}(2020)]%
        {polynomial_regression}
\bibfield{author}{\bibinfo{person}{Felipe de Jes{\'u}s Leal-Romo}, \bibinfo{person}{Jos{\'e}~Ernesto Rayas-S{\'a}nchez}, {and} \bibinfo{person}{Jos{\'e}~Luis Ch{\'a}vez-Hurtado}.} \bibinfo{year}{2020}\natexlab{}.
\newblock \showarticletitle{Surrogate-based analysis and design optimization of power delivery networks}.
\newblock \bibinfo{journal}{\emph{IEEE Transactions on Electromagnetic Compatibility}} \bibinfo{volume}{62}, \bibinfo{number}{6} (\bibinfo{year}{2020}), \bibinfo{pages}{2528--2537}.
\newblock


\bibitem[Deb et~al\mbox{.}(1995)]%
        {PM1}
\bibfield{author}{\bibinfo{person}{Kalyanmoy Deb}, \bibinfo{person}{Ram~Bhushan Agrawal}, {et~al\mbox{.}}} \bibinfo{year}{1995}\natexlab{}.
\newblock \showarticletitle{Simulated binary crossover for continuous search space}.
\newblock \bibinfo{journal}{\emph{Complex systems}} \bibinfo{volume}{9}, \bibinfo{number}{2} (\bibinfo{year}{1995}), \bibinfo{pages}{115--148}.
\newblock


\bibitem[Derrac et~al\mbox{.}(2011)]%
        {wilcoxon}
\bibfield{author}{\bibinfo{person}{Joaqu{\'\i}n Derrac}, \bibinfo{person}{Salvador Garc{\'\i}a}, \bibinfo{person}{Daniel Molina}, {and} \bibinfo{person}{Francisco Herrera}.} \bibinfo{year}{2011}\natexlab{}.
\newblock \showarticletitle{A practical tutorial on the use of nonparametric statistical tests as a methodology for comparing evolutionary and swarm intelligence algorithms}.
\newblock \bibinfo{journal}{\emph{Swarm and Evolutionary Computation}} \bibinfo{volume}{1}, \bibinfo{number}{1} (\bibinfo{year}{2011}), \bibinfo{pages}{3--18}.
\newblock


\bibitem[Dobnikar et~al\mbox{.}(1999)]%
        {PM2}
\bibfield{author}{\bibinfo{person}{Andrej Dobnikar}, \bibinfo{person}{Nigel~C Steele}, \bibinfo{person}{David~W Pearson}, \bibinfo{person}{Rudolf~F Albrecht}, \bibinfo{person}{Kalyanmoy Deb}, {and} \bibinfo{person}{Samir Agrawal}.} \bibinfo{year}{1999}\natexlab{}.
\newblock \showarticletitle{A niched-penalty approach for constraint handling in genetic algorithms}. In \bibinfo{booktitle}{\emph{Artificial Neural Nets and Genetic Algorithms: Proceedings of the International Conference in Portoro{\v{z}}, Slovenia, 1999}}. Springer, \bibinfo{pages}{235--243}.
\newblock


\bibitem[Dolan and Mor{\'e}(2002)]%
        {dolan2002benchmarking}
\bibfield{author}{\bibinfo{person}{Elizabeth~D Dolan} {and} \bibinfo{person}{Jorge~J Mor{\'e}}.} \bibinfo{year}{2002}\natexlab{}.
\newblock \showarticletitle{Benchmarking optimization software with performance profiles}.
\newblock \bibinfo{journal}{\emph{Mathematical programming}}  \bibinfo{volume}{91} (\bibinfo{year}{2002}), \bibinfo{pages}{201--213}.
\newblock


\bibitem[Dong and Dong(2020)]%
        {RBFN2}
\bibfield{author}{\bibinfo{person}{Huachao Dong} {and} \bibinfo{person}{Zuomin Dong}.} \bibinfo{year}{2020}\natexlab{}.
\newblock \showarticletitle{Surrogate-assisted grey wolf optimization for high-dimensional, computationally expensive black-box problems}.
\newblock \bibinfo{journal}{\emph{Swarm and Evolutionary Computation}}  \bibinfo{volume}{57} (\bibinfo{year}{2020}), \bibinfo{pages}{100713}.
\newblock


\bibitem[Duarte et~al\mbox{.}(2017)]%
        {dynamic}
\bibfield{author}{\bibinfo{person}{Grasiele Duarte}, \bibinfo{person}{Afonso Lemonge}, {and} \bibinfo{person}{Leonardo Goliatt}.} \bibinfo{year}{2017}\natexlab{}.
\newblock \showarticletitle{A dynamic migration policy to the island model}. In \bibinfo{booktitle}{\emph{2017 IEEE Congress on evolutionary computation (CEC)}}. IEEE, \bibinfo{pages}{1135--1142}.
\newblock


\bibitem[Duarte et~al\mbox{.}(2021)]%
        {Stigmergy}
\bibfield{author}{\bibinfo{person}{Grasiele~Regina Duarte}, \bibinfo{person}{Afonso~Celso de Castro~Lemonge}, \bibinfo{person}{Leonardo~Goliatt da Fonseca}, {and} \bibinfo{person}{Beatriz Souza Leite~Pires de Lima}.} \bibinfo{year}{2021}\natexlab{}.
\newblock \showarticletitle{An Island Model based on Stigmergy to solve optimization problems}.
\newblock \bibinfo{journal}{\emph{Natural Computing}} \bibinfo{volume}{20}, \bibinfo{number}{3} (\bibinfo{year}{2021}), \bibinfo{pages}{413--441}.
\newblock


\bibitem[Dulebenets(2020)]%
        {berth_scheduling}
\bibfield{author}{\bibinfo{person}{Maxim~A Dulebenets}.} \bibinfo{year}{2020}\natexlab{}.
\newblock \showarticletitle{An Adaptive Island Evolutionary Algorithm for the berth scheduling problem}.
\newblock \bibinfo{journal}{\emph{Memetic Computing}} \bibinfo{volume}{12}, \bibinfo{number}{1} (\bibinfo{year}{2020}), \bibinfo{pages}{51--72}.
\newblock


\bibitem[El-Abd(2010)]%
        {fix_2}
\bibfield{author}{\bibinfo{person}{Mohammed El-Abd}.} \bibinfo{year}{2010}\natexlab{}.
\newblock \showarticletitle{A cooperative approach to the artificial bee colony algorithm}. In \bibinfo{booktitle}{\emph{IEEE congress on evolutionary computation}}. IEEE, \bibinfo{pages}{1--5}.
\newblock


\bibitem[El-Abd(2016)]%
        {step1}
\bibfield{author}{\bibinfo{person}{Mohammed El-Abd}.} \bibinfo{year}{2016}\natexlab{}.
\newblock \showarticletitle{Cooperative coevolution using the brain storm optimization algorithm}. In \bibinfo{booktitle}{\emph{2016 IEEE Symposium Series on Computational Intelligence (SSCI)}}. IEEE, \bibinfo{pages}{1--7}.
\newblock


\bibitem[ElHara et~al\mbox{.}(2019)]%
        {bbob-largescale}
\bibfield{author}{\bibinfo{person}{Ouassim~Ait ElHara}, \bibinfo{person}{Konstantinos Varelas}, \bibinfo{person}{Duc~Manh Nguyen}, \bibinfo{person}{Tu{\v{s}ar}}, \bibinfo{person}{Dimo Brockhoff}, \bibinfo{person}{Nikolaus Hansen}, {and} \bibinfo{person}{Anne Auger}.} \bibinfo{year}{2019}\natexlab{}.
\newblock \showarticletitle{{COCO}: The Large Scale Black-Box Optimization Benchmarking (bbob-largescale) Test Suite}.
\newblock \bibinfo{journal}{\emph{ArXiv}}  \bibinfo{volume}{abs/1903.06396} (\bibinfo{year}{2019}).
\newblock
\urldef\tempurl%
\url{https://api.semanticscholar.org/CorpusID:80628369}
\showURL{%
\tempurl}


\bibitem[Farzaneh and Mahdian~Toroghi(2020)]%
        {peole}
\bibfield{author}{\bibinfo{person}{Majid Farzaneh} {and} \bibinfo{person}{Rahil Mahdian~Toroghi}.} \bibinfo{year}{2020}\natexlab{}.
\newblock \showarticletitle{Music generation using an interactive evolutionary algorithm}. In \bibinfo{booktitle}{\emph{Pattern Recognition and Artificial Intelligence: Third Mediterranean Conference, MedPRAI 2019, Istanbul, Turkey, December 22--23, 2019, Proceedings 3}}. Springer, \bibinfo{pages}{207--217}.
\newblock


\bibitem[Federici et~al\mbox{.}(2020)]%
        {constrained}
\bibfield{author}{\bibinfo{person}{Lorenzo Federici}, \bibinfo{person}{Boris Benedikter}, {and} \bibinfo{person}{Alessandro Zavoli}.} \bibinfo{year}{2020}\natexlab{}.
\newblock \showarticletitle{EOS: a parallel, self-adaptive, multi-population evolutionary algorithm for constrained global optimization}. In \bibinfo{booktitle}{\emph{2020 IEEE Congress on Evolutionary Computation (CEC)}}. IEEE, \bibinfo{pages}{1--10}.
\newblock


\bibitem[Fu et~al\mbox{.}(2020)]%
        {SAEA-RFS}
\bibfield{author}{\bibinfo{person}{Guoxia Fu}, \bibinfo{person}{Chaoli Sun}, \bibinfo{person}{Ying Tan}, \bibinfo{person}{Guochen Zhang}, {and} \bibinfo{person}{Yaochu Jin}.} \bibinfo{year}{2020}\natexlab{}.
\newblock \showarticletitle{A surrogate-assisted evolutionary algorithm with random feature selection for large-scale expensive problems}. In \bibinfo{booktitle}{\emph{Parallel Problem Solving from Nature--PPSN XVI: 16th International Conference, PPSN 2020, Leiden, The Netherlands, September 5-9, 2020, Proceedings, Part I 16}}. Springer, \bibinfo{pages}{125--139}.
\newblock


\bibitem[Gao et~al\mbox{.}(2020)]%
        {ANN}
\bibfield{author}{\bibinfo{person}{Yuan Gao}, \bibinfo{person}{Tao Yang}, \bibinfo{person}{Serhiy Bozhko}, \bibinfo{person}{Patrick Wheeler}, {and} \bibinfo{person}{Tomislav Dragi{\v{c}}evi{\'c}}.} \bibinfo{year}{2020}\natexlab{}.
\newblock \showarticletitle{Filter design and optimization of electromechanical actuation systems using search and surrogate algorithms for more-electric aircraft applications}.
\newblock \bibinfo{journal}{\emph{IEEE Transactions on Transportation Electrification}} \bibinfo{volume}{6}, \bibinfo{number}{4} (\bibinfo{year}{2020}), \bibinfo{pages}{1434--1447}.
\newblock


\bibitem[Gong et~al\mbox{.}(2023)]%
        {CC-DDEA}
\bibfield{author}{\bibinfo{person}{Yue-Jiao Gong}, \bibinfo{person}{Yuan-Ting Zhong}, {and} \bibinfo{person}{Hao-Gan Huang}.} \bibinfo{year}{2023}\natexlab{}.
\newblock \showarticletitle{Offline Data-Driven Optimization at Scale: A Cooperative Coevolutionary Approach}.
\newblock \bibinfo{journal}{\emph{IEEE Transactions on Evolutionary Computation}} (\bibinfo{year}{2023}), \bibinfo{pages}{1--1}.
\newblock
\href{https://doi.org/10.1109/TEVC.2023.3338693}{doi:\nolinkurl{10.1109/TEVC.2023.3338693}}


\bibitem[Gozali and Fujimura(2019)]%
        {master_slave}
\bibfield{author}{\bibinfo{person}{Alfian~Akbar Gozali} {and} \bibinfo{person}{Shigeru Fujimura}.} \bibinfo{year}{2019}\natexlab{}.
\newblock \showarticletitle{Localized island model genetic algorithm in population diversity preservation}. In \bibinfo{booktitle}{\emph{2018 international conference on industrial enterprise and system engineering (IcoIESE 2018)}}. Atlantis Press, \bibinfo{pages}{122--128}.
\newblock


\bibitem[Gu et~al\mbox{.}(2023)]%
        {SADE-AMSS}
\bibfield{author}{\bibinfo{person}{Haoran Gu}, \bibinfo{person}{Handing Wang}, {and} \bibinfo{person}{Yaochu Jin}.} \bibinfo{year}{2023}\natexlab{}.
\newblock \showarticletitle{Surrogate-Assisted Differential Evolution With Adaptive Multisubspace Search for Large-Scale Expensive Optimization}.
\newblock \bibinfo{journal}{\emph{IEEE Transactions on Evolutionary Computation}} \bibinfo{volume}{27}, \bibinfo{number}{6} (\bibinfo{year}{2023}), \bibinfo{pages}{1765--1779}.
\newblock


\bibitem[Guo et~al\mbox{.}(2016)]%
        {fused_magnesium_furnaces}
\bibfield{author}{\bibinfo{person}{Dan Guo}, \bibinfo{person}{Tianyou Chai}, \bibinfo{person}{Jinliang Ding}, {and} \bibinfo{person}{Yaochu Jin}.} \bibinfo{year}{2016}\natexlab{}.
\newblock \showarticletitle{Small data driven evolutionary multi-objective optimization of fused magnesium furnaces}. In \bibinfo{booktitle}{\emph{2016 IEEE Symposium Series on Computational Intelligence (SSCI)}}. \bibinfo{pages}{1--8}.
\newblock
\href{https://doi.org/10.1109/SSCI.2016.7850211}{doi:\nolinkurl{10.1109/SSCI.2016.7850211}}


\bibitem[Han et~al\mbox{.}(2014)]%
        {high_fidelity}
\bibfield{author}{\bibinfo{person}{Hyejin Han}, \bibinfo{person}{Jounghuem Kwon}, \bibinfo{person}{Jiyong Lee}, \bibinfo{person}{Romain Destenay}, {and} \bibinfo{person}{Bum-Jae You}.} \bibinfo{year}{2014}\natexlab{}.
\newblock \showarticletitle{Real-time optimization for the high-fidelity of human motion imitation}. In \bibinfo{booktitle}{\emph{2014 11th International Conference on Ubiquitous Robots and Ambient Intelligence (URAI)}}. IEEE, \bibinfo{pages}{692--695}.
\newblock


\bibitem[Hansen(2019)]%
        {lq-CMA-ES}
\bibfield{author}{\bibinfo{person}{Nikolaus Hansen}.} \bibinfo{year}{2019}\natexlab{}.
\newblock \showarticletitle{A global surrogate assisted {CMA-ES}}. In \bibinfo{booktitle}{\emph{Proceedings of the Genetic and Evolutionary Computation Conference}} (Prague, Czech Republic). \bibinfo{address}{New York, NY, USA}, \bibinfo{pages}{664–672}.
\newblock
\urldef\tempurl%
\url{https://doi.org/10.1145/3321707.3321842}
\showURL{%
\tempurl}


\bibitem[Hansen et~al\mbox{.}(2021)]%
        {COCO}
\bibfield{author}{\bibinfo{person}{N. Hansen}, \bibinfo{person}{A. Auger}, \bibinfo{person}{R. Ros}, \bibinfo{person}{O. Mersmann}, \bibinfo{person}{T. Tu{\v s}ar}, {and} \bibinfo{person}{D. Brockhoff}.} \bibinfo{year}{2021}\natexlab{}.
\newblock \showarticletitle{{COCO}: A Platform for Comparing Continuous Optimizers in a Black-Box Setting}.
\newblock \bibinfo{journal}{\emph{Optimization Methods and Software}}  \bibinfo{volume}{36} (\bibinfo{year}{2021}), \bibinfo{pages}{114--144}.
\newblock
Issue 1.
\href{https://doi.org/10.1080/10556788.2020.1808977}{doi:\nolinkurl{10.1080/10556788.2020.1808977}}


\bibitem[Harpham et~al\mbox{.}(2004)]%
        {rbfn_gc}
\bibfield{author}{\bibinfo{person}{C Harpham}, \bibinfo{person}{Christian~W Dawson}, {and} \bibinfo{person}{Martin~R Brown}.} \bibinfo{year}{2004}\natexlab{}.
\newblock \showarticletitle{A review of genetic algorithms applied to training radial basis function networks}.
\newblock \bibinfo{journal}{\emph{Neural Computing \& Applications}}  \bibinfo{volume}{13} (\bibinfo{year}{2004}), \bibinfo{pages}{193--201}.
\newblock


\bibitem[Hasanzadeh et~al\mbox{.}(2013)]%
        {hasanzadeh2013adaptive}
\bibfield{author}{\bibinfo{person}{Mohammad Hasanzadeh}, \bibinfo{person}{Mohammad~Reza Meybodi}, {and} \bibinfo{person}{Mohammad~Mehdi Ebadzadeh}.} \bibinfo{year}{2013}\natexlab{}.
\newblock \showarticletitle{Adaptive cooperative particle swarm optimizer}.
\newblock \bibinfo{journal}{\emph{Applied Intelligence}}  \bibinfo{volume}{39} (\bibinfo{year}{2013}), \bibinfo{pages}{397--420}.
\newblock


\bibitem[He et~al\mbox{.}(2023)]%
        {SAEAs}
\bibfield{author}{\bibinfo{person}{Chunlin He}, \bibinfo{person}{Yong Zhang}, \bibinfo{person}{Dunwei Gong}, {and} \bibinfo{person}{Xinfang Ji}.} \bibinfo{year}{2023}\natexlab{}.
\newblock \showarticletitle{A review of surrogate-assisted evolutionary algorithms for expensive optimization problems}.
\newblock \bibinfo{journal}{\emph{Expert Systems with Applications}} (\bibinfo{year}{2023}), \bibinfo{pages}{119495}.
\newblock


\bibitem[Hu et~al\mbox{.}(2017)]%
        {affect1}
\bibfield{author}{\bibinfo{person}{Xiao-Min Hu}, \bibinfo{person}{Fei-Long He}, \bibinfo{person}{Wei-Neng Chen}, {and} \bibinfo{person}{Jun Zhang}.} \bibinfo{year}{2017}\natexlab{}.
\newblock \showarticletitle{Cooperation coevolution with fast interdependency identification for large scale optimization}.
\newblock \bibinfo{journal}{\emph{Information Sciences}}  \bibinfo{volume}{381} (\bibinfo{year}{2017}), \bibinfo{pages}{142--160}.
\newblock


\bibitem[Huang and Gong(2022)]%
        {CL-DDEA}
\bibfield{author}{\bibinfo{person}{Hao-Gan Huang} {and} \bibinfo{person}{Yue-Jiao Gong}.} \bibinfo{year}{2022}\natexlab{}.
\newblock \showarticletitle{Contrastive learning: An alternative surrogate for offline data-driven evolutionary computation}.
\newblock \bibinfo{journal}{\emph{IEEE Transactions on Evolutionary Computation}} \bibinfo{volume}{27}, \bibinfo{number}{2} (\bibinfo{year}{2022}), \bibinfo{pages}{370--384}.
\newblock


\bibitem[Huang et~al\mbox{.}(2021)]%
        {TT-DDEA}
\bibfield{author}{\bibinfo{person}{Pengfei Huang}, \bibinfo{person}{Handing Wang}, {and} \bibinfo{person}{Yaochu Jin}.} \bibinfo{year}{2021}\natexlab{}.
\newblock \showarticletitle{Offline data-driven evolutionary optimization based on tri-training}.
\newblock \bibinfo{journal}{\emph{Swarm and evolutionary computation}}  \bibinfo{volume}{60} (\bibinfo{year}{2021}), \bibinfo{pages}{100800}.
\newblock


\bibitem[Huang et~al\mbox{.}(2019)]%
        {SRK-DDEA}
\bibfield{author}{\bibinfo{person}{Pengfei Huang}, \bibinfo{person}{Handing Wang}, {and} \bibinfo{person}{Wenping Ma}.} \bibinfo{year}{2019}\natexlab{}.
\newblock \showarticletitle{Stochastic ranking for offline data-driven evolutionary optimization using radial basis function networks with multiple kernels}. In \bibinfo{booktitle}{\emph{2019 IEEE Symposium Series on Computational Intelligence (SSCI)}}. IEEE, \bibinfo{pages}{2050--2057}.
\newblock


\bibitem[Husain and Kim(2010)]%
        {ensemble_d}
\bibfield{author}{\bibinfo{person}{Afzal Husain} {and} \bibinfo{person}{Kwang-Yong Kim}.} \bibinfo{year}{2010}\natexlab{}.
\newblock \showarticletitle{Enhanced multi-objective optimization of a microchannel heat sink through evolutionary algorithm coupled with multiple surrogate models}.
\newblock \bibinfo{journal}{\emph{Applied Thermal Engineering}} \bibinfo{volume}{30}, \bibinfo{number}{13} (\bibinfo{year}{2010}), \bibinfo{pages}{1683--1691}.
\newblock


\bibitem[Jin(2011)]%
        {SAEA_DDEA}
\bibfield{author}{\bibinfo{person}{Yaochu Jin}.} \bibinfo{year}{2011}\natexlab{}.
\newblock \showarticletitle{Surrogate-assisted evolutionary computation: Recent advances and future challenges}.
\newblock \bibinfo{journal}{\emph{Swarm and Evolutionary Computation}} \bibinfo{volume}{1}, \bibinfo{number}{2} (\bibinfo{year}{2011}), \bibinfo{pages}{61--70}.
\newblock


\bibitem[Jin et~al\mbox{.}(2018)]%
        {DDEA}
\bibfield{author}{\bibinfo{person}{Yaochu Jin}, \bibinfo{person}{Handing Wang}, \bibinfo{person}{Tinkle Chugh}, \bibinfo{person}{Dan Guo}, {and} \bibinfo{person}{Kaisa Miettinen}.} \bibinfo{year}{2018}\natexlab{}.
\newblock \showarticletitle{Data-driven evolutionary optimization: An overview and case studies}.
\newblock \bibinfo{journal}{\emph{IEEE Transactions on Evolutionary Computation}} \bibinfo{volume}{23}, \bibinfo{number}{3} (\bibinfo{year}{2018}), \bibinfo{pages}{442--458}.
\newblock


\bibitem[Kropp et~al\mbox{.}(2023)]%
        {SBX}
\bibfield{author}{\bibinfo{person}{Ian Kropp}, \bibinfo{person}{A~Pouyan Nejadhashemi}, {and} \bibinfo{person}{Kalyanmoy Deb}.} \bibinfo{year}{2023}\natexlab{}.
\newblock \showarticletitle{Improved Evolutionary Operators for Sparse Large-Scale Multiobjective Optimization Problems}.
\newblock \bibinfo{journal}{\emph{IEEE Transactions on Evolutionary Computation}} (\bibinfo{year}{2023}).
\newblock


\bibitem[Ku et~al\mbox{.}(2023)]%
        {offline_DDEA}
\bibfield{author}{\bibinfo{person}{Junhua Ku}, \bibinfo{person}{Huixiang Zhen}, {and} \bibinfo{person}{Wenyin Gong}.} \bibinfo{year}{2023}\natexlab{}.
\newblock \showarticletitle{Offline data-driven optimization based on dual-scale surrogate ensemble}.
\newblock \bibinfo{journal}{\emph{Memetic Computing}} \bibinfo{volume}{15}, \bibinfo{number}{2} (\bibinfo{year}{2023}), \bibinfo{pages}{139--154}.
\newblock


\bibitem[Kumar et~al\mbox{.}(2022)]%
        {Hardware_Accelerators}
\bibfield{author}{\bibinfo{person}{Aviral Kumar}, \bibinfo{person}{Amir Yazdanbakhsh}, \bibinfo{person}{Milad Hashemi}, \bibinfo{person}{Kevin Swersky}, {and} \bibinfo{person}{Sergey Levine}.} \bibinfo{year}{2022}\natexlab{}.
\newblock \showarticletitle{Data-Driven Offline Optimization for Architecting Hardware Accelerators}. In \bibinfo{booktitle}{\emph{International Conference on Learning Representations}}.
\newblock
\urldef\tempurl%
\url{https://openreview.net/forum?id=GsH-K1VIyy}
\showURL{%
\tempurl}


\bibitem[Kushida et~al\mbox{.}(2013)]%
        {control_parameters}
\bibfield{author}{\bibinfo{person}{Jun-ichi Kushida}, \bibinfo{person}{Akira Hara}, \bibinfo{person}{Tetsuyuki Takahama}, {and} \bibinfo{person}{Ayumi Kido}.} \bibinfo{year}{2013}\natexlab{}.
\newblock \showarticletitle{Island-based differential evolution with varying subpopulation size}. In \bibinfo{booktitle}{\emph{2013 IEEE 6th international workshop on computational intelligence and applications (IWCIA)}}. IEEE, \bibinfo{pages}{119--124}.
\newblock


\bibitem[Li et~al\mbox{.}(2020a)]%
        {SAMSO}
\bibfield{author}{\bibinfo{person}{Fan Li}, \bibinfo{person}{Xiwen Cai}, \bibinfo{person}{Liang Gao}, {and} \bibinfo{person}{Weiming Shen}.} \bibinfo{year}{2020}\natexlab{a}.
\newblock \showarticletitle{A surrogate-assisted multiswarm optimization algorithm for high-dimensional computationally expensive problems}.
\newblock \bibinfo{journal}{\emph{IEEE transactions on cybernetics}} \bibinfo{volume}{51}, \bibinfo{number}{3} (\bibinfo{year}{2020}), \bibinfo{pages}{1390--1402}.
\newblock


\bibitem[Li et~al\mbox{.}(2022)]%
        {decomposition}
\bibfield{author}{\bibinfo{person}{Jian-Yu Li}, \bibinfo{person}{Zhi-Hui Zhan}, \bibinfo{person}{Kay~Chen Tan}, {and} \bibinfo{person}{Jun Zhang}.} \bibinfo{year}{2022}\natexlab{}.
\newblock \showarticletitle{Dual differential grouping: A more general decomposition method for large-scale optimization}.
\newblock \bibinfo{journal}{\emph{IEEE Transactions on Cybernetics}} (\bibinfo{year}{2022}).
\newblock


\bibitem[Li et~al\mbox{.}(2020c)]%
        {BDDEA-LDG}
\bibfield{author}{\bibinfo{person}{Jian-Yu Li}, \bibinfo{person}{Zhi-Hui Zhan}, \bibinfo{person}{Chuan Wang}, \bibinfo{person}{Hu Jin}, {and} \bibinfo{person}{Jun Zhang}.} \bibinfo{year}{2020}\natexlab{c}.
\newblock \showarticletitle{Boosting data-driven evolutionary algorithm with localized data generation}.
\newblock \bibinfo{journal}{\emph{IEEE Transactions on Evolutionary Computation}} \bibinfo{volume}{24}, \bibinfo{number}{5} (\bibinfo{year}{2020}), \bibinfo{pages}{923--937}.
\newblock


\bibitem[Li et~al\mbox{.}(2020b)]%
        {DDEA-PES}
\bibfield{author}{\bibinfo{person}{Jian-Yu Li}, \bibinfo{person}{Zhi-Hui Zhan}, \bibinfo{person}{Hua Wang}, {and} \bibinfo{person}{Jun Zhang}.} \bibinfo{year}{2020}\natexlab{b}.
\newblock \showarticletitle{Data-driven evolutionary algorithm with perturbation-based ensemble surrogates}.
\newblock \bibinfo{journal}{\emph{IEEE Transactions on Cybernetics}} \bibinfo{volume}{51}, \bibinfo{number}{8} (\bibinfo{year}{2020}), \bibinfo{pages}{3925--3937}.
\newblock


\bibitem[Lim et~al\mbox{.}(2009)]%
        {ensemble_c}
\bibfield{author}{\bibinfo{person}{Dudy Lim}, \bibinfo{person}{Yaochu Jin}, \bibinfo{person}{Yew-Soon Ong}, {and} \bibinfo{person}{Bernhard Sendhoff}.} \bibinfo{year}{2009}\natexlab{}.
\newblock \showarticletitle{Generalizing surrogate-assisted evolutionary computation}.
\newblock \bibinfo{journal}{\emph{IEEE Transactions on Evolutionary Computation}} \bibinfo{volume}{14}, \bibinfo{number}{3} (\bibinfo{year}{2009}), \bibinfo{pages}{329--355}.
\newblock


\bibitem[Lim et~al\mbox{.}(2007)]%
        {ensemble_b}
\bibfield{author}{\bibinfo{person}{Dudy Lim}, \bibinfo{person}{Yew-Soon Ong}, \bibinfo{person}{Yaochu Jin}, {and} \bibinfo{person}{Bernhard Sendhoff}.} \bibinfo{year}{2007}\natexlab{}.
\newblock \showarticletitle{A study on metamodeling techniques, ensembles, and multi-surrogates in evolutionary computation}. In \bibinfo{booktitle}{\emph{Proceedings of the 9th annual conference on Genetic and evolutionary computation}}. \bibinfo{pages}{1288--1295}.
\newblock


\bibitem[Lin et~al\mbox{.}(2022)]%
        {ensemble6}
\bibfield{author}{\bibinfo{person}{Qiuzhen Lin}, \bibinfo{person}{Xunfeng Wu}, \bibinfo{person}{Lijia Ma}, \bibinfo{person}{Jianqiang Li}, \bibinfo{person}{Maoguo Gong}, {and} \bibinfo{person}{Carlos A.~Coello Coello}.} \bibinfo{year}{2022}\natexlab{}.
\newblock \showarticletitle{An Ensemble Surrogate-Based Framework for Expensive Multiobjective Evolutionary Optimization}.
\newblock \bibinfo{journal}{\emph{IEEE Transactions on Evolutionary Computation}} \bibinfo{volume}{26}, \bibinfo{number}{4} (\bibinfo{year}{2022}), \bibinfo{pages}{631--645}.
\newblock
\href{https://doi.org/10.1109/TEVC.2021.3103936}{doi:\nolinkurl{10.1109/TEVC.2021.3103936}}


\bibitem[Liu et~al\mbox{.}(2022)]%
        {SAEA-RPG}
\bibfield{author}{\bibinfo{person}{Shulei Liu}, \bibinfo{person}{Handing Wang}, \bibinfo{person}{Wei Peng}, {and} \bibinfo{person}{Wen Yao}.} \bibinfo{year}{2022}\natexlab{}.
\newblock \showarticletitle{A surrogate-assisted evolutionary feature selection algorithm with parallel random grouping for high-dimensional classification}.
\newblock \bibinfo{journal}{\emph{IEEE Transactions on Evolutionary Computation}} \bibinfo{volume}{26}, \bibinfo{number}{5} (\bibinfo{year}{2022}), \bibinfo{pages}{1087--1101}.
\newblock


\bibitem[Liu et~al\mbox{.}(2021)]%
        {online_DDEA}
\bibfield{author}{\bibinfo{person}{Xiao Liu}, \bibinfo{person}{Chunfu Hu}, \bibinfo{person}{Xiongsong Li}, \bibinfo{person}{Jian Gao}, {and} \bibinfo{person}{Shoudao Huang}.} \bibinfo{year}{2021}\natexlab{}.
\newblock \showarticletitle{An online data-driven multi-objective optimization of a permanent magnet linear synchronous motor}.
\newblock \bibinfo{journal}{\emph{IEEE Transactions on Magnetics}} \bibinfo{volume}{57}, \bibinfo{number}{7} (\bibinfo{year}{2021}), \bibinfo{pages}{1--4}.
\newblock


\bibitem[Liu and Wang(2022)]%
        {Kriging-assisted}
\bibfield{author}{\bibinfo{person}{Zhening Liu} {and} \bibinfo{person}{Handing Wang}.} \bibinfo{year}{2022}\natexlab{}.
\newblock \showarticletitle{A data augmentation based Kriging-assisted reference vector guided evolutionary algorithm for expensive dynamic multi-objective optimization}.
\newblock \bibinfo{journal}{\emph{Swarm and Evolutionary Computation}}  \bibinfo{volume}{75} (\bibinfo{year}{2022}), \bibinfo{pages}{101173}.
\newblock


\bibitem[Lynn et~al\mbox{.}(2018)]%
        {island_cate2}
\bibfield{author}{\bibinfo{person}{Nandar Lynn}, \bibinfo{person}{Mostafa~Z Ali}, {and} \bibinfo{person}{Ponnuthurai~Nagaratnam Suganthan}.} \bibinfo{year}{2018}\natexlab{}.
\newblock \showarticletitle{Population topologies for particle swarm optimization and differential evolution}.
\newblock \bibinfo{journal}{\emph{Swarm and evolutionary computation}}  \bibinfo{volume}{39} (\bibinfo{year}{2018}), \bibinfo{pages}{24--35}.
\newblock


\bibitem[Ma et~al\mbox{.}(2021)]%
        {refer_v}
\bibfield{author}{\bibinfo{person}{Lianbo Ma}, \bibinfo{person}{Min Huang}, \bibinfo{person}{Shengxiang Yang}, \bibinfo{person}{Rui Wang}, {and} \bibinfo{person}{Xingwei Wang}.} \bibinfo{year}{2021}\natexlab{}.
\newblock \showarticletitle{An adaptive localized decision variable analysis approach to large-scale multiobjective and many-objective optimization}.
\newblock \bibinfo{journal}{\emph{IEEE Transactions on Cybernetics}} \bibinfo{volume}{52}, \bibinfo{number}{7} (\bibinfo{year}{2021}), \bibinfo{pages}{6684--6696}.
\newblock


\bibitem[Ma et~al\mbox{.}(2022)]%
        {MDG}
\bibfield{author}{\bibinfo{person}{Xiaoliang Ma}, \bibinfo{person}{Zhitao Huang}, \bibinfo{person}{Xiaodong Li}, \bibinfo{person}{Lei Wang}, \bibinfo{person}{Yutao Qi}, {and} \bibinfo{person}{Zexuan Zhu}.} \bibinfo{year}{2022}\natexlab{}.
\newblock \showarticletitle{Merged differential grouping for large-scale global optimization}.
\newblock \bibinfo{journal}{\emph{IEEE Transactions on Evolutionary Computation}} \bibinfo{volume}{26}, \bibinfo{number}{6} (\bibinfo{year}{2022}), \bibinfo{pages}{1439--1451}.
\newblock


\bibitem[Ma et~al\mbox{.}(2018)]%
        {CCEA}
\bibfield{author}{\bibinfo{person}{Xiaoliang Ma}, \bibinfo{person}{Xiaodong Li}, \bibinfo{person}{Qingfu Zhang}, \bibinfo{person}{Ke Tang}, \bibinfo{person}{Zhengping Liang}, \bibinfo{person}{Weixin Xie}, {and} \bibinfo{person}{Zexuan Zhu}.} \bibinfo{year}{2018}\natexlab{}.
\newblock \showarticletitle{A survey on cooperative co-evolutionary algorithms}.
\newblock \bibinfo{journal}{\emph{IEEE Transactions on Evolutionary Computation}} \bibinfo{volume}{23}, \bibinfo{number}{3} (\bibinfo{year}{2018}), \bibinfo{pages}{421--441}.
\newblock


\bibitem[Ma et~al\mbox{.}(2023)]%
        {MetaBox}
\bibfield{author}{\bibinfo{person}{Zeyuan Ma}, \bibinfo{person}{Hongshu Guo}, \bibinfo{person}{Jiacheng Chen}, \bibinfo{person}{Zhenrui Li}, \bibinfo{person}{Guojun Peng}, \bibinfo{person}{Yue-Jiao Gong}, \bibinfo{person}{Yining Ma}, {and} \bibinfo{person}{Zhiguang Cao}.} \bibinfo{year}{2023}\natexlab{}.
\newblock \showarticletitle{{MetaBox}: A Benchmark Platform for Meta-Black-Box Optimization with Reinforcement Learning}. In \bibinfo{booktitle}{\emph{Advances in Neural Information Processing Systems}}, \bibfield{editor}{\bibinfo{person}{A.~Oh}, \bibinfo{person}{T.~Naumann}, \bibinfo{person}{A.~Globerson}, \bibinfo{person}{K.~Saenko}, \bibinfo{person}{M.~Hardt}, {and} \bibinfo{person}{S.~Levine}} (Eds.), Vol.~\bibinfo{volume}{36}. \bibinfo{publisher}{Curran Associates, Inc.}, \bibinfo{pages}{10775--10795}.
\newblock
\urldef\tempurl%
\url{https://proceedings.neurips.cc/paper_files/paper/2023/file/232eee8ef411a0a316efa298d7be3c2b-Paper-Datasets_and_Benchmarks.pdf}
\showURL{%
\tempurl}


\bibitem[Mahdavi et~al\mbox{.}(2015)]%
        {step2}
\bibfield{author}{\bibinfo{person}{Sedigheh Mahdavi}, \bibinfo{person}{Mohammad~Ebrahim Shiri}, {and} \bibinfo{person}{Shahryar Rahnamayan}.} \bibinfo{year}{2015}\natexlab{}.
\newblock \showarticletitle{Metaheuristics in large-scale global continues optimization: A survey}.
\newblock \bibinfo{journal}{\emph{Information Sciences}}  \bibinfo{volume}{295} (\bibinfo{year}{2015}), \bibinfo{pages}{407--428}.
\newblock


\bibitem[Mei et~al\mbox{.}(2013)]%
        {rdg}
\bibfield{author}{\bibinfo{person}{Yi Mei}, \bibinfo{person}{Xiaodong Li}, {and} \bibinfo{person}{Xin Yao}.} \bibinfo{year}{2013}\natexlab{}.
\newblock \showarticletitle{Cooperative coevolution with route distance grouping for large-scale capacitated arc routing problems}.
\newblock \bibinfo{journal}{\emph{IEEE Transactions on Evolutionary Computation}} \bibinfo{volume}{18}, \bibinfo{number}{3} (\bibinfo{year}{2013}), \bibinfo{pages}{435--449}.
\newblock


\bibitem[Mohapatra et~al\mbox{.}(2017)]%
        {CSO}
\bibfield{author}{\bibinfo{person}{Prabhujit Mohapatra}, \bibinfo{person}{Kedar~Nath Das}, {and} \bibinfo{person}{Santanu Roy}.} \bibinfo{year}{2017}\natexlab{}.
\newblock \showarticletitle{A modified competitive swarm optimizer for large scale optimization problems}.
\newblock \bibinfo{journal}{\emph{Applied Soft Computing}}  \bibinfo{volume}{59} (\bibinfo{year}{2017}), \bibinfo{pages}{340--362}.
\newblock


\bibitem[Nadimi-Shahraki et~al\mbox{.}(2021)]%
        {feature5}
\bibfield{author}{\bibinfo{person}{Mohammad~H. Nadimi-Shahraki}, \bibinfo{person}{Mahdis Banaie-Dezfouli}, \bibinfo{person}{Hoda Zamani}, \bibinfo{person}{Shokooh Taghian}, {and} \bibinfo{person}{Seyedali Mirjalili}.} \bibinfo{year}{2021}\natexlab{}.
\newblock \showarticletitle{B-MFO: A Binary Moth-Flame Optimization for Feature Selection from Medical Datasets}.
\newblock \bibinfo{journal}{\emph{Computers}} \bibinfo{volume}{10}, \bibinfo{number}{11} (\bibinfo{year}{2021}).
\newblock
\showISSN{2073-431X}
\href{https://doi.org/10.3390/computers10110136}{doi:\nolinkurl{10.3390/computers10110136}}


\bibitem[Nadimi-Shahraki and Zamani(2022)]%
        {dmde}
\bibfield{author}{\bibinfo{person}{Mohammad~H Nadimi-Shahraki} {and} \bibinfo{person}{Hoda Zamani}.} \bibinfo{year}{2022}\natexlab{}.
\newblock \showarticletitle{DMDE: Diversity-maintained multi-trial vector differential evolution algorithm for non-decomposition large-scale global optimization}.
\newblock \bibinfo{journal}{\emph{Expert Systems with Applications}}  \bibinfo{volume}{198} (\bibinfo{year}{2022}), \bibinfo{pages}{116895}.
\newblock


\bibitem[Omidvar et~al\mbox{.}(2013)]%
        {DG}
\bibfield{author}{\bibinfo{person}{Mohammad~Nabi Omidvar}, \bibinfo{person}{Xiaodong Li}, \bibinfo{person}{Yi Mei}, {and} \bibinfo{person}{Xin Yao}.} \bibinfo{year}{2013}\natexlab{}.
\newblock \showarticletitle{Cooperative co-evolution with differential grouping for large scale optimization}.
\newblock \bibinfo{journal}{\emph{IEEE Transactions on evolutionary computation}} \bibinfo{volume}{18}, \bibinfo{number}{3} (\bibinfo{year}{2013}), \bibinfo{pages}{378--393}.
\newblock


\bibitem[Omidvar et~al\mbox{.}(2017)]%
        {dg2}
\bibfield{author}{\bibinfo{person}{Mohammad~Nabi Omidvar}, \bibinfo{person}{Ming Yang}, \bibinfo{person}{Yi Mei}, \bibinfo{person}{Xiaodong Li}, {and} \bibinfo{person}{Xin Yao}.} \bibinfo{year}{2017}\natexlab{}.
\newblock \showarticletitle{DG2: A faster and more accurate differential grouping for large-scale black-box optimization}.
\newblock \bibinfo{journal}{\emph{IEEE Transactions on Evolutionary Computation}} \bibinfo{volume}{21}, \bibinfo{number}{6} (\bibinfo{year}{2017}), \bibinfo{pages}{929--942}.
\newblock


\bibitem[Qian and Yu(2016)]%
        {feature10}
\bibfield{author}{\bibinfo{person}{Hong Qian} {and} \bibinfo{person}{Yang Yu}.} \bibinfo{year}{2016}\natexlab{}.
\newblock \showarticletitle{Scaling simultaneous optimistic optimization for high-dimensional non-convex functions with low effective dimensions}. In \bibinfo{booktitle}{\emph{Proceedings of the AAAI Conference on Artificial Intelligence}}, Vol.~\bibinfo{volume}{30}.
\newblock


\bibitem[Ren et~al\mbox{.}(2021)]%
        {depends}
\bibfield{author}{\bibinfo{person}{Zhigang Ren}, \bibinfo{person}{Yongsheng Liang}, \bibinfo{person}{Muyi Wang}, \bibinfo{person}{Yang Yang}, {and} \bibinfo{person}{An Chen}.} \bibinfo{year}{2021}\natexlab{}.
\newblock \showarticletitle{An eigenspace divide-and-conquer approach for large-scale optimization}.
\newblock \bibinfo{journal}{\emph{Applied Soft Computing}}  \bibinfo{volume}{99} (\bibinfo{year}{2021}), \bibinfo{pages}{106911}.
\newblock


\bibitem[Seghouane and Shokouhi(2019)]%
        {RBFN1}
\bibfield{author}{\bibinfo{person}{Abd-Krim Seghouane} {and} \bibinfo{person}{Navid Shokouhi}.} \bibinfo{year}{2019}\natexlab{}.
\newblock \showarticletitle{Adaptive learning for robust radial basis function networks}.
\newblock \bibinfo{journal}{\emph{IEEE Transactions on Cybernetics}} \bibinfo{volume}{51}, \bibinfo{number}{5} (\bibinfo{year}{2019}), \bibinfo{pages}{2847--2856}.
\newblock


\bibitem[Shan et~al\mbox{.}(2021)]%
        {DDEA-SE}
\bibfield{author}{\bibinfo{person}{Yawen Shan}, \bibinfo{person}{Yongjin Hou}, \bibinfo{person}{Mengzhen Wang}, {and} \bibinfo{person}{Fei Xu}.} \bibinfo{year}{2021}\natexlab{}.
\newblock \showarticletitle{Trimmed data-driven evolutionary optimization using selective surrogate ensembles}. In \bibinfo{booktitle}{\emph{Bio-Inspired Computing: Theories and Applications: 15th International Conference, BIC-TA 2020, Qingdao, China, October 23-25, 2020, Revised Selected Papers 15}}. Springer, \bibinfo{pages}{106--116}.
\newblock


\bibitem[Shao and Chen(2009)]%
        {fix_1}
\bibfield{author}{\bibinfo{person}{Yichuan Shao} {and} \bibinfo{person}{Hanning Chen}.} \bibinfo{year}{2009}\natexlab{}.
\newblock \showarticletitle{Cooperative bacterial foraging optimization}. In \bibinfo{booktitle}{\emph{2009 International Conference on Future BioMedical Information Engineering (FBIE)}}. IEEE, \bibinfo{pages}{486--488}.
\newblock


\bibitem[Shen et~al\mbox{.}(2011)]%
        {rbfn_cluster}
\bibfield{author}{\bibinfo{person}{Wei Shen}, \bibinfo{person}{Xiaopen Guo}, \bibinfo{person}{Chao Wu}, {and} \bibinfo{person}{Desheng Wu}.} \bibinfo{year}{2011}\natexlab{}.
\newblock \showarticletitle{Forecasting stock indices using radial basis function neural networks optimized by artificial fish swarm algorithm}.
\newblock \bibinfo{journal}{\emph{Knowledge-Based Systems}} \bibinfo{volume}{24}, \bibinfo{number}{3} (\bibinfo{year}{2011}), \bibinfo{pages}{378--385}.
\newblock


\bibitem[Song et~al\mbox{.}(2023a)]%
        {Ceramic_Formula}
\bibfield{author}{\bibinfo{person}{Wen-Xiang Song}, \bibinfo{person}{Wei-Neng Chen}, {and} \bibinfo{person}{Ya-Hui Jia}.} \bibinfo{year}{2023}\natexlab{a}.
\newblock \showarticletitle{An Interactive Evolutionary Algorithm for Ceramic Formula Design}. In \bibinfo{booktitle}{\emph{Neural Information Processing: 30th International Conference, ICONIP 2023, Changsha, China, November 20–23, 2023, Proceedings, Part I}}. \bibinfo{publisher}{Springer-Verlag}, \bibinfo{pages}{381–394}.
\newblock
\href{https://doi.org/10.1007/978-981-99-8079-6_30}{doi:\nolinkurl{10.1007/978-981-99-8079-6_30}}


\bibitem[Song et~al\mbox{.}(2023b)]%
        {ensemble8}
\bibfield{author}{\bibinfo{person}{Xianfang Song}, \bibinfo{person}{Yong Zhang}, \bibinfo{person}{Dunwei Gong}, \bibinfo{person}{Hui Liu}, {and} \bibinfo{person}{Wanqiu Zhang}.} \bibinfo{year}{2023}\natexlab{b}.
\newblock \showarticletitle{Surrogate Sample-Assisted Particle Swarm Optimization for Feature Selection on High-Dimensional Data}.
\newblock \bibinfo{journal}{\emph{IEEE Transactions on Evolutionary Computation}} \bibinfo{volume}{27}, \bibinfo{number}{3} (\bibinfo{year}{2023}), \bibinfo{pages}{595--609}.
\newblock
\href{https://doi.org/10.1109/TEVC.2022.3175226}{doi:\nolinkurl{10.1109/TEVC.2022.3175226}}


\bibitem[Sun et~al\mbox{.}(2017)]%
        {ensemble2}
\bibfield{author}{\bibinfo{person}{Chaoli Sun}, \bibinfo{person}{Yaochu Jin}, \bibinfo{person}{Ran Cheng}, \bibinfo{person}{Jinliang Ding}, {and} \bibinfo{person}{Jianchao Zeng}.} \bibinfo{year}{2017}\natexlab{}.
\newblock \showarticletitle{Surrogate-assisted cooperative swarm optimization of high-dimensional expensive problems}.
\newblock \bibinfo{journal}{\emph{IEEE Transactions on Evolutionary Computation}} \bibinfo{volume}{21}, \bibinfo{number}{4} (\bibinfo{year}{2017}), \bibinfo{pages}{644--660}.
\newblock


\bibitem[Sun et~al\mbox{.}(2019)]%
        {feature3}
\bibfield{author}{\bibinfo{person}{Yongjun Sun}, \bibinfo{person}{Tong Yang}, {and} \bibinfo{person}{Zujun Liu}.} \bibinfo{year}{2019}\natexlab{}.
\newblock \showarticletitle{A whale optimization algorithm based on quadratic interpolation for high-dimensional global optimization problems}.
\newblock \bibinfo{journal}{\emph{Applied Soft Computing}}  \bibinfo{volume}{85} (\bibinfo{year}{2019}), \bibinfo{pages}{105744}.
\newblock
\showISSN{1568-4946}
\href{https://doi.org/10.1016/j.asoc.2019.105744}{doi:\nolinkurl{10.1016/j.asoc.2019.105744}}


\bibitem[Taghian et~al\mbox{.}(2018)]%
        {feature4}
\bibfield{author}{\bibinfo{person}{Shokooh Taghian}, \bibinfo{person}{Mohammad~H Nadimi-Shahraki}, {and} \bibinfo{person}{Hoda Zamani}.} \bibinfo{year}{2018}\natexlab{}.
\newblock \showarticletitle{Comparative analysis of transfer function-based binary Metaheuristic algorithms for feature selection}. In \bibinfo{booktitle}{\emph{2018 International Conference on Artificial Intelligence and Data Processing (IDAP)}}. IEEE, \bibinfo{pages}{1--6}.
\newblock


\bibitem[Tang et~al\mbox{.}(2010)]%
        {CEC2010}
\bibfield{author}{\bibinfo{person}{Ke Tang}, \bibinfo{person}{Xiaodong Li}, \bibinfo{person}{Ponnuthurai~Nagaratnam Suganthan}, \bibinfo{person}{Zhenyu Yang}, {and} \bibinfo{person}{Thomas Weise}.} \bibinfo{year}{2010}\natexlab{}.
\newblock \bibinfo{booktitle}{\emph{{Benchmark Functions for the CEC'2010 Special Session and Competition on Large-Scale Global Optimization}}}.
\newblock \bibinfo{type}{{T}echnical {R}eport}.
\newblock
\urldef\tempurl%
\url{http://www.it-weise.de/documents/files/TLSYW2009BFFTCSSACOLSGO.pdf}
\showURL{%
\tempurl}


\bibitem[Turgut et~al\mbox{.}(2020)]%
        {crow_search}
\bibfield{author}{\bibinfo{person}{Mert~Sinan Turgut}, \bibinfo{person}{Oguz~Emrah Turgut}, {and} \bibinfo{person}{Deniz~T{\"u}rsel Eliiyi}.} \bibinfo{year}{2020}\natexlab{}.
\newblock \showarticletitle{Island-based crow search algorithm for solving optimal control problems}.
\newblock \bibinfo{journal}{\emph{Applied Soft Computing}}  \bibinfo{volume}{90} (\bibinfo{year}{2020}), \bibinfo{pages}{106170}.
\newblock


\bibitem[Wang and Jin(2018)]%
        {trauma_systems}
\bibfield{author}{\bibinfo{person}{Handing Wang} {and} \bibinfo{person}{Yaochu Jin}.} \bibinfo{year}{2018}\natexlab{}.
\newblock \showarticletitle{A random forest-assisted evolutionary algorithm for data-driven constrained multiobjective combinatorial optimization of trauma systems}.
\newblock \bibinfo{journal}{\emph{IEEE transactions on cybernetics}} \bibinfo{volume}{50}, \bibinfo{number}{2} (\bibinfo{year}{2018}), \bibinfo{pages}{536--549}.
\newblock


\bibitem[Wang et~al\mbox{.}(2017)]%
        {ensemble1}
\bibfield{author}{\bibinfo{person}{Handing Wang}, \bibinfo{person}{Yaochu Jin}, {and} \bibinfo{person}{John Doherty}.} \bibinfo{year}{2017}\natexlab{}.
\newblock \showarticletitle{Committee-based active learning for surrogate-assisted particle swarm optimization of expensive problems}.
\newblock \bibinfo{journal}{\emph{IEEE transactions on cybernetics}} \bibinfo{volume}{47}, \bibinfo{number}{9} (\bibinfo{year}{2017}), \bibinfo{pages}{2664--2677}.
\newblock


\bibitem[Wang et~al\mbox{.}(2016)]%
        {trauma_systems2}
\bibfield{author}{\bibinfo{person}{Handing Wang}, \bibinfo{person}{Yaochu Jin}, {and} \bibinfo{person}{Jan~O Jansen}.} \bibinfo{year}{2016}\natexlab{}.
\newblock \showarticletitle{Data-driven surrogate-assisted multiobjective evolutionary optimization of a trauma system}.
\newblock \bibinfo{journal}{\emph{IEEE Transactions on Evolutionary Computation}} \bibinfo{volume}{20}, \bibinfo{number}{6} (\bibinfo{year}{2016}), \bibinfo{pages}{939--952}.
\newblock


\bibitem[Wang et~al\mbox{.}(2018a)]%
        {ensemble3}
\bibfield{author}{\bibinfo{person}{Handing Wang}, \bibinfo{person}{Yaochu Jin}, \bibinfo{person}{Chaoli Sun}, {and} \bibinfo{person}{John Doherty}.} \bibinfo{year}{2018}\natexlab{a}.
\newblock \showarticletitle{Offline data-driven evolutionary optimization using selective surrogate ensembles}.
\newblock \bibinfo{journal}{\emph{IEEE Transactions on Evolutionary Computation}} \bibinfo{volume}{23}, \bibinfo{number}{2} (\bibinfo{year}{2018}), \bibinfo{pages}{203--216}.
\newblock


\bibitem[Wang et~al\mbox{.}(2019)]%
        {island_large}
\bibfield{author}{\bibinfo{person}{Ting-Chen Wang}, \bibinfo{person}{Chih-Yu Lin}, \bibinfo{person}{Rung-Tzuo Liaw}, {and} \bibinfo{person}{Chuan-Kang Ting}.} \bibinfo{year}{2019}\natexlab{}.
\newblock \showarticletitle{Empirical analysis of island model on large scale global optimization}. In \bibinfo{booktitle}{\emph{2019 IEEE Congress on Evolutionary Computation (CEC)}}. IEEE, \bibinfo{pages}{342--349}.
\newblock


\bibitem[Wang et~al\mbox{.}(2021)]%
        {CSO2}
\bibfield{author}{\bibinfo{person}{Xiangyu Wang}, \bibinfo{person}{Kai Zhang}, \bibinfo{person}{Jian Wang}, {and} \bibinfo{person}{Yaochu Jin}.} \bibinfo{year}{2021}\natexlab{}.
\newblock \showarticletitle{An enhanced competitive swarm optimizer with strongly convex sparse operator for large-scale multiobjective optimization}.
\newblock \bibinfo{journal}{\emph{IEEE transactions on evolutionary computation}} \bibinfo{volume}{26}, \bibinfo{number}{5} (\bibinfo{year}{2021}), \bibinfo{pages}{859--871}.
\newblock


\bibitem[Wang et~al\mbox{.}(2018b)]%
        {affect2}
\bibfield{author}{\bibinfo{person}{Yuping Wang}, \bibinfo{person}{Haiyan Liu}, \bibinfo{person}{Fei Wei}, \bibinfo{person}{Tingting Zong}, {and} \bibinfo{person}{Xiaodong Li}.} \bibinfo{year}{2018}\natexlab{b}.
\newblock \showarticletitle{Cooperative coevolution with formula-based variable grouping for large-scale global optimization}.
\newblock \bibinfo{journal}{\emph{Evolutionary computation}} \bibinfo{volume}{26}, \bibinfo{number}{4} (\bibinfo{year}{2018}), \bibinfo{pages}{569--596}.
\newblock


\bibitem[Wei et~al\mbox{.}(2020)]%
        {CA-LLSO}
\bibfield{author}{\bibinfo{person}{Feng-Feng Wei}, \bibinfo{person}{Wei-Neng Chen}, \bibinfo{person}{Qiang Yang}, \bibinfo{person}{Jeremiah Deng}, \bibinfo{person}{Xiao-Nan Luo}, \bibinfo{person}{Hu Jin}, {and} \bibinfo{person}{Jun Zhang}.} \bibinfo{year}{2020}\natexlab{}.
\newblock \showarticletitle{A classifier-assisted level-based learning swarm optimizer for expensive optimization}.
\newblock \bibinfo{journal}{\emph{IEEE Transactions on Evolutionary Computation}} \bibinfo{volume}{25}, \bibinfo{number}{2} (\bibinfo{year}{2020}), \bibinfo{pages}{219--233}.
\newblock


\bibitem[Wu et~al\mbox{.}(2019)]%
        {island_cate1}
\bibfield{author}{\bibinfo{person}{Guohua Wu}, \bibinfo{person}{Rammohan Mallipeddi}, {and} \bibinfo{person}{Ponnuthurai~Nagaratnam Suganthan}.} \bibinfo{year}{2019}\natexlab{}.
\newblock \showarticletitle{Ensemble strategies for population-based optimization algorithms--A survey}.
\newblock \bibinfo{journal}{\emph{Swarm and evolutionary computation}}  \bibinfo{volume}{44} (\bibinfo{year}{2019}), \bibinfo{pages}{695--711}.
\newblock


\bibitem[Wu et~al\mbox{.}(2022)]%
        {ESCO}
\bibfield{author}{\bibinfo{person}{Xunfeng Wu}, \bibinfo{person}{Qiuzhen Lin}, \bibinfo{person}{Jianqiang Li}, \bibinfo{person}{Kay~Chen Tan}, {and} \bibinfo{person}{Victor~CM Leung}.} \bibinfo{year}{2022}\natexlab{}.
\newblock \showarticletitle{An Ensemble Surrogate-Based Coevolutionary Algorithm for Solving Large-Scale Expensive Optimization Problems}.
\newblock \bibinfo{journal}{\emph{IEEE Transactions on Cybernetics}} (\bibinfo{year}{2022}).
\newblock


\bibitem[Yang et~al\mbox{.}(2020)]%
        {erdg}
\bibfield{author}{\bibinfo{person}{Ming Yang}, \bibinfo{person}{Aimin Zhou}, \bibinfo{person}{Changhe Li}, {and} \bibinfo{person}{Xin Yao}.} \bibinfo{year}{2020}\natexlab{}.
\newblock \showarticletitle{An efficient recursive differential grouping for large-scale continuous problems}.
\newblock \bibinfo{journal}{\emph{IEEE Transactions on Evolutionary Computation}} \bibinfo{volume}{25}, \bibinfo{number}{1} (\bibinfo{year}{2020}), \bibinfo{pages}{159--171}.
\newblock


\bibitem[Yang et~al\mbox{.}(2017)]%
        {LLSO}
\bibfield{author}{\bibinfo{person}{Qiang Yang}, \bibinfo{person}{Wei-Neng Chen}, \bibinfo{person}{Jeremiah Da~Deng}, \bibinfo{person}{Yun Li}, \bibinfo{person}{Tianlong Gu}, {and} \bibinfo{person}{Jun Zhang}.} \bibinfo{year}{2017}\natexlab{}.
\newblock \showarticletitle{A level-based learning swarm optimizer for large-scale optimization}.
\newblock \bibinfo{journal}{\emph{IEEE Transactions on Evolutionary Computation}} \bibinfo{volume}{22}, \bibinfo{number}{4} (\bibinfo{year}{2017}), \bibinfo{pages}{578--594}.
\newblock


\bibitem[Yu et~al\mbox{.}(2019)]%
        {SL-PSO}
\bibfield{author}{\bibinfo{person}{Haibo Yu}, \bibinfo{person}{Ying Tan}, \bibinfo{person}{Chaoli Sun}, {and} \bibinfo{person}{Jianchao Zeng}.} \bibinfo{year}{2019}\natexlab{}.
\newblock \showarticletitle{A generation-based optimal restart strategy for surrogate-assisted social learning particle swarm optimization}.
\newblock \bibinfo{journal}{\emph{Knowledge-Based Systems}}  \bibinfo{volume}{163} (\bibinfo{year}{2019}), \bibinfo{pages}{14--25}.
\newblock


\bibitem[Zeng et~al\mbox{.}(2020)]%
        {dynamic_neighbor}
\bibfield{author}{\bibinfo{person}{Nianyin Zeng}, \bibinfo{person}{Zidong Wang}, \bibinfo{person}{Weibo Liu}, \bibinfo{person}{Hong Zhang}, \bibinfo{person}{Kate Hone}, {and} \bibinfo{person}{Xiaohui Liu}.} \bibinfo{year}{2020}\natexlab{}.
\newblock \showarticletitle{A dynamic neighborhood-based switching particle swarm optimization algorithm}.
\newblock \bibinfo{journal}{\emph{IEEE Transactions on Cybernetics}} \bibinfo{volume}{52}, \bibinfo{number}{9} (\bibinfo{year}{2020}), \bibinfo{pages}{9290--9301}.
\newblock


\bibitem[Zhao et~al\mbox{.}(2008)]%
        {zhao2008dynamic}
\bibfield{author}{\bibinfo{person}{Shi-Zheng Zhao}, \bibinfo{person}{Jing~J Liang}, \bibinfo{person}{Ponnuthurai~N Suganthan}, {and} \bibinfo{person}{Mehmet~Fatih Tasgetiren}.} \bibinfo{year}{2008}\natexlab{}.
\newblock \showarticletitle{Dynamic multi-swarm particle swarm optimizer with local search for large scale global optimization}. In \bibinfo{booktitle}{\emph{2008 IEEE congress on evolutionary computation (IEEE world congress on computational intelligence)}}. IEEE, \bibinfo{pages}{3845--3852}.
\newblock


\bibitem[Zhen et~al\mbox{.}(2022)]%
        {MS-DDEO}
\bibfield{author}{\bibinfo{person}{Huixiang Zhen}, \bibinfo{person}{Wenyin Gong}, {and} \bibinfo{person}{Ling Wang}.} \bibinfo{year}{2022}\natexlab{}.
\newblock \showarticletitle{Offline data-driven evolutionary optimization based on model selection}.
\newblock \bibinfo{journal}{\emph{Swarm and Evolutionary Computation}}  \bibinfo{volume}{71} (\bibinfo{year}{2022}), \bibinfo{pages}{101080}.
\newblock


\bibitem[Zhong et~al\mbox{.}(2024)]%
        {SDDObench}
\bibfield{author}{\bibinfo{person}{Yuanting Zhong}, \bibinfo{person}{Xincan Wang}, \bibinfo{person}{Yuhong Sun}, {and} \bibinfo{person}{Yue-Jiao Gong}.} \bibinfo{year}{2024}\natexlab{}.
\newblock \showarticletitle{SDDObench: A Benchmark for Streaming Data-Driven Optimization with Concept Drift}. In \bibinfo{booktitle}{\emph{Proceedings of the Genetic and Evolutionary Computation Conference}} (Melbourne, VIC, Australia) \emph{(\bibinfo{series}{GECCO '24})}. \bibinfo{publisher}{Association for Computing Machinery}, \bibinfo{address}{New York, NY, USA}, \bibinfo{pages}{59–67}.
\newblock
\showISBNx{9798400704949}
\href{https://doi.org/10.1145/3638529.3654063}{doi:\nolinkurl{10.1145/3638529.3654063}}


\bibitem[Zhou et~al\mbox{.}(2022)]%
        {blast_furnace}
\bibfield{author}{\bibinfo{person}{Qun Zhou}, \bibinfo{person}{Yongliang Yin}, \bibinfo{person}{Daogang Peng}, \bibinfo{person}{Huirong Zhao}, \bibinfo{person}{Lei Xing}, \bibinfo{person}{Xuebin Jiang}, \bibinfo{person}{Zhenchao Xu}, {and} \bibinfo{person}{Chunmei Xu}.} \bibinfo{year}{2022}\natexlab{}.
\newblock \showarticletitle{Multi-objective optimization of blast furnace dosing and operation based on NSGA-II}. In \bibinfo{booktitle}{\emph{2022 4th International Conference on Electrical Engineering and Control Technologies (CEECT)}}. IEEE, \bibinfo{pages}{165--169}.
\newblock


\bibitem[Zhou et~al\mbox{.}(2006)]%
        {ensemble_a}
\bibfield{author}{\bibinfo{person}{Zongzhao Zhou}, \bibinfo{person}{Yew~Soon Ong}, \bibinfo{person}{Prasanth~B Nair}, \bibinfo{person}{Andy~J Keane}, {and} \bibinfo{person}{Kai~Yew Lum}.} \bibinfo{year}{2006}\natexlab{}.
\newblock \showarticletitle{Combining global and local surrogate models to accelerate evolutionary optimization}.
\newblock \bibinfo{journal}{\emph{IEEE Transactions on Systems, Man, and Cybernetics, Part C (Applications and Reviews)}} \bibinfo{volume}{37}, \bibinfo{number}{1} (\bibinfo{year}{2006}), \bibinfo{pages}{66--76}.
\newblock


\end{thebibliography}

\appendix

\end{document}